%% file: rss-main.tex
\documentclass[conference]{IEEEtran}
\IEEEoverridecommandlockouts 

\usepackage{flushend}
\usepackage{balance} 
\let\labelindent\relax

\input{0-preamble}

\input{0-user-macros}

\graphicspath{Figures/}

\let\oldthebibliography\thebibliography
\renewcommand\thebibliography[1]{%
  \oldthebibliography{#1}%
  \setlength{\parskip}{0pt}%
  \setlength{\itemsep}{0pt}%
}

\usepackage{rotating}
\usepackage{placeins}
\usepackage{import}

\input{math_commands.tex}

\definecolor{Lightapricot}{rgb}{0.99,0.84,0.69}

\newcommand{\ourmodel}{\textsc{RePLan}\xspace}

\begin{document}

\makeatletter
    \let\@oldmaketitle\@maketitle
    \renewcommand{\@maketitle}{\@oldmaketitle
    \centering
    \includegraphics[width=\linewidth]{figures/overview.pdf}
    \captionof{figure}{\textbf{Overview of \ourmodel} using an example user goal: ``place apple in the bowl". \ourmodel generates a high-level plan of a possible solution, followed by low-level reward functions. If it is unable to accomplish a subtask, perception is used to diagnose any issues that may be present in the scene. For example, in this instance there is a lemon already in the bowl, which the robot must first remove. \ourmodel generates a plan to remove the obstacle and continue with the task. Once completing all the subtasks, if the original goal still is not accomplished, \ourmodel can reason at a higher level to try a new solution. Meanwhile, prompting vanilla GPT-4V with an image of the scene and the user goal returns an infeasible plan because it does not take into account object states.}
    \label{fig:overview}      
    }%
\makeatother

\title{\ourmodel: Robotic Replanning with Perception and Language Models}
\author{\IEEEauthorblockN{Marta Skreta\IEEEauthorrefmark{1}, Zihan Zhou\IEEEauthorrefmark{1}, Jia Lin Yuan\IEEEauthorrefmark{1}, Kourosh Darvish, Alán Aspuru-Guzik, and Animesh Garg}
\IEEEauthorblockA{
University of Toronto, Vector Institute for Artificial Intelligence, Georgia Institute of Technology, NVIDIA \\
{\small\texttt{\{martaskreta, footoredo, jly, kdarvish\}@cs.toronto.edu, alan@aspuru.com, animesh.garg@gatech.edu}}
\\
\IEEEauthorrefmark{1}These authors contributed equally to this work.}
}

\maketitle
\IEEEpeerreviewmaketitle

\begin{abstract}
Advancements in large language models (LLMs) have demonstrated their potential in facilitating high-level reasoning, logical reasoning and robotics planning. Recently, LLMs have also been able to generate reward functions for low-level robot actions, effectively bridging the interface between high-level planning and low-level robot control. However, the challenge remains that even with syntactically correct plans, robots can still fail to achieve their intended goals due to imperfect plans or unexpected environmental issues. To overcome this,   Vision Language Models (VLMs) have shown remarkable success in tasks such as visual question answering. Leveraging the capabilities of VLMs, we present a novel framework called Robotic \textbf{Re}planning with \textbf{P}erception and \textbf{Lan}guage Models (\ourmodel) that enables online replanning capabilities for long-horizon tasks. This framework utilizes the physical grounding provided by a VLM's understanding of the world's state to adapt robot actions when the initial plan fails to achieve the desired goal.
We developed a \textbf{R}easoning and \textbf{C}ontrol (RC) benchmark with eight long-horizon tasks to test our approach.
We find that \ourmodel enables a robot to successfully adapt to unforeseen obstacles while accomplishing open-ended, long-horizon goals, where baseline models cannot, and can be readily applied to real robots. Find more information at \href{https://replan-lm.github.io/replan.github.io/}{https://replan-lm.github.io/replan.github.io/}
\end{abstract}

\section{Introduction}
\label{intro}

Designing embodied agents to execute multi-stage, long-horizon tasks is challenging. Firstly, agents need manipulation skills for physical engagement with their environment. They also need to be adept at perceiving their surrounding environment and reasoning on cause-and-effect relationships of their actions on the environment. Moreover, these agents should be able to plan and carry out a series of actions that are in line with the main goals they are tasked to accomplish~\citep{wu2023m}, with minimal human intervention.

Methods based on rule-driven frameworks like Task and Motion Planning (TAMP)~\citep{garrett2021integrated} and learning approaches, such as Hierarchical Reinforcement Learning (HRL) and Imitation Learning (IL), have advanced the field of long-horizon planning. Yet, these methods often require extensive domain knowledge, intricate reward engineering, and time-consuming dataset creation efforts~\citep{hussein2017imitation, brohan2022rt}. 

In contrast, the rise of Large Language Models (LLMs) has shown considerable promise in robot planning~\citep{driess2023palm, brohan2023rt}. The integration of LLMs into complex, long-horizon tasks in robot planning still poses substantial challenges. We identify four primary areas where these challenges are most pronounced, which we expand on below: a) long-horizon plan generation, b) verifying plan correctness, c) obtaining feedback to adjust plans online, and d) open-ended task execution.  

Long-horizon, multi-stage task planning requires reasoning over extended periods, which is a challenge for LLMs \cite{wang2023describe}. In the context of LLMs, tackling large-scale problems often leads to issues like hallucination or failing to consider important details, rendering their plans ineffective or error-ridden~\citep{bang2023multitask, wang2023survey}. To address this, prompting schemes like ReAct and Chain-of-Thought distill complex problems into intermediate reasoning steps with the aid of exemplars, facilitating effective reasoning~\citep{yao2023react, wei2022chain}. 
However, these efforts are still constrained by the number of stages the robot can handle.

Another challenge in using LLMs for task planning is verifying that their outputs are correct, such as syntax checking~\citep{skreta2023errors, wang2023voyager, darvish2024organa} and semantics~\citep{rana2023sayplan} verification, as well as simulating task execution ~\citep{liu2022mind} to provide success/failure feedback to the LLM. Verification enables the ability to plan over multiple steps in complex domains. 


Working with multi-step tasks in robotics also involves dealing with uncertainties and changes in the environment. Effectively handling these tasks requires combining task instructions with sensory data, which helps the robot adapt to the changing surroundings.
Recent research has shown that Vision-Language Models (VLMs) can guide robots in interpreting their environment more accurately and generalizing to unseen scenarios. We find that VLMs are able to provide more accurate feedback about object states compared with LLMs. When converting object states to pose estimations and asking an LLM to reason about obstacles, we find that it is uncapable of inferring spatial relations between objects (Appendix~\ref{sec:no_vlm_diagnosis}).   By integrating visual cues with linguistic context, VLMs enable robots to better interpret their surrounding environment~\citep{brohan2023rt} and can even provide rewards for training RL agents ~\citep{baumli2023vision}. However, prompting VLMs to directly generate out-of-the-box plans for multi-stage tasks may not work as task complexity increases. For example, in Figure~\ref{fig:overview} and Appendix~\ref{sec:gpt_4v_baseline}, we show that prompting GPT-4V~\citep{gpt4v-2023}(state-of-the-art VLM with public access) does not take into account environment obstacles, and therefore generates an unfeasible plan.

\begin{table*}[!t]
  \centering
  \caption{Comparison of \ourmodel with other LLM-based multi-stage planners. Autonomous online replanning refers to replanning without human intervention. If a feature is unclear from the paper description, it is denoted by a question mark.}
  \label{tab:paper_comparison}
    \small 
    \setlength{\tabcolsep}{3pt}  
    \resizebox{\textwidth}{!}{%
  \begin{tabular}{ccccccccc}
    \toprule
    \textbf{Paper} & \specialcell{No human \\assistance} & \specialcell{Open-ended \\ problem-solving} & \specialcell{Verifier} & \specialcell{ Sensory \\ feedback} & \specialcell{Reward \\generator } & \specialcell{Unstructured state\\ representation } & \specialcell{No \\fine-tuning}  & \specialcell{Autonomous online \\replanning } \\
    \midrule
    Text2Reward~\cite{xie2023text2reward} & $\times$ & $\times$ & \checkmark & \checkmark  & \checkmark & $\times$ & $\times$ & $\times$   \\ 
    \specialcell{Language to Rewards~\cite{yu2023language}} & $\times$ & $\times$ & $\times$ & \checkmark &  \checkmark & \checkmark & \checkmark & $\times$ \\ 
    REFLECT~\cite{liu2023reflect} & \checkmark & \checkmark & \checkmark & \checkmark & $\times$  & $\times$ & \checkmark & \checkmark  \\ 
    DoReMi~\cite{guo2023doremi} & \checkmark & $\times$ & \checkmark & \checkmark & $\times$ & \checkmark& \checkmark & \checkmark \\
    DROC~\cite{zha2023distilling} & $\times$ & $\times$ & \checkmark & $\times$ & $\times$ & \checkmark & \checkmark  &  $\times$\\
    SayCan~\cite{brohan2023can} & \checkmark &\checkmark & $\times$ &   \checkmark & $\times$ &  \checkmark &  \checkmark & $\times$ \\
    Inner Monologue~\cite{huang2022inner} & $\times$ & \checkmark & \checkmark & \checkmark & $\times$ & \checkmark & ? & \checkmark \\
    \midrule
    \ourmodel & \checkmark & \checkmark & \checkmark & \checkmark &  \checkmark & \checkmark & \checkmark  & \checkmark \\
    \bottomrule
  \end{tabular}
    }
\end{table*}

In order for robots to generalize across diverse environments, the acquisition of rich low-level skills is essential~\citep{wang2023voyager, zhou2023learning}. Many works have utilized pre-trained and encoded skills, supplying a skill list to LLMs for skill selection~\citep{lin2023text2motion}. It should also be noted that merely relying on user-provided skills is insufficient for enabling robots to operate effectively in an open-ended world. Overcoming this limitation necessitates a focus on reward generation strategies. Recently, there has been a shift towards directly setting up 
rewards generated by LLMs~\citep{yu2023language, kwon2023reward, xie2023text2reward}. This approach, employed in both RL for policy learning~\citep{kwon2023reward, xie2023text2reward} and Model Predictive Control (MPC)~\citep{garcia1989model, rawlings2000tutorial} to enhance data efficiency~\citep{miyaoka2023chatmpc, yu2023language}, enables users to more easily guide robot behavior by creating and combining rewards.

In this paper, we present a framework called Robotic \textbf{Re}planning with \textbf{P}erception and \textbf{Lan}guage Models (\ourmodel) that enables multi-stage, long-horizon task execution without human intervention. Specifically, our contributions are the following:
\begin{enumerate}[
    topsep=0pt,
    noitemsep,
    leftmargin=*,
    ]
    \item \ourmodel: an end-to-end framework that does multi-level replanning with verification through zero-shot models, incorporating high-level planning and low-level reward generation (Figure~\ref{fig:overview})
    \item We use vision-language signals for grounding and closing the \ourmodel feedback loop as an error correction system
    \item We demonstrate that \ourmodel can solve tasks that are (a) ambiguous (i.e., the correct solution to the problem is not known at the beginning) and (b) adapt its plan online as challenges are encountered during task execution without human intervention
    \item We present a Reasoning and Control (RC) benchmark of eight tasks that combine both of these aspects. We show that \ourmodel is able to succeed almost $4\times$ as often compared to baselines and can perform real-world experiments . 
\end{enumerate}

\section{Related work}
\label{sec:related_work}

\textbf{Long-horizon Robot Planning.}
Addressing long-horizon planning in robotics has been a persistent challenge. Rule-based methods~\citep{10.1126/science.abc2986, baier2009heuristic}, such as Planning Domain Definition Language (PDDL)~\citep{aeronautiques1998pddl}, attempted to solve task and motion planning sequentially, however, the planned task may not feasible when the domain knowledge is incomplete. The task and motion planning (TAMP) approach~\citep{garrett2021integrated}, addresses this by simultaneously determining symbolic actions and low-level motions.
While these methods excel in \textit{verifying} task and motion plans during planning, their \textit{generalization} to new environments and tasks as well as their efficiency on replanning are constrained.
In addressing multi-stage planning challenges, many works focus on learning task plans from input task specifications, leveraging reinforcement learning (RL) and imitation learning (IL) techniques. For example, Behavior-1K~\citep{li2023behavior} employs RL to acquire semantics and physical manipulation skills, often benefiting from classical motion planners and simplifying assumptions.

However, it's important to note that these learning-based techniques demand significant domain expertise for reward engineering and rely on large datasets for task learning~\citep{heo2023furniturebench}. While they adeptly \textit{react} to environmental uncertainties by iteratively updating policies based on observations, their zero-shot generalization across multi-stage tasks remains a persistent challenge.


\textbf{Robot Control with Physically Grounded Language Models.}
Recent advancements in LLMs have resulted in their adoption in robot planning, leveraging their natural language capabilities and common-sense reasoning for generating robot task and motion plans\citep{wang2023survey, xi2023rise}. Notably, LLMs have been applied to planning multi-stage tasks~\citep{singh2022progprompt, driess2023palm}, by utilizing LLMs to improve sample efficiency in reinforcement learning. 
 ProgPrompt and Code-As-Policies are among the early approaches using code-writing LLMs to generate code for robot policies~\citep{singh2022progprompt, liang2022code}. However,  generated codes are prone to compilation errors.

Language models with a verifier have been used for generating long-horizon tasks in an iterative prompting technique with lower error rates~\citep{yoshikawa2023large}; however, there is no guarantee that the output task plan can be executed. To overcome that shortcoming, SayPlan used LLMs to reason over scene graphs and generate plans across large environments, using iterative replanning to ensure scene graph constraints were not violated~\citep{rana2023sayplan}. 

Toward grounding the language belief with visual and motor control feedback, several works have employed a variation of vision-language models with agent embodiment for solving planning problems~\citep{ha2023scaling, brohan2023rt, brohan2023can, huang2022inner}. Table~\ref{tab:paper_comparison} provides a detailed comparison of \ourmodel with most related works.

\textbf{Language to Reward Shaping.}
In contrast to approaches that map natural task descriptions to robot actions and subsequently to rewards, an alternative approach seeks to directly infer rewards from natural language inputs, addressing the challenge of reward engineering~\citep{lin2022inferring}. This language-driven reward-shaping approach has demonstrated utility in various domains, including negotiation~\citep{kwon2023reward} and gaming~\citep{goyal2019using}, facilitating desired behavior learning through RL.
\citep{mahmoudieh2022zero} introduce a visuo-language model that generates robot motion policy reward on goal text and raw pixel observations, in a manner similar to ~\citep{radford2021learning}, enabling zero-shot prompting for unseen scenarios.
\citep{yu2023language} employs an iterative prompting method using an LLM to link user task specifications and robot motion through reward functions. While excelling in motion generation with minimal data, their approach falls short in handling long-horizon multistage tasks and lacks real-time environment feedback, necessitating user intervention for adaptation.~\citep{xie2023text2reward} extended the previous work for robot reward policy refinement by requiring substantial human involvement and Pythonic observation from the environment. Both of these methods struggle with open-ended problems and multi-stage tasks. To mitigate these limitations, our work autonomously performs long-horizon tasks and adapts to execution outcomes by leveraging motor control and raw visual feedback.

\textbf{Online Replanning Based on Sensory Feedback.} Recent works have explored online robotic adaption for recovering from errors during task execution. CAPE queries an LLM for a possible corrective action if a precondition is not met~\citep{raman2023cape}. DROC introduces a framework that can provide a robot with low-level and high-level corrections, which are then stored in a knowledge base that can be retrieved later~\citep{zha2023distilling}. However, the corrections must be provided by humans, whereas our framework does not require any human intervention. DoReMi uses an LLM to generate constraints that must be met during a given subtask that the robot is executing, and then queries a VLM to ensure those preconditions are met~\citep{guo2023doremi}. However, this framework has to rely on constraints that an LLM generates and only probes the VLM for a yes-or-no answer, which may miss unexpected obstacles, and does not allow for tasks which have ambiguous solutions. The framework also queries a VLM every $t$ seconds, which may increase latency depending on the VLM used. REFLECT uses audio and video sensors to hierarchically generate captions for object states, and then uses an LLM to reason upon possible failure reasons if a subtask is not accomplished~\citep{liu2023reflect}. However, this method requires predefining caption states and sensor labels. 
Furthermore, none of these methods allow for task completion where the task is ambiguous (except for CAPE and REFLECT), and none output unconstrained robot rewards directly.

\begin{figure*}
     \centering         \includegraphics[width=0.95\textwidth]{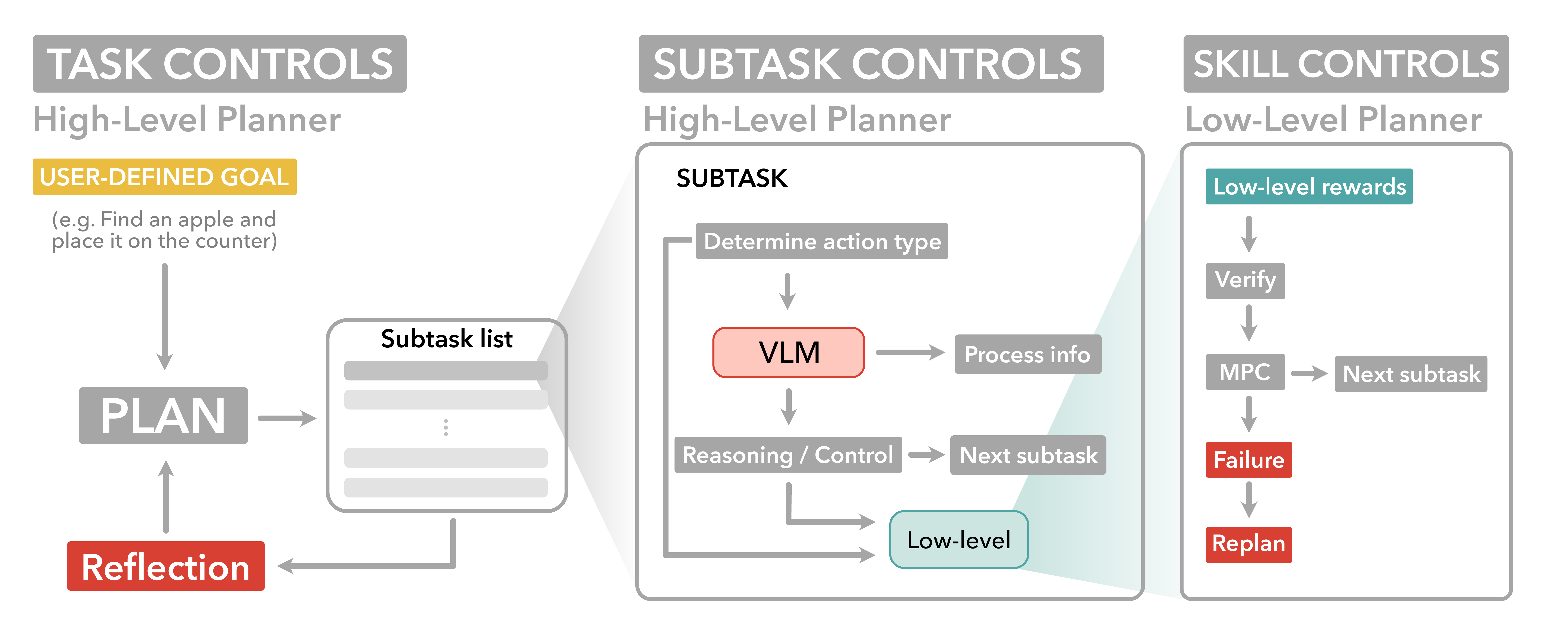}
    \caption{\textbf{\ourmodel model architecture for hierarchical reasoning and control}. In the Task Control level, \ourmodel first generates high-level subtasks conditioned on the user's goal and a scene image. Next, the Subtask Control level determines what action types need to be taken to execute the subtask. If the action requires obtaining information about object attributes or states, the VLM Perceiver may be called to answer any questions. VLM answers are reasoned upon and used to update the world knowledge of the Planners. Otherwise, if the subtasks require low-level actions, a low level motion plan is generated and passed to the Low-Level Planner at the Skill Controls level. The Planner generates robot skill-level rewards for execution via MPC. If failures occur, feedback is provided to subtask and task controls for replanning. If the goal is not reached even after replanning and completing the subtasks, the High-Level Planner reflects on past experiences and proposes a new plan. A more detailed version is shown in Appendix Figure~\ref{fig:roadmap}.
 }
    \label{fig:pipeline}
\end{figure*}

\section{\ourmodel: Model Architecture}

We present an overview of our method in Figure~\ref{fig:pipeline}. The input to our system is a goal described in natural language. The goal can be specific (e.g. place kettle on stove) or open-ended (e.g. search for the banana).
\ourmodel has five modules, which are described below. All prompts used for the modules can be found in Appendix \ref{sec:prompts}.
\begin{enumerate}[
    topsep=0pt,
    noitemsep,
    leftmargin=*,
    ]
  \item a High-Level LLM Planner for planning, replanning, reflecting, and reasoning
  \item a VLM Perceiver for physically-grounded reasoning
    \item a Low-Level LLM Planner for converting high-level tasks to low-level rewards
  \item a Motion Controller for low-level robot actions
  \item a LLM Verifier to check that the Planner/Perceiver is correct and fixes them if applicable
\end{enumerate}

\subsection{High-Level LLM Planner}

Inspired by the ability of LLMs to generate actionable steps from high-level tasks \citep{ huang2022language}, we employ a High-Level Planner to take in as input a user-specified task and return a list of subtasks on how a robot should accomplish the task. We use a prompting scheme similar to ReAct \cite{yao2023react} for generating subtasks. The benefit of using a High-Level Planner is that there are no restrictions on the abstraction level of the user input. The user input can be a specific task (e.g. ``Place my keys on the counter''), or an open-ended task where the procedure requires exploration (e.g. ``Find my keys''). This is because the Planner can propose a procedure, even if it doesn't know the exact answer a priori. If the robot completes the sequence of actions proposed by the Planner and the overall goal is still not accomplished, the High-Level Planner can propose a new procedure. The High-Level Planner can also utilize the past recommended procedures to prevent the redundancy of having the agent perform the same tasks repetitively \citep{skreta2023errors}. The High-Level Planner is also used to incorporate feedback from perception models when generating high-level plans. This is important because the Planner should not generate plans that are not aligned with the physical state of the world, or should be able to replan if there are any obstacles. 

\subsection{VLM Perceiver}
While LLMs have demonstrated powerful reasoning skills over text, they lack grounding in the physical world \citep{liu2022mind}. Consequently, although LLMs can generate plans that sound reasonable, they may fail to account for any obstacles or uncertainties that are present because they cannot perceive the environment. In constrast, while VLMs offer physical grounding to text queries, they are much weaker in language-only reasoning than LLMs. Hence, we use the High-Level Planner to decide what it wants to query from the Perceiver, and then the Planner incorporates feedback from the Perceiver when it needs to know about the object states or replan because the robot failed to do an action. The High-Level LLM-based Planner decides on simple, yet specific, questions to ask the Perceiver (Figures \ref{fig:llm_vlm_question}-\ref{fig:vlm_object_state}) and samples multiple answers before consolidating them into a summary observation that is consistent with the environment state (Figures \ref{fig:vlm_action_failed_new}-\ref{fig:llm_summary}).

\subsection{Low-Level LLM Planner}
Recently, it has been demonstrated that LLMs are capable of producing low-level plans that enable robot motion control~\citep{yu2023language, xie2023text2reward}. This is exciting because it bridges the gap between high-level, human-specified goals and robot actions in a zero-shot manner (without the need for extensive training datasets). However, while previous works are able to generate low-level robot actions for a concrete task (e.g. ``Open the drawer"), we find that they fail when asked to generate plans for long-horizon, open-ended tasks. Thus, we utilize the High-Level Planner to generate concrete subtasks from a high-level goal, which are then passed to the Low-Level Planner to generate the corresponding low-level actions. Our Low-Level Planner uses the same Reward Translator as in \citep{yu2023language} which we have found works well. The Low-Level Planner works in two stages. First, it generates a motion plan from a user-specified input. The motion plan is a natural language description of the actions a robot should do to achieve the goal. Then, the motion plan is then translated to reward functions, which serve as a representation of the desired robot motion. These reward functions are then passed to the Motion Controller.

\subsection{Motion Controller}
\begin{table*}[ht]
  \centering
  \caption{Different features in \ourmodel benchmark tasks.}
  \label{tab:env_comparison}
    \small 
    \setlength{\tabcolsep}{3pt}  
    \resizebox{.8\textwidth}{!}{%
  \begin{tabular}{ccccccccc}
    \toprule
    \textbf{Feature} & \specialcell{\textit{Cabinet-} \\\textit{Open}} & \specialcell{\textit{Cabinet-} \\ \textit{Closed}} & \specialcell{\textit{Cabinet-}\\ \textit{Blocked}} & \specialcell{ \textit{Cabinet-}\\ \textit{Locked} } & \specialcell{\textit{Cubes-}\\ \textit{Color}} & \specialcell{\textit{Cubes-}\\ \textit{Blocked}} & \specialcell{\textit{Kitchen-}\\ \textit{Explore}}  & \specialcell{\textit{Composite-}\\ \textit{Explore}} \\
    \midrule
    
     Multi-step planning& $\times$ & \checkmark & \checkmark & \checkmark  & $\times$ & \checkmark & \checkmark & \checkmark   \\
     Visual feedback& $\times$ & $\times$ & \checkmark & \checkmark  & \checkmark & \checkmark & \checkmark & \checkmark   \\
     Causal reasoning& $\times$ & $\times$ & \checkmark & \checkmark  & $\times$ & \checkmark & \checkmark & \checkmark   \\
     Exploration& $\times$ & $\times$ & $\times$ & $\times$  & $\times$ & $\times$ & \checkmark & \checkmark   \\
    \bottomrule
  \end{tabular}
    }
\end{table*}
 
The Motion Controller receives reward functions and instructs the robot on what actions to take in order to satisfy those functions. For motion control, we use MuJoCo MPC (MJPC), an open-source real-time predictive controller, implemented on MuJoCo physics~\citep{howell2022}. Given the initial condition $x_0$, the control problem is defined as:
\begin{equation*}
    \underset{x_{1:{T}}, u_{1:T}}{\text{minimize }}  \quad \sum \limits_{t = 0}^{T} c(x_t, u_t),~~~
    \text{subject to}  \quad x_{t+1} = f(x_t, u_t),
\end{equation*}
where $x_t$ and $u_t$ are the state and control signals at time step $t$, and the mapping $f$ is the transition dynamics. The goal is to minimize the cost function $c(.)$ along the trajectory from the current time step to the horizon $T$. We define the $M$ output rewards provided by the low-level planner as the negative of the cost function, i.e., $c(x_t, u_t) = -\sum \limits_{i = 1}^{M} w_i~r_i(x_t, u_t, \phi_i)$, where $\phi_i$ and $w_i$ are the $i$'th reward parameters and weight.
To solve the optimization problem in our work, the predictive sampling implementation is used~\citep{howell2022}.

A subtask can have more than one reward function, the Low-Level Planner also reasons about which reward function actually determines the success state of the action. For example, for the subtask: \texttt{Move the kettle away from the microwave}, the Planner generates:

\begin{lstlisting}[language=Python]
minimize_l2_distance_reward("palm", "kettle")
maximize_l2_distance_reward("kettle", "microwave_handle")
\end{lstlisting}

The Planner is able to reason that once the function \code{maximize\_l2\_distance\_reward ("kettle", "microwave\_handle")} has been satisfied, the kettle has been moved from the microwave. Thus, the Planner correctly labels this as the primary reward function.

\subsection{LLM \& VLM Verifier}
LLM outputs can contain errors and are often non-deterministic. One way to increase consistency and reliability is to verify the outputs~\citep{skreta2023errors}. We do that by taking plans produced by both the High-Level and Low-Level Planners and asking the Verifier to check that every step proposed is necessary to achieve the goal. For the Low-Level Planner, the Verifier is used to determine whether each step in the generated motion plan is useful for completing the subtask. This eliminates any unrelated actions that the Low-Level Planner. Motion control is a hard problem, eliminating any unnecessary actions increases the chances of the subtask being completed. 

Since the VLMs are prone to hallucination~\citep{dai-etal-2023-plausible}, they are used in a very constrained manner. Thus, the Verifier also corrects any observations made by the Perceiver based on objects that it knows exist in the environment. For instance, VLMs can identify an object using different synonyms. While it may be easy for humans to disambiguate, a robot could require instructions that adhere to strict syntax rules. Thus, the Verifier ensures that plans generated using perceived objects conform to the robot-interpretable entity names.

\section{Reasoning \& Control (RC) Benchmark}

In order to assess the long-term planning, as well as the logical and low-level planning capabilities of our system, we devised a new benchmark of eight long-horizon tasks that demand causal reasoning and exploration (Table~\ref{tab:env_comparison}) for evaluation. The readers can refer to Figure~\ref{fig:env-init} or our website for visualizations and Appendix~\ref{sec:env-details} for detailed descriptions.\\
\textbf{Cabinet-Open} A scene with an opened cabinet and a cube. The robot is required to put the cube inside the cabinet.\\ 
\textbf{Cabinet-Closed} Same as \textit{Cabinet-Open} except that the cabinet is closed.\\  
\textbf{Cabinet-Blocked} Same as \textit{Cabinet-Closed} except that there is a bar blocking the cabinet handles.\\ 
\textbf{Cabinet-Locked} The robot is tasked to retrieve a cube from a cabinet locked by a lever.\\  
\textbf{Cubes-Color} There are two cubes with different colors and one crate. The task is to move the cube that matches the color of the crate into the crate.\\  
\textbf{Cubes-Blocked} \textit{Cubes-Color}, however, a cube with the wrong color is in the crate. It needs to be moved first.\\ 
\textbf{Kitchen-Explore} Explore a complex kitchen scene, in search of a green apple while dealing with obstacles.\\  
\textbf{Composite-Explore} The task is to open a sliding cabinet, which is locked by a weight sensor. The robot has to explore the scene to locate the weight.  

These environments were implemented in MuJoCo \citep{todorov2012mujoco}. We used furniture\_sim\footnote{https://github.com/vikashplus/furniture\_sim/} for the assets. We used the MuJoCo MPC~\citep{howell2022} to generate the motion control.

\section{Experimental Evaluation}
\subsection{Experiment Setup}

We evaluate our framework using a  Franka Emika Panda arm robot equipped with a Robotiq 2F-85 gripper and a ZED Mini camera mounted on the gripper. We use MuJoCo MPC for physics simulation and real-time predictive control~\citep{howell2022}. 

For the LLM modules in \ourmodel, we use OpenAI GPT \texttt{GPT-4}~\cite{openai2023gpt4}. 
For the VLM Perceiver, we used \texttt{Qwen-VL-Chat-7B}~\citep{Qwen-VL} (except for \textit{Cubes-Blocked}, where we used \texttt{GPT-4V} due to hardware constraints). We show the performance of state-of-the-art VLMs on Perceiver tasks in Appendix~\ref{sec:vlm_ablation}. We found that from the open-source models we tested, \texttt{Qwen-VL-Chat-7B} has the best object reasoning skills; however, its object recognition capabilities improve when we first segment the scene using a segmentation model (we use Segment Anything ~\cite{kirillov2023segany}). \texttt{GPT-4V} had the best performance overall across object recognition and object reasoning, but there is a restrictive rate limit on its API. 


\subsection{Baselines and Ablations}

We compare our method to the Language to Rewards \citep{yu2023language} framework, a one-shot, in-context learning agent.  Language to Rewards uses a Reward Translator to translate a high-level goal (such as ``open the drawer'') to low-level reward functions that are used by a Motion Controller to instruct a robot on what to do. While Language to Rewards does not utilize a VLM to perceive the scene, we give it access to the same objects as our model identifies at the start of the scene. We also show sample plans using PDDL and PDDLStream in Appendix~\ref{sec:pddl_exps} for \textit{Cabinet-\{Open, Closed, Blocked\}} and GPT-4V for \textit{Cabinet-Blocked} and \textit{Cubes-Color} in Appendix~\ref{sec:gpt_4v_baseline}. 
Notably, since our focus is on zero-shot online (re)planning, we do not use RL based baselines, including recent LLMs to reward models. However, we note that if pretrained multi-task controllers were available, they could be utilized in low-level motion generation. 

Finally, to demonstrate the importance of all the modules in our pipeline, we do an ablation study on how well the robot can perform each task without each module. We systematically remove the following modules: VLM Perceiver, LLM Verifier, and replanning of High-Level Planner. We also provide additional ablations for choice of VLM (Appendix~\ref{sec:vlm_ablation}) as well as comparisons to naive GPT-4V (Appendix~\ref{sec:gpt_4v_baseline}) and TAMP (Appendix~\ref{sec:pddl_exps}).

\begin{table}[!t]
  \centering
  \footnotesize
  \setlength{\tabcolsep}{2pt}
    \caption{Completion rates in all 8 tasks (measured with 10 runs each). Average completion rates across 8 tasks are listed in the last column.}
\scalebox{0.99}{
  \begin{tabular}{lrrrrr}
      \toprule
     \textbf{Task} & \ourmodel & \textbf{$-$verifier} & \textbf{$-$perceiver}  & \textbf{$-$replan} & \textbf{L2R}  \\
    \midrule
    \textit{CabinetOpen} & \textbf{100\%} & 80\% & 90\% & 80\%  & 90\% \\
    \textit{CabinetClosed} & \textbf{100\%} & 80\% & 90\% & 70\% & 20\% \\
    \textit{CabinetBlocked} & \textbf{60\%} & 20\% & 30\% & 0\% &  0\% \\
    \textit{CabinetLocked} & \textbf{100\%} & \textbf{100\%} & 50\% & 0\%  & 0\% \\
    \midrule
    \textit{CubesColor} & 90\% & \textbf{100\%} &  20\% & 80\% & 50\% \\
    \textit{CubesBlocked} & \textbf{90\%} & 80\% & 0\% & 10\% & 10\% \\
    \midrule
    \textit{KitchenExplore} & \textbf{80\%} & 60\% & 20\%  & 0\% & 0\% \\
    \textit{CompositeExplore} & \textbf{70\%} & 20\% & 0\% & 20\%  & 0\% \\
    \midrule
    Average & \textbf{86.25\%} & 67.05\% & 37.50\% & 32.50\% & 21.25\% \\
    \bottomrule
  \end{tabular}

  }
\label{tab:tab1}
\vspace{-1mm}
\end{table}

\subsection{Experiment 1: Motion Controller}\label{sec:motion-controller-exprs}
First, we run experiments to evaluate the consistency of our motion controller. Specifically, we pick \emph{4} important motions in our main experiments and run the motion controller \emph{20} times each. The picked motions are:
\begin{enumerate}[
    topsep=0pt,
    noitemsep,
    partopsep=1ex,
    leftmargin=*,
    ]
    \item \emph{Opening door.} A common motion that opens the cabinet door, specifically from \textit{Cabinet-Closed}.
    \item \emph{Removing bar.} A task where a bar should be removed from handles, unblocking the door. This task is from \textit{Cabinet-Blocked}, 
    \item \emph{Pulling lever.} This task is from \textit{Cabinet-Locked}. The robot needs to pull the lever which unlocks a door.
    \item \emph{Removing kettle.} This task is from \textit{Kitchen-Explore}, where the robot needs to remove the kettle that is obstructing the microwave door.
\end{enumerate}

\begin{center}
\begin{tabular}{ |c|c| } 
 \hline
 \emph{Motion} & \emph{Success rate} \\ 
 \hline
 Opening door & 100\% \\ 
 Removing bar & 100\% \\ 
 Pulling lever & 100\% \\ 
 Removing kettle & 80\% \\ 
 \hline
\end{tabular}
\end{center}

It is worth noting that these tests are carried out in an ideal setting, hence the success rates are overly optimistic. During actual planning, multiple facts can affect the motion success rate. For example, the motion planner could generate unnecessary reward functions; and previous steps could change the locations or poses of the objects to interact.

\subsection{Experiment 2: Long-horizon task completion}
\begin{figure}[!t]
     \centering   
     \includegraphics[width=\linewidth]{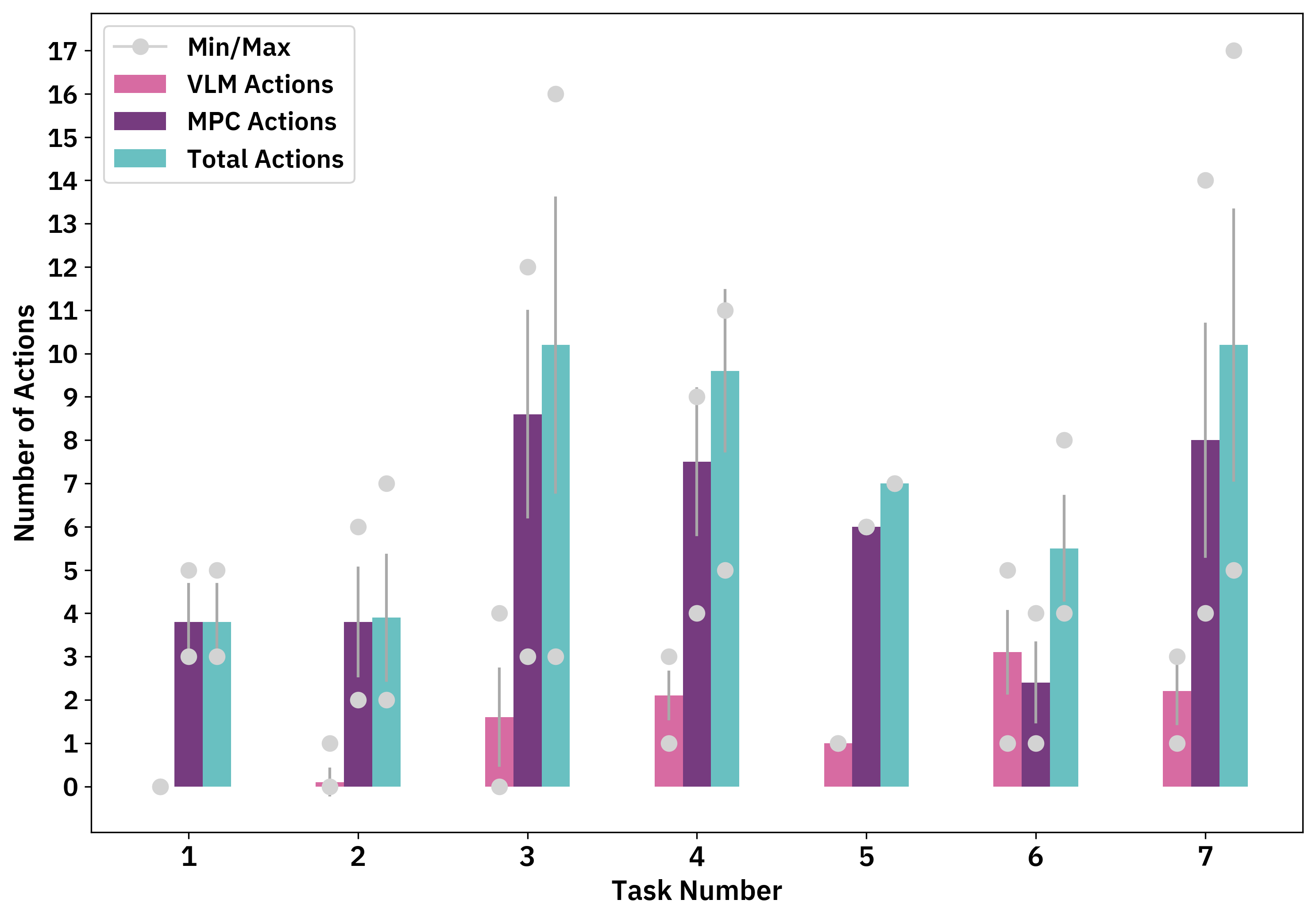}
     \vspace{-3mm}
    \caption{Number of actions the robot executed in each task averaged over ten runs. Actions requiring the Perceiver are shown in pink while those executed using MPC are shown in purple. Standard deviations are shown using gray bars while the minimum and maximum number of actions are shown using gray dots. 
}
    \label{fig:num_actions}
\end{figure}

We report the success rates of all models on their ability to complete our benchmark tasks in Table~\ref{tab:tab1}. We report the percentage of successful runs (ten runs were done for each task). The number of actions performed by the robot in each task is shown in Figure~\ref{fig:num_actions}. On average, most tasks require 7 to 11 steps, with some runs using up to 17 (all without human intervention). 



Overall, \ourmodel achieves a 4$\times$ improvement over Language to Rewards. From the eight tasks, Language to Rewards only shows non-negligible success rates in \textit{Cabinet-Open}, which is a single-motion task, and \textit{Cubes-Color}, where the algorithm can guess the correct cube to move, resulting in a 50\% success rate. Meanwhile, \ourmodel achieves at least 60\% completion rate in all tested tasks, with a close to 90\% average completion rate. \ourmodel performs the worst in \textit{Cabinet-Blocked}, which we conjecture to be a result of both the requirement of precise bar manipulation in tight spaces and the fact that the target block can get stuck in irrecoverable positions. 
Furthermore, we find that PDDL and naive GPT-4V are unable to solve these tasks out-of-the-box. PDDL requires human-provided ground truth information (for example, that the door is blocked, see Appendix \ref{sec:pddl_exps}). PDDLStream is able to solve the problem eventually, but in general, it requires a much longer time. When we prompt GPT-4V naively to create a plan for solving some tasks, it is unable to naturally identify obstacles or object states in the scene (Appendix ~\ref{sec:gpt_4v_baseline}). This indicates that constrained prompting is required for VLMs to provide useful feedback.

\begin{figure*}[th]
\centering  
 \begin{minipage}{0.8\textwidth}
      \centering  
      \vspace{-3mm}
      \includegraphics[width=1.0
     \textwidth, trim={0 5cm 0 5cm}, clip]{figures/kitchen_rollout-compressed.pdf}
     \vspace{-5mm}
       \end{minipage}%
   \begin{minipage}{0.2\textwidth}
     \caption{\textbf{Rollout of robot solving\textit{Kitchen-Explore}}. The high-level plan is shown in the top row. The second row shows each subtask and the corresponding reward functions generated by the Low-Level Planner, as well as Perceiver feedback. If the subtask fails, its box is colored in red. If the plan is completed and the goal is achieved, its box is green. }
     \label{fig:kitchen_rollout}
       \end{minipage}
 \end{figure*}
 
To evaluate the significance of the Verifier, Perceiver, and Replan modules, we notice that \ourmodel achieves +18.8\%, +48.8\%, and +53.8\% improvement compared with removing the three modules, respectively. We also notice that removing each of the modules impacts the performance more in more complex tasks. For example, in \textit{Kitchen-Explore}, without the Perceiver, the agent is unable to figure out that a kettle is blocking the microwave and is, therefore, unable to open the microwave to complete the task. In \textit{Cubes-Blocked}, the agent is unable to see that a cube is already blocking the crate if it does not have access to a Perceiver, so it cannot successfully place the cube within the crate. 

Noticeably, even in simpler tasks such as \textit{Cabinet-\{Open, Closed\}}, \ourmodel still demonstrates superior performances compared to other tested variants even though replanning is not necessary for the ideal scenario. This shows \ourmodel's ability to act as a safe-net mechanism for unexpected errors through using a Verifier, improving the overall consistency.

We show an example rollout of the robot completing \textit{Kitchen-Explore} in Figure~\ref{fig:kitchen_rollout}. For \textit{Kitchen-Explore}, the High-Level Planner first instructs the robot to look inside the cabinet. The Perceiver informs the Planner that the apple is not in the cabinet. The Planner then instructs the robots to look inside the microwave. However, the motion control fails. The Perceiver is again called and informs the Planner that there is a kettle in front of the microwave. The Planner replans, instructing the robot to first move the kettle out of the way. The robot then resumes with the remaining subtasks. It opens the microwave and the Perceiver informs the Planner that the green apple is there and the task is complete. For another example rollout of \textit{Cabinet-Blocked}, the reader can refer to Appendix Figure~\ref{fig:cabinet_rollout}.

\subsection{\ourmodel with Real-Robot Environment}
\begin{figure}[!t]
     \centering         \includegraphics[width=\linewidth]{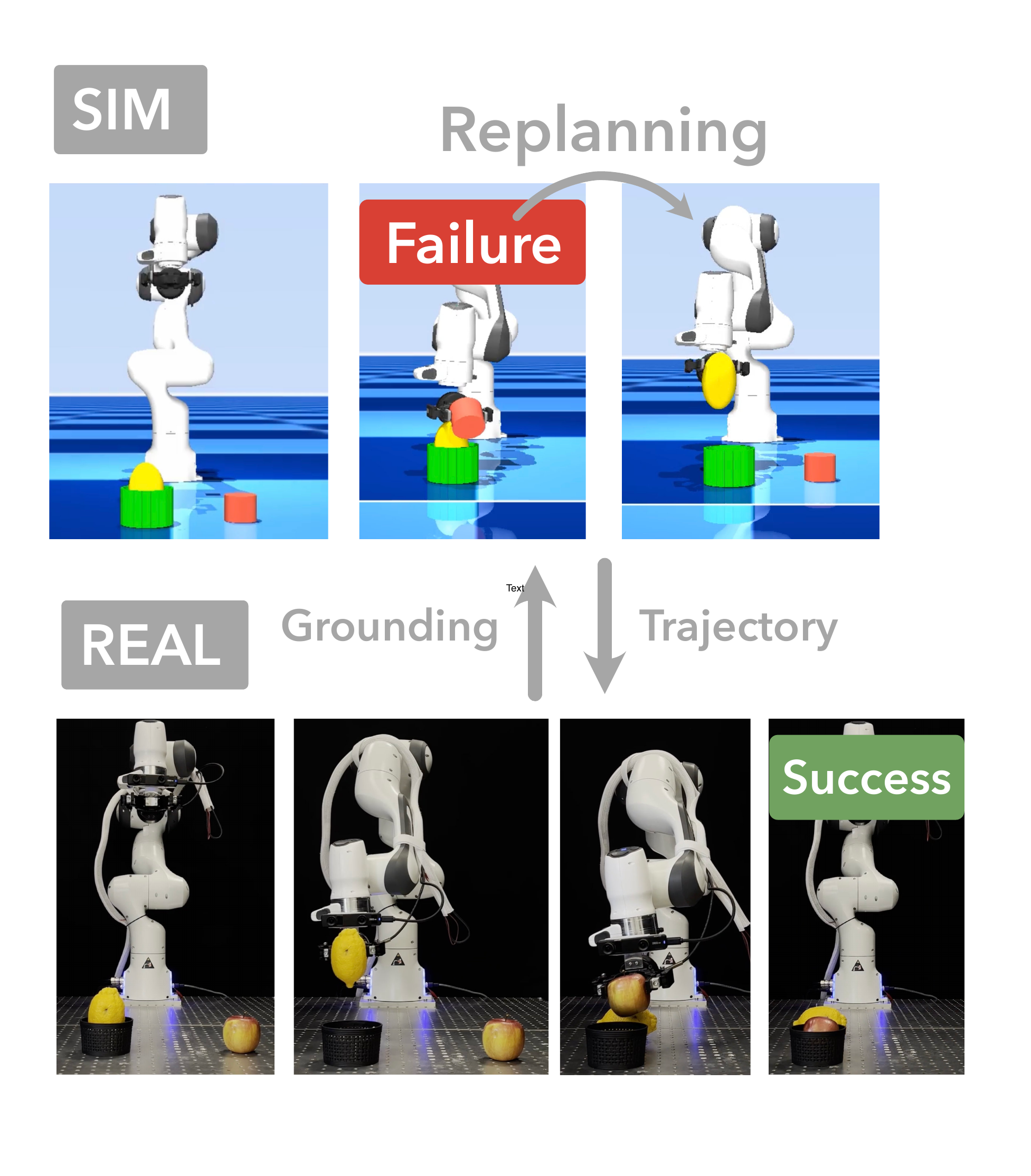}
    \caption{\textbf{Real-world experiment}. The robot is tasked with placing an apple inside the bowl, but it has to figure out that the lemon must first be removed in order to complete the task. The full experiment trajectory can be seen in the video in Supplementary Material.
 }
    \label{fig:realworld}
\end{figure}

For real robot evaluation, we adapt \textit{Cubes-Blocked} by replacing the colored cubes with a lemon and an apple (Figure~\ref{fig:realworld}). 

\paragraph{Pose estimation} In a real robot setting, we use  GroundingDINO~\citep{liu2023grounding} to detect object bounding boxes and Non-Maximum Suppression~\citep{zou2023object} to remove redundant ones. We then use the Segment Anything Model (SAM)~\citep{kirillov2023segment} to generate object segmentations. Using depth information from a ZED camera\footnote{https://www.stereolabs.com/products/zed-2}, a 3D point cloud for each segmented object is generated. Object poses are estimated by fitting the smallest bounding box to each point cloud. This perception setup builds on a recent Language guided model used in  ORGANA~\citep{darvish2024organa}. 

\paragraph{Planning and replanning} For long-horizon planning, we follow the same \ourmodel's pipeline as in simulated environments, with the exception that the real-world images are used for the Perceiver module. During the MPC low-level planning stage, we first run the simulation with the estimated object poses from our perception module. If the simulation ends in a failure state, we return to \ourmodel to replan with the updated real-world images for diagnosis; if the simulation succeeds, we then execute the simulated trajectory on our real robot. Additionally, we have found that MPC often generates actions that would not be deemed safe in a lab. To compensate this, we have incorporated supplementary terms into our residuals to comply with safety constraints.

\section{Limitations}
\begin{figure}[!t]
     \centering  
     \includegraphics[width=\linewidth]{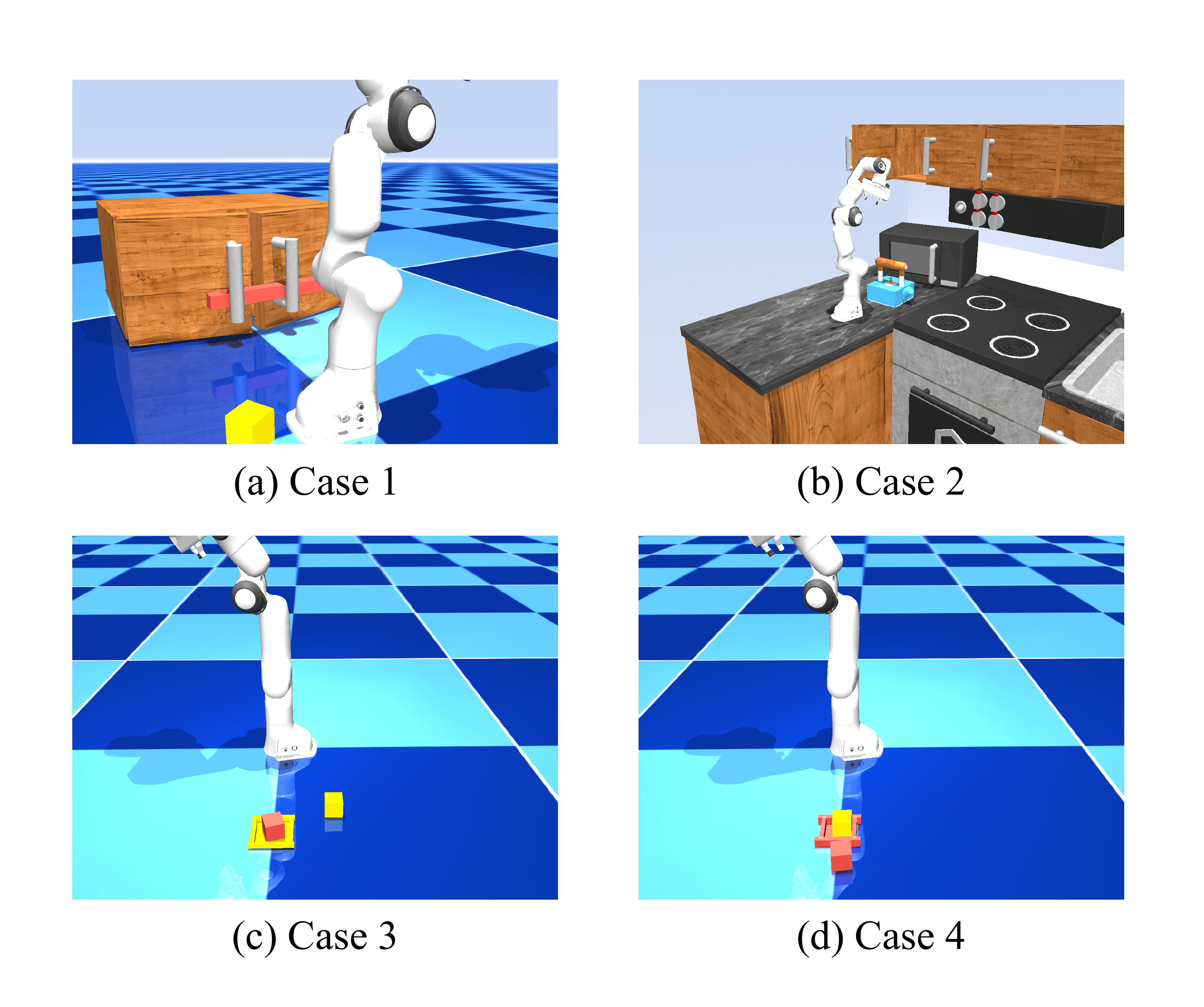}
    \caption{
    \textbf{Examples of error cases} that occurred in \ourmodel include a) failure to remove the bar due to incorrect reward generation (CabinetBlocked); b) failure to identify the reason for the inability to open the microwave door (KitchenExplore); c) vague VLM question about object attributes resulting in incorrect cube selection (CubesColor); d) failure to identify why the robot could not place the red cube in the crate (CubesBlocked).
    }
    \label{fig:error-cases}
\end{figure}

We note one limitation of our method is its reliance on VLM's understanding and interpretation of spatial states. If the VLM cannot accurately recognize an object or interpret the reason why a task is incomplete, it may lead to the inaccurate processing of the robotic task. This will likely be improved as VLMs become trained to directly predict spatial states, such as SpatialVLM ~\citep{chen2024spatialvlm}. In the meantime, we find that by using an LLM to probe specific questions, sampling multiple answers, and generating a summary consistent with the LLM's knowledge of the scene, \ourmodel is able to utilize the VLM to accurately reason about the scene state.
We have also identified some  failure points:
\textbf{Case 1: LLM-MPC communication failure (Figure~\ref{fig:error-cases}(a))} In our \textit{Cabinet-Close} environment robot tried to open the cabinet door but failed. The Perceiver gave a correct diagnosis to remove the bar from the handle. However, when generating the reward functions to remove the bar, the LLM selected the wrong primary reward function, as demonstrated below:
\begin{lstlisting}[language=Python]
reset_reward()
minimize_l2_distance_reward("palm", "red_block_right_side", primary_reward=True)
maximize_l2_distance_reward("red_block_right_side", "target_position_in_cabinet")
execute_plan()
\end{lstlisting}
\texttt{maximize_l2_distance_reward} should be the primary function. As a result, MPC ends prematurely. The robot will not be able to remove the bar. Another case of this is failure attributed to the controller could be incorrectly interpreted as the task being undoable.This misinterpretation can lead to unnecessary task abandonment or repeated attempts, resulting in a waste of time and resources.
\textbf{Case 2: LLM consistency failure (Figure~\ref{fig:error-cases}(b))} The robot is blocked by a kettle obstructing the path. The Perceiver gives five diagnoses, of which three claimed that the kettle was blocking the way, one claimed the cabinet door was blocking the way, and one did not give any conclusive diagnosis. The summary LLM concludes that the cabinet door blocked the action. The robot interacts with the cabinet and never removes the kettle.
\textbf{Case 3: LLM-VLM communication failure (Figure~\ref{fig:error-cases}(c))} The high-level planner proposed a plan where the first step is \emph{``Identify the cube with the same colour as the crate''} using the VLM. However, the LLM queries the VLM with \emph{``Which cube has the same colour?''}, which is too vague. This results in the VLM answering \emph{``The same color cube is the yellow cube and the yellow cube in the middle of the blue cube group.''}. This answer did not provide the necessary information to solve the task. The robot put the wrong cube on the crate. \textbf{Case 4: preciever failure (Figure~\ref{fig:error-cases}(d))} After the robot's failure to execute the task ``Place the red cube on the crate", the Perceiver was called to help identify any issues. The Perceiver's diagnoses all mentioned that the robot was holding the red cube but did not identify the yellow cube as blocking the crate, and so the Planner's summary of the VLMs diagnoses was: ``Based on the given information, the most probable reason why the robot cannot place the red\_cube on the crate is that it is currently holding the red cube." However, it's also important to note that \textit{Cubes-Blocked} used GPT-4V which severely limits the number of ouput tokens from the model, and so a lot of explanations were cut off (for example: ``In the image provided, the robot is holding the red cube, which is currently"). 
Additional details in Appendix~\ref{sec:error_analysis}.

\section{Discussion and Conclusions}
In Section \ref{sec:no_vlm_diagnosis}, we demonstrate that only providing coordinates and bounding boxes is insufficient for the LLM to correctly reason about the error. We also find that out-of-the-box VLMs are unable to diagnose errors in one shot either, as shown by our experiments in Appendix \ref{sec:gpt_4v_baseline}. The perceiver is important for diagnosing errors.
However, we also found that we were able to make MPC reward generation more robust by using a Verifier to eliminate any reward functions that were not essential to completing the task. In essence, we found that using a Verifier at multiple stages in our workflow was essential in improving the success rate of long-horizon task execution.


In summary, our paper introduces \ourmodel, a robust solution for multi-stage planning, utilizing the power of LLMs for plan generation and VLMs for insightful feedback. Our multi-level planning approach, coupled with step-wise verification and replanning, demonstrates promising results in addressing multi-stage tasks. 

\section{Acknowledgements}
We would like to thank Miroslav Bogdanovic for insightful discussions and Yuchi Zhao for assistance with real robot setup. This research was undertaken thanks in part to funding provided to the University of Toronto's Acceleration Consortium from the Canada First Research Excellence Fund Grant number - CFREF-2022-00042.

\renewcommand*{\bibfont}{\small}

\bibliographystyle{plainnat}
\bibliography{replan}

\newpage

\begin{appendices}
  \crefalias{section}{appendix}
  \crefalias{subsection}{appendix}
  \crefalias{subsubsection}{appendix}
  \onecolumn
  {
    \LARGE
    \textbf{Appendices} \par
  }
  \input{appendix}

\end{appendices}


\end{document}

%% file: 0-preamble.tex
\usepackage{etoolbox} 
\usepackage{times}
\usepackage[utf8]{inputenc} 
\usepackage[T1]{fontenc}    
\usepackage[english]{babel} 
\usepackage[protrusion=true,expansion=true]{microtype}
\usepackage{comment}
\usepackage{cancel}
\usepackage{csquotes}
\usepackage{listings}
\usepackage{nameref}
\usepackage{paralist}
\usepackage{xspace}
\usepackage{setspace}
\usepackage{makeidx}
\usepackage{pdfpages}
\usepackage{chngcntr}

\usepackage{amsmath,amssymb} 
\usepackage{amsfonts}       
\usepackage{mathtools}
\usepackage{array} 
\usepackage{nicefrac}       
\usepackage{breqn}          
\usepackage{textcomp}
\usepackage{bm} 
\usepackage{pifont}
\usepackage[
  separate-uncertainty = true,
  multi-part-units = repeat
]{siunitx}

\usepackage{graphicx}
\usepackage{adjustbox} 
\usepackage{epsfig}

\usepackage{float}
\usepackage[font=small]{subcaption}
\usepackage[font=small,labelfont=bf]{caption}
\usepackage{lscape}      
\usepackage{wrapfig}

\usepackage{tikz}  
\usepackage{makecell}
\usepackage{overpic}
\usepackage{algorithm}
\usepackage[noend]{algpseudocode}

\usepackage{array} 
\usepackage{tabularx}
\usepackage{multirow}
\usepackage{multicol}
\usepackage{booktabs}   

\usepackage{paralist}
\usepackage{enumitem}
\setlist[itemize]{noitemsep,leftmargin=*,topsep=0in}
\setlist[enumerate]{noitemsep,leftmargin=*,topsep=0in}

\usepackage{url}            
\PassOptionsToPackage{table}{xcolor}
\usepackage{color,colortbl} 
\usepackage{soul}

\PassOptionsToPackage{capitalize, noabbrev}{cleveref}
\usepackage[pagebackref=false,breaklinks=true,colorlinks,urlcolor=blue,citecolor=blue,linkcolor=blue,bookmarks=true]{hyperref}

\usepackage[capitalize,noabbrev]{cleveref}
\makeatletter
\let\NAT@parse\undefined
\makeatother
\usepackage[numbers,sort&compress]{natbib}

\usepackage{blindtext}
\usepackage{bbm}

\usepackage{lipsum}

\newcommand{\includegraphicsmaybe}[1]{\IfFileExists{#1}{\includegraphics{#1}}{\includegraphics{dummy.pdf}}}

\usepackage{titlesec}
\titlespacing{\section}{0pt}{0.3\baselineskip}{0.25\baselineskip}
\titlespacing{\subsection}{0pt}{0.25\baselineskip}{0.15\baselineskip}
\titlespacing{\subsubsection}{0pt}{0.05\baselineskip}{0.03\baselineskip}

\renewcommand{\paragraph}[1]{\vspace{0.2em}\noindent\textit{#1} --}

\makeatletter
\newcolumntype{B}[3]{>{\boldmath\DC@{#1}{#2}{#3}}c<{\DC@end}}
\makeatother


%% file: 0-user-macros.tex
\algnewcommand{\IfThenElse}[3]{
  \State \algorithmicif\ #1\ \algorithmicthen\ #2\ \algorithmicelse\ #3}
\algnewcommand{\IfThen}[2]{
  \State \algorithmicif\ #1\ \algorithmicthen\ #2}
\algrenewcommand\algorithmicrequire{\textbf{Input:}}
\algrenewcommand\algorithmicensure{\textbf{Output:}}

\algrenewcommand\alglinenumber[1]{\tiny #1:}

\usepackage{listings}
\def\code#1{\texttt{#1}}
\definecolor{codegreen}{rgb}{0,0.6,0}
\definecolor{codegray}{rgb}{0.4,0.4,0.4}
\definecolor{codepurple}{rgb}{0.5,0,0.9}
\definecolor{backcolour}{rgb}{0.95,0.95,0.95}

\definecolor{lightgray}{rgb}{0.9,0.9,0.9}
\definecolor{lightpink}{rgb}{0.98,0.85,0.86}
\definecolor{lightblue}{rgb}{0.68,0.84,0.9}

\lstdefinestyle{mystyle}{
    backgroundcolor=\color{backcolour},   
    commentstyle=\color{codegreen},
    keywordstyle=\color{magenta},
    numberstyle=\tiny\color{codegray},
    stringstyle=\color{codepurple},
    basicstyle=\fontsize{7.5}{8}\selectfont\ttfamily\ttfamily,
    breakatwhitespace=false,         
    breaklines=true,
    breakindent=0pt,
    captionpos=b,                    
    keepspaces=true,                 
    numbers=none,                    
    numbersep=5pt,                  
    showspaces=false,                
    showstringspaces=false,
    showtabs=false,                  
    tabsize=2
}
\newcommand{\specialcell}[2][c]{%
\begin{tabular}[#1]{@{}c@{}}#2\end{tabular}}



%% file: math_commands.tex
\usepackage{amsmath,amsfonts,bm}

\def\eqref#1{equation~\ref{#1}}

\def\1{\bm{1}}

\DeclareMathAlphabet{\mathsfit}{\encodingdefault}{\sfdefault}{m}{sl}
\SetMathAlphabet{\mathsfit}{bold}{\encodingdefault}{\sfdefault}{bx}{n}



%% file: appendix.tex
\counterwithin{figure}{section}
\section{Environment Details}
\label{sec:env-details}

\subsection{Wooden cabinet scene} \label{sec:env_1} A room where there is a yellow cube placed on the floor beside a wooden cabinet. There is a red bar holding the handles of the wooden cabinet closed. The doors of the cabinet cannot be opened without removing the bar. We implemented three tasks in this scene:
 
\paragraph{\texttt{CabinetOpen}: Place the yellow cube in the wooden cabinet (easy mode).} The robot must pick up the yellow cube and place it in the cabinet, which is open. This is a simple 1-step task that evaluates the Low-Level Planner for single motion planning. 

\paragraph{\texttt{CabinetClosed}: Place the yellow cube in the wooden cabinet (hard mode).} This task is the same as the previous task, but now the wooden cabinet doors are closed. This task requires two steps: 1) opening the door and 2) placing the cube inside the cabinet. The High-Level Planner is assessed to plan a sequence of actions and pass them to the Low-Level Planner for motion generation.

\paragraph{\texttt{CabinetBlocked}: Place the yellow cube inside the wooden cabinet (expert mode).}  The challenge with this task is that the robot must identify that it cannot open the wooden cabinet because there is a bar across the handles of the door. After removing the bar, the robot can open the cabinet door and finish the task. This task is challenging because it requires vision to identify that the door cannot be opened, followed by replanning to remove the item blocking the door. 

\subsection{Kitchen environment scene} A kitchen that contains a cabinet with two doors, a microwave, a kettle, and a green apple.

\paragraph{\texttt{KitchenExplore}: Find the green apple.} A green apple was hidden in the microwave is not visible to the robot at the start of the scene. The robot must search for the apple in the kitchen. There is an additional challenge where the kettle is blocking the microwave door from being opened, and so to open the door, the robot must first remove the kettle. Same as \texttt{CabinetBlocked}, \texttt{Kitchen-Explore} also requires both vision and replanning to solve the task, but it has an additional challenge because the goal requires open-ended exploration (it is unclear where the apple is), which requires replanning at a high level.

\subsection{Wooden cabinet and lever scene} 
A room containing a wooden cabinet, a blue block, and a lever that controls whether the cabinet door is locked or unlocked.

\paragraph{\texttt{CabinetLocked}: Remove the blue cube from the cabinet.} Just as with tasks 1-3, this task requires the robot to open a cabinet. There is no physical obstruction preventing the cabinet from being opened; however, the cabinet is locked. The cabinet becomes unlocked once a lever close to the cabinet is pulled. Thus, after (unsuccessfully) trying to open the cabinet door, the robot must reason that it should pull the lever first and then open the door.

\subsection{Coloured cubes scene} \label{sec:env_4}
A room containing a small red crate and two cubes (one is yellow and the other is red).

\paragraph{\texttt{CubesColor}: Place the cube with the same colour as the crate on the crate.} In this task, the robot has to identify the cube with the same colour as the crate and place it on the crate.

\paragraph{\texttt{CubesBlocked}: Blocking cube.} The robot is given the colour of a cube it must put on the crate. However, there is already a cube on the crate with a different colour and the crate can only hold one cube at a time. The robot must remove the cube that is already on the crate before placing the target one.

\paragraph{\texttt{CompositeExplore}: Weight sensor} 
The robot is asked to open a sliding cabinet. The cabinet is locked by a weight sensor, which can be activated by putting a weight on top. To do this, the robot needs first to explore the scene to find the weight.

The instructions we use for each task are listed below:

\begin{center}
\begin{tabular}{ |l|p{10cm}| } 
 \hline
\label{tab:TaskDescription}
\emph{Environment} & \emph{Instruction} \\ 
 \hline
 \texttt{Cabinet-\{Open, Closed, Blocked\}} & \texttt{move the yellow\_cube to target\_position inside the wooden\_cabinet} \\ 
 \texttt{Kitchen-Explore} & \texttt{find the green\_cube} \\ 
 \texttt{Cabinet-Locked} & \texttt{find the blue\_cube} \\ 
 \texttt{TwoCube-Color} & \texttt{place the cube with the same color as the crate on the crate} \\ 
 \texttt{TwoCube-Blocked} & \texttt{place the red cube on the crate} \\ 
 \texttt{Composite-Explore} & \texttt{open the stone\_cabinet. The weight sensor lock can be unlocked by putting the red\_cube on it.}\\
 \hline
\end{tabular}
\end{center}

\begin{figure}[!p]
     \centering
     \begin{subfigure}[b]{0.3\textwidth}     \centering

     \includegraphics[width=\textwidth]{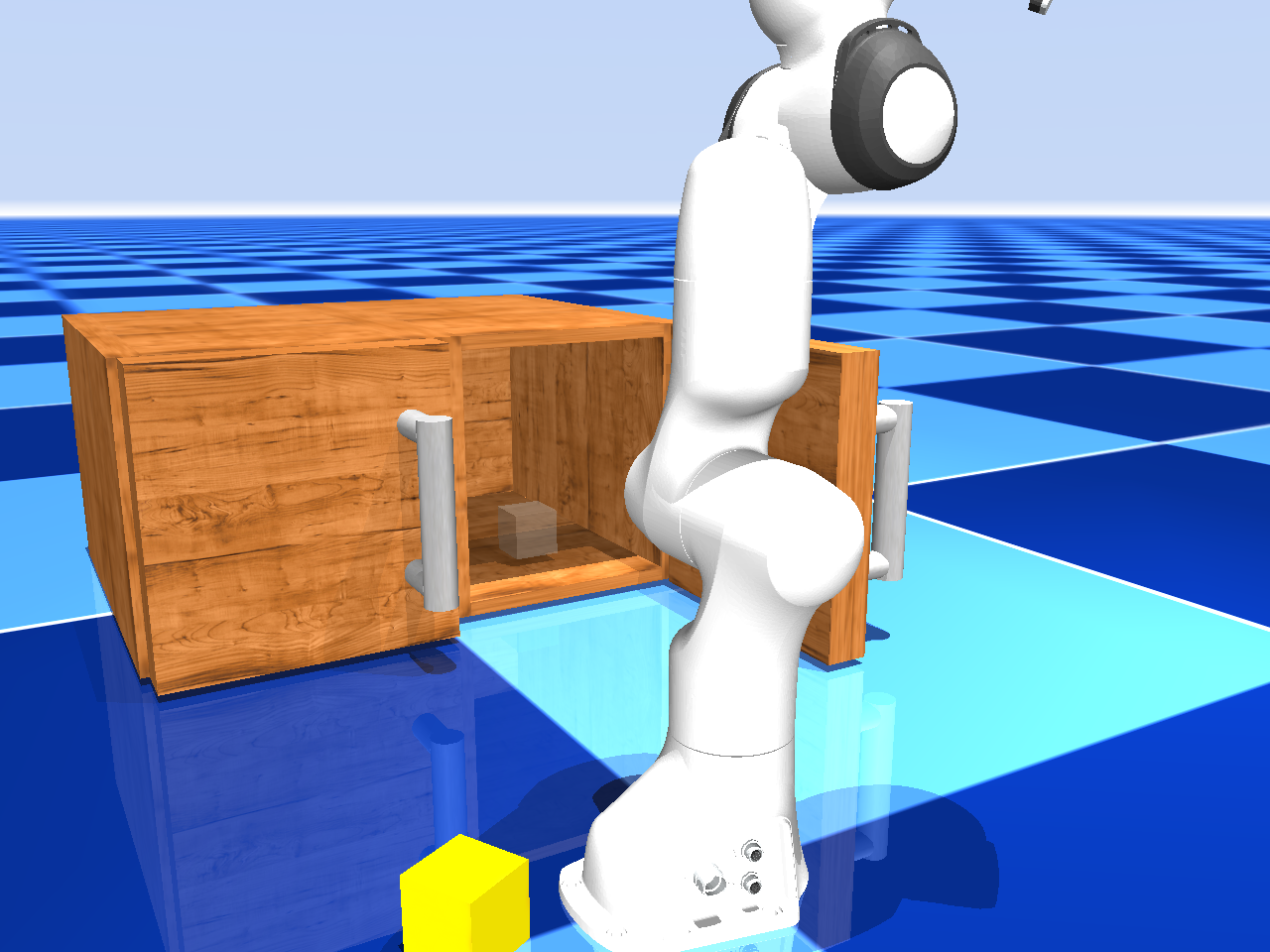}
         \caption{\texttt{Cabinet-Open} - initial}
         \label{fig:cabinet-open-init}
     \end{subfigure}
    \begin{subfigure}[b]{0.3\textwidth}
         \centering

             \includegraphics[width=\textwidth]{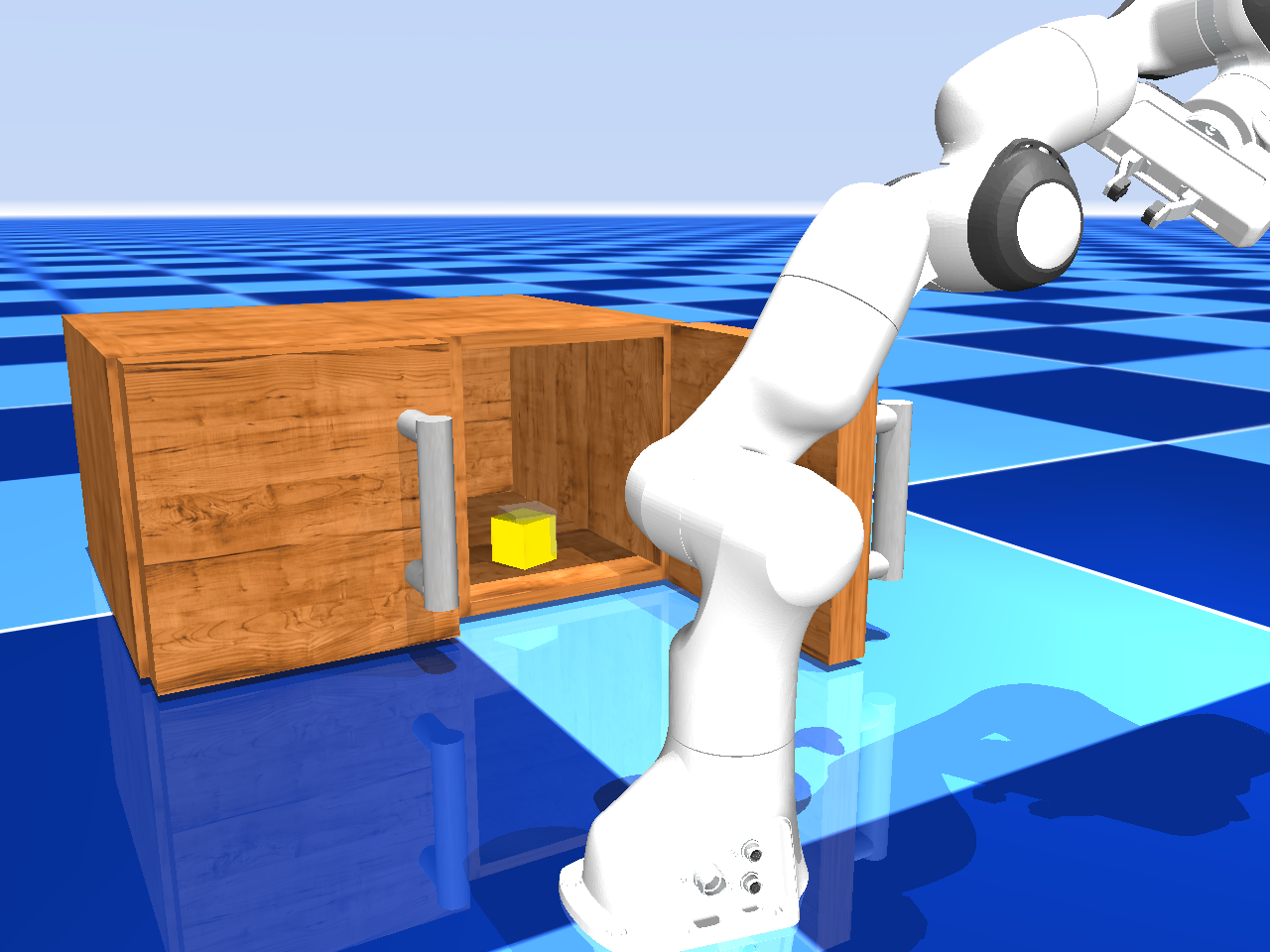}
         \caption{\texttt{Cabinet-Open} - complete}
         \label{fig:cabinet-open-done}
     \end{subfigure}
     \\

     \begin{subfigure}[b]{0.3\textwidth}
           \includegraphics[width=\textwidth]{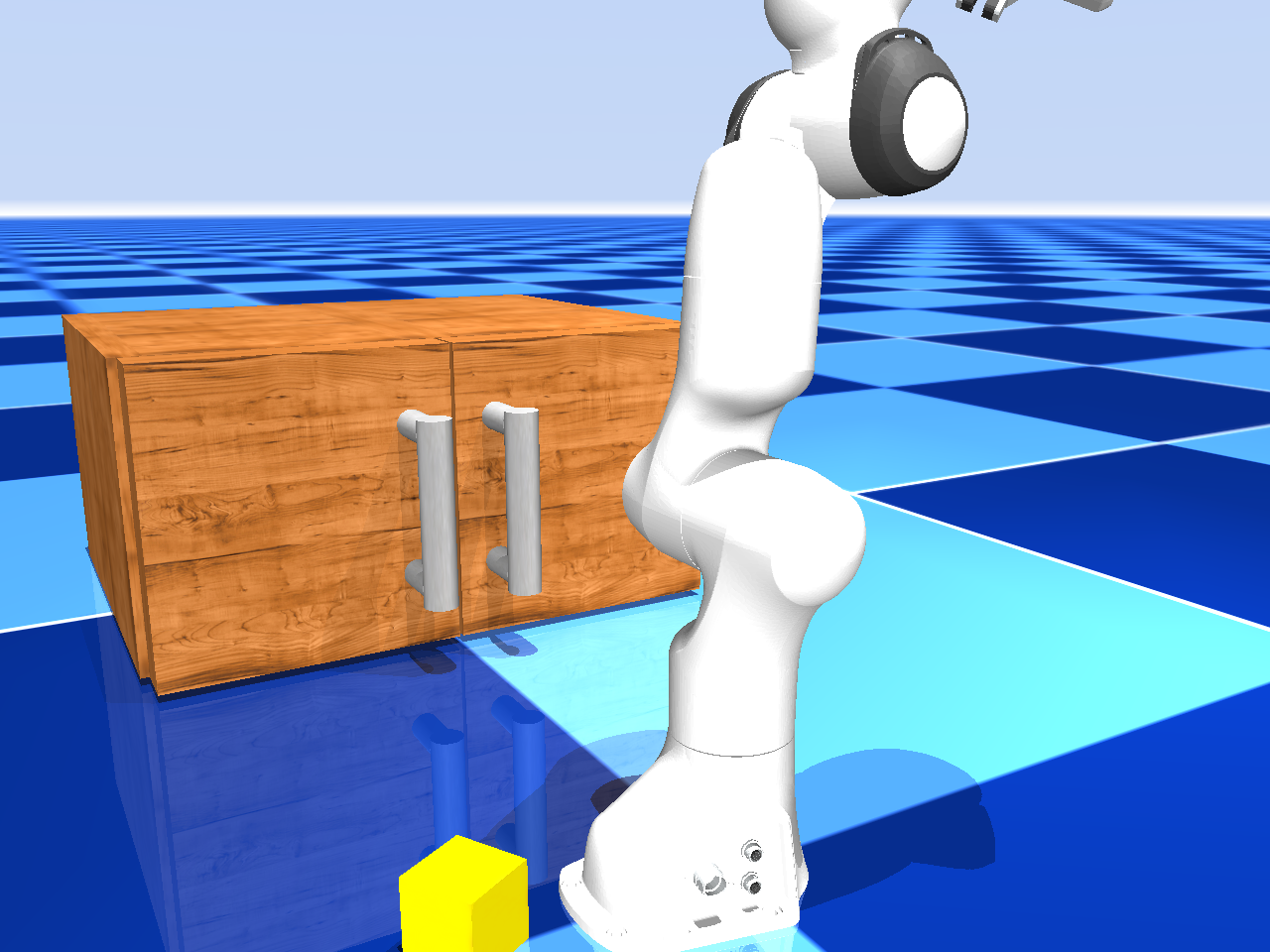}
         \caption{\texttt{Cabinet-Closed} - initial}
         \label{fig:cabinet-nolock-init}
     \end{subfigure}
          \begin{subfigure}[b]{0.3\textwidth}
           \includegraphics[width=\textwidth]{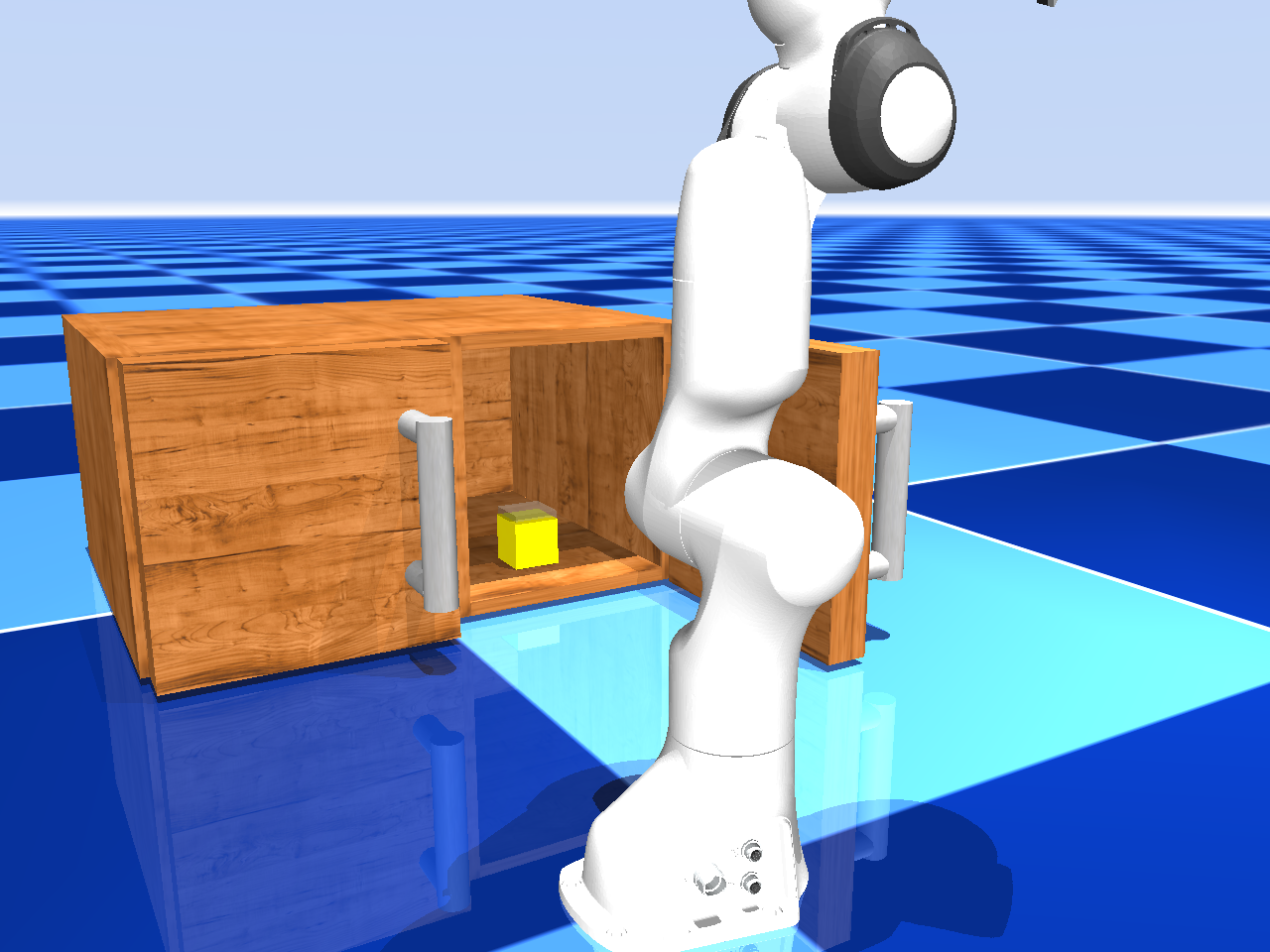}
         \caption{\texttt{Cabinet-Closed} - complete}
         \label{fig:cabinet-nolock-done}
     \end{subfigure}
     \\
     \begin{subfigure}[b]{0.3\textwidth}         \includegraphics[width=\textwidth]{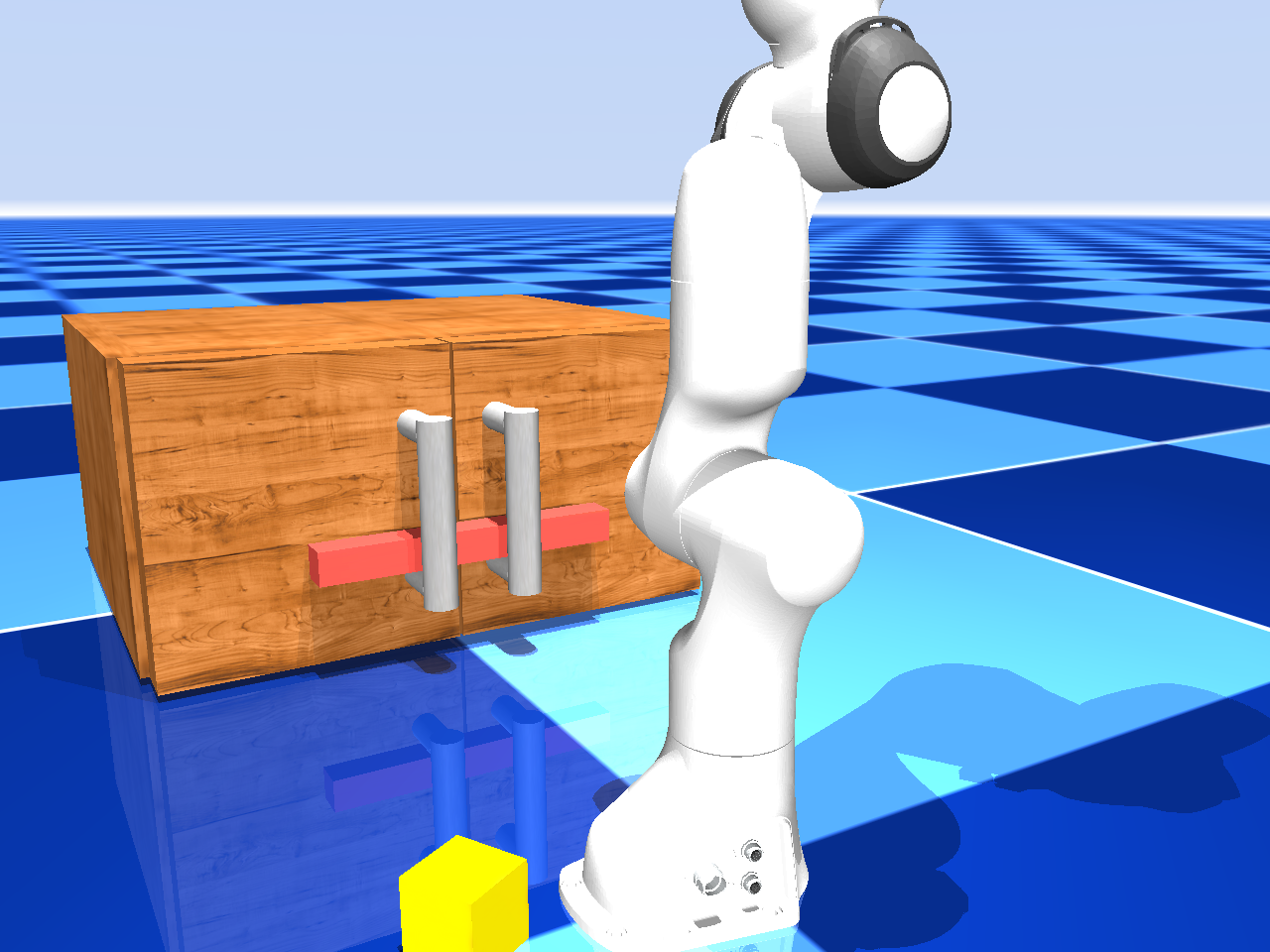}
         \caption{\texttt{Cabinet-Blocked} - initial}
         \label{fig:cabinet-init}
     \end{subfigure}
        \begin{subfigure}[b]{0.3\textwidth}
           \includegraphics[width=\textwidth]{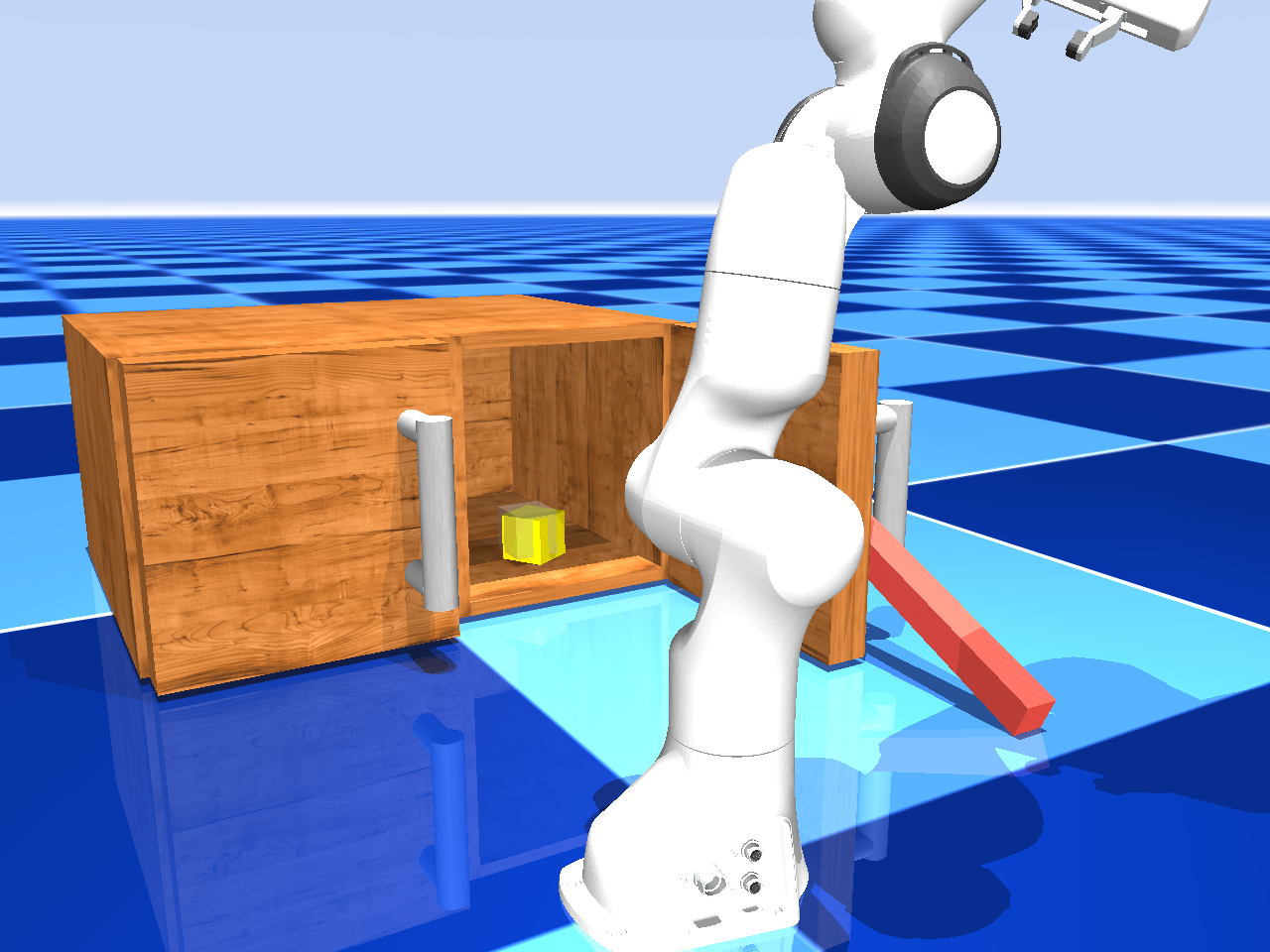}
         \caption{\texttt{Cabinet-Blocked} - complete}
         \label{fig:cabinet-nolock-done}
     \end{subfigure}
     \\
     \begin{subfigure}[b]{0.3\textwidth}
         \centering
         \includegraphics[width=\textwidth]{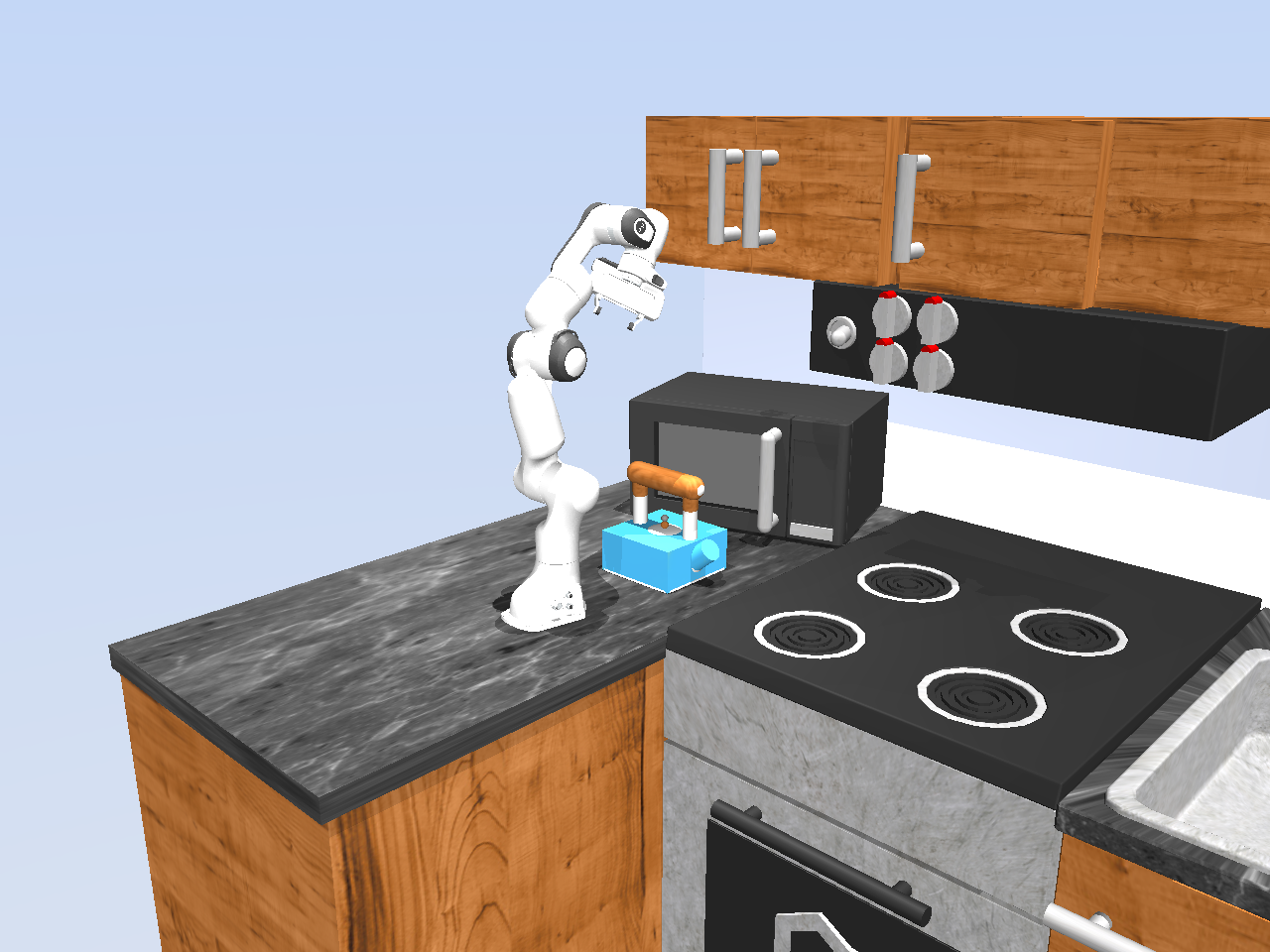}
         \caption{\texttt{Kitchen-Explore} - initial}
         \label{fig:kitchen-init}
     \end{subfigure}
          \begin{subfigure}[b]{0.3\textwidth}
         \centering
         \includegraphics[width=\textwidth]{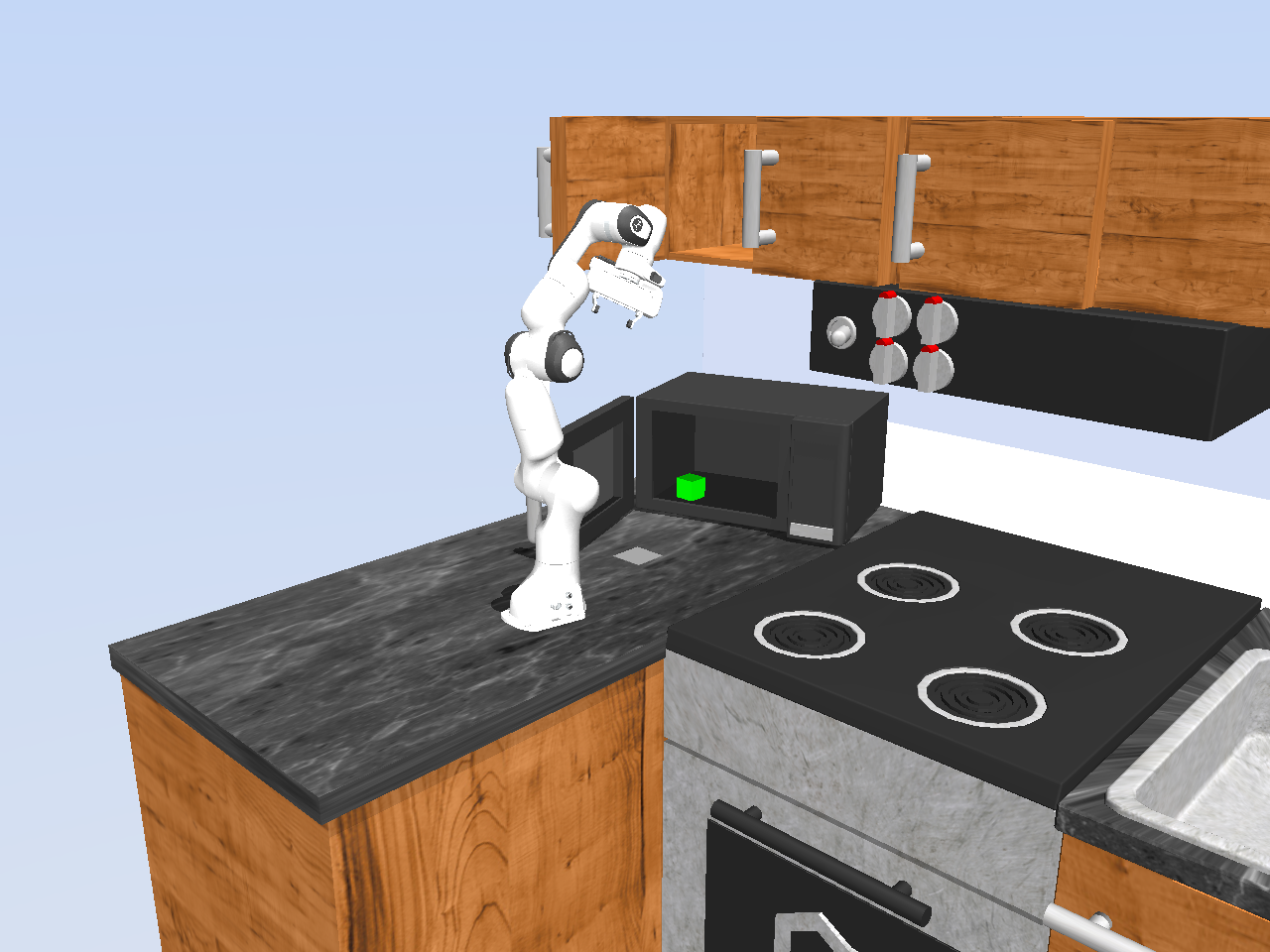}
         \caption{\texttt{Kitchen-Explore} - complete}
         \label{fig:kitchen-done}
     \end{subfigure}
     \\
             \caption{Initial and final scenes of the tested environments.}
        \label{fig:env-init}
\end{figure}
\FloatBarrier
\begin{figure}[!p]
\ContinuedFloat 
    \centering
    \begin{subfigure}[b]{0.3\textwidth}
        \centering
        \includegraphics[width=\textwidth]{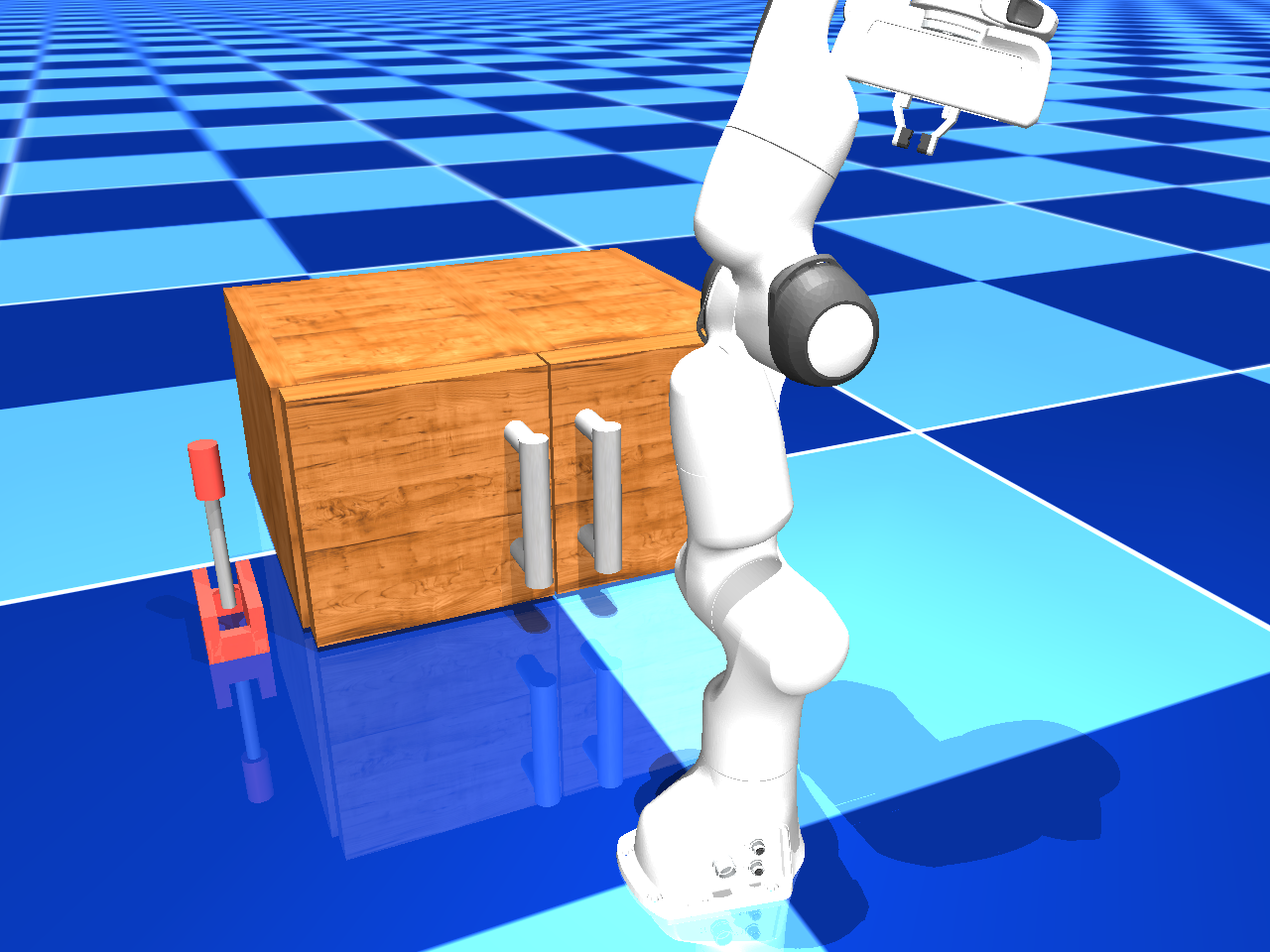}
        \caption{\texttt{Cabinet-Locked} - initial}
        \label{fig:locklock-init}
    \end{subfigure}
    \begin{subfigure}[b]{0.3\textwidth}
         \centering
         \includegraphics[width=\textwidth]{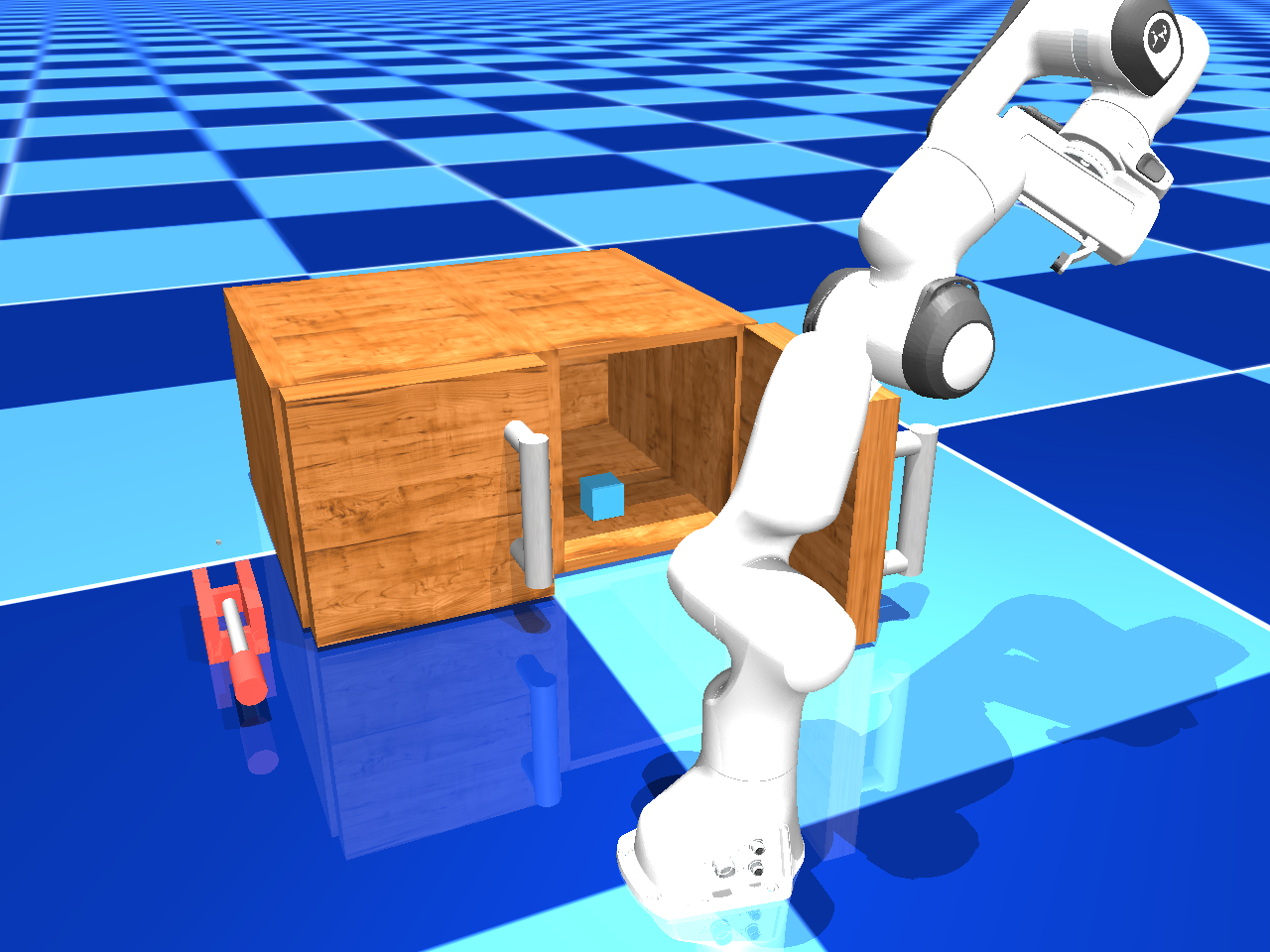}
         \caption{\texttt{Cabinet-Locked} - complete}
         \label{fig:locklock-done}
     \end{subfigure}
     \\
     \begin{subfigure}[b]{0.3\textwidth}
         \centering
         \includegraphics[width=\textwidth]{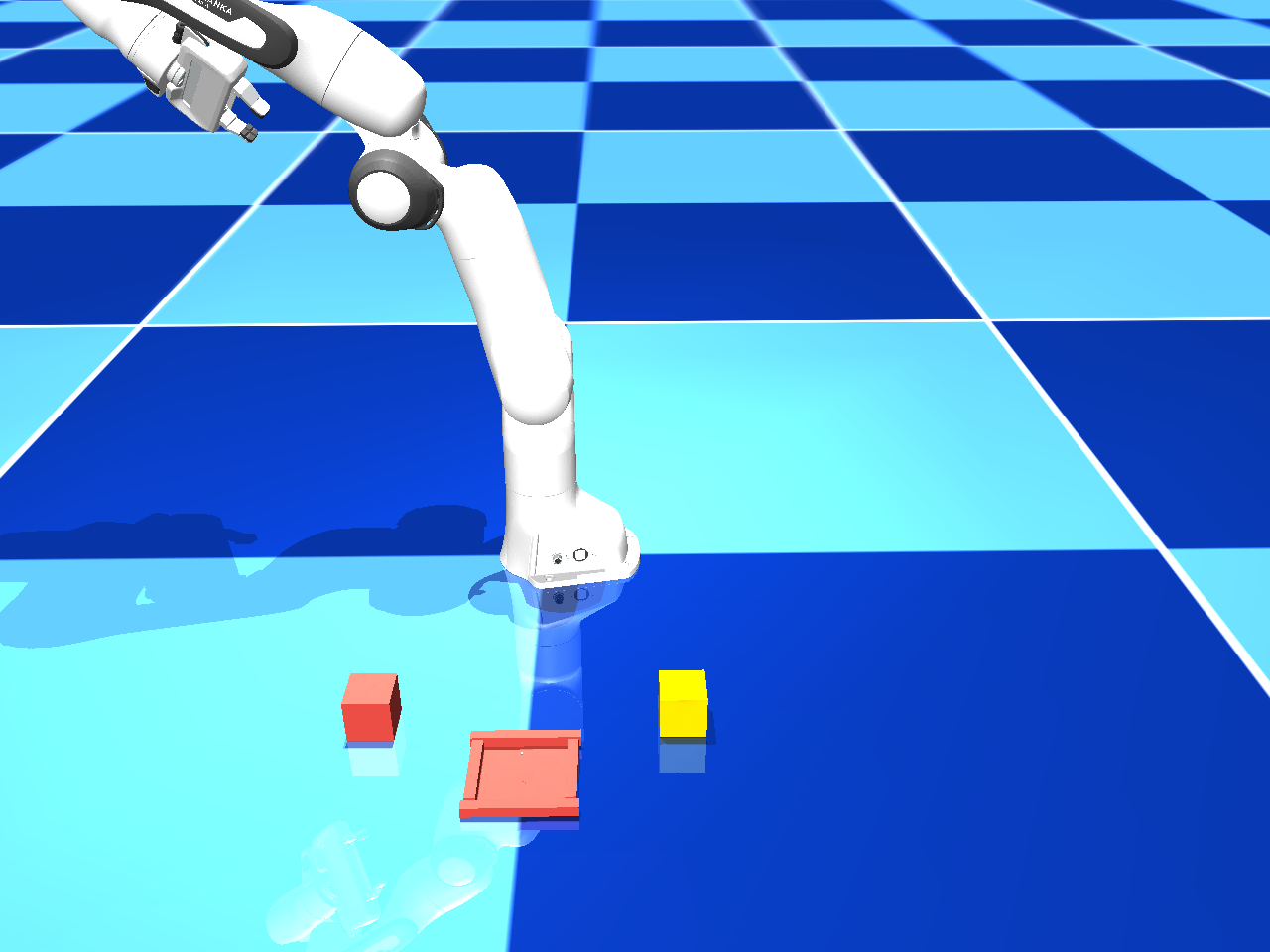}
         \caption{\texttt{TwoCube-Color} - initial}
         \label{fig:blocks-init}
     \end{subfigure}
     \begin{subfigure}[b]{0.3\textwidth}
         \centering
         \includegraphics[width=\textwidth]{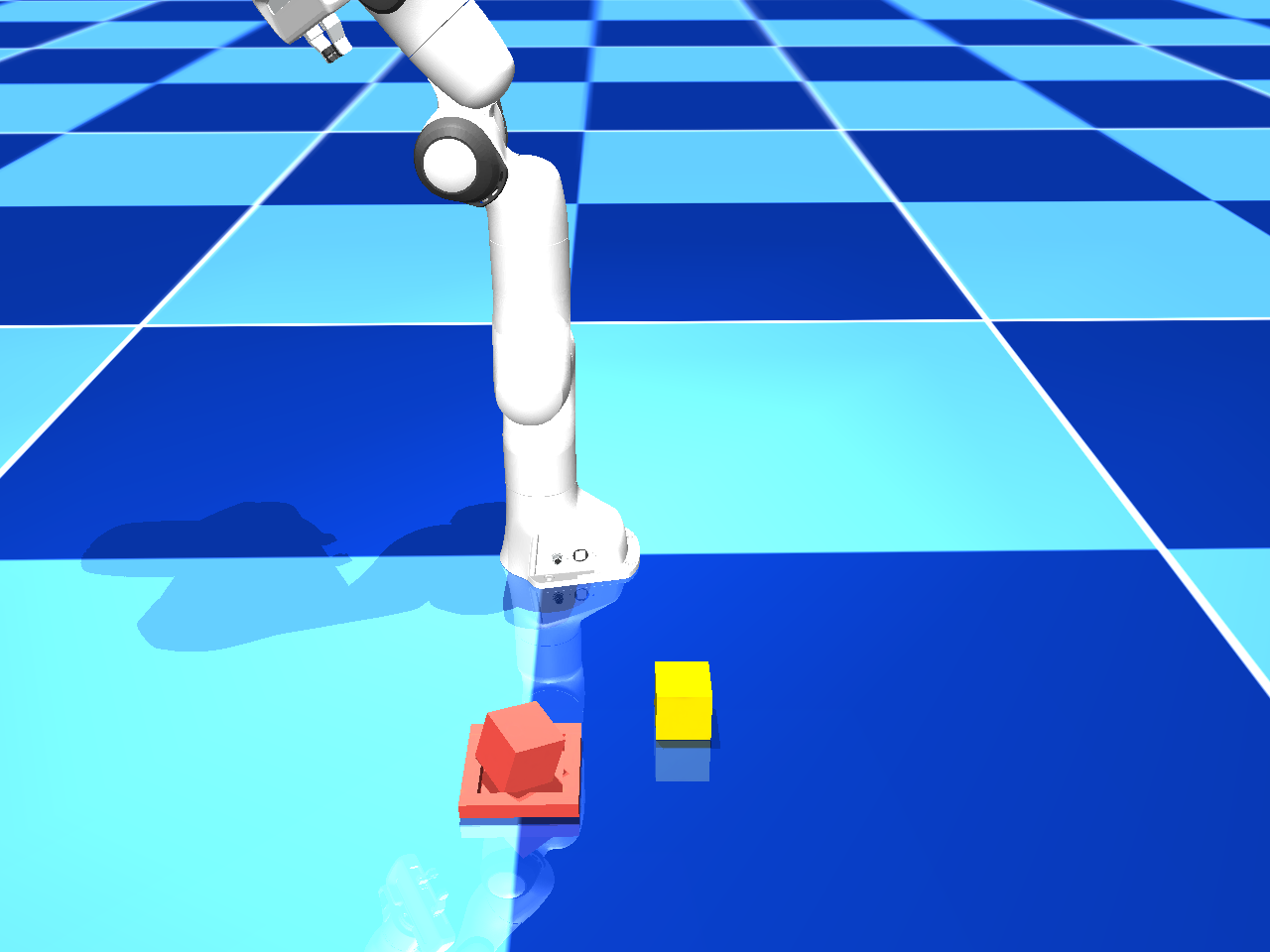}
         \caption{\texttt{TwoCube-Color} - complete}
         \label{fig:blocks-done}
     \end{subfigure}
     \\
    \begin{subfigure}[b]{0.3\textwidth}
         \centering
         \includegraphics[width=\textwidth]{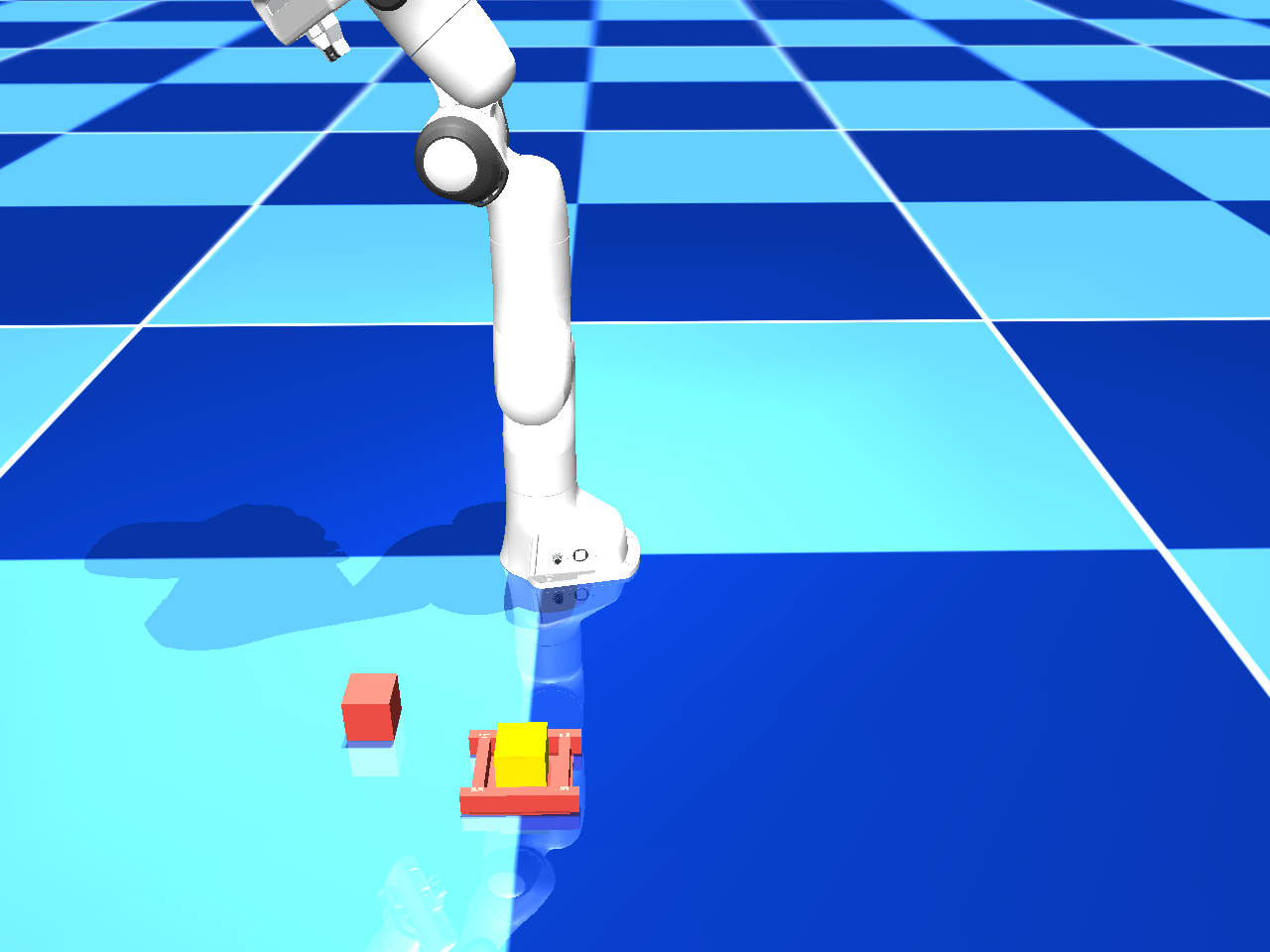}
         \caption{\texttt{TwoCube-Blocked} - initial}
         \label{fig:blockblocks-init}     
    \end{subfigure}
    \begin{subfigure}[b]{0.3\textwidth}
         \centering
         \includegraphics[width=\textwidth]{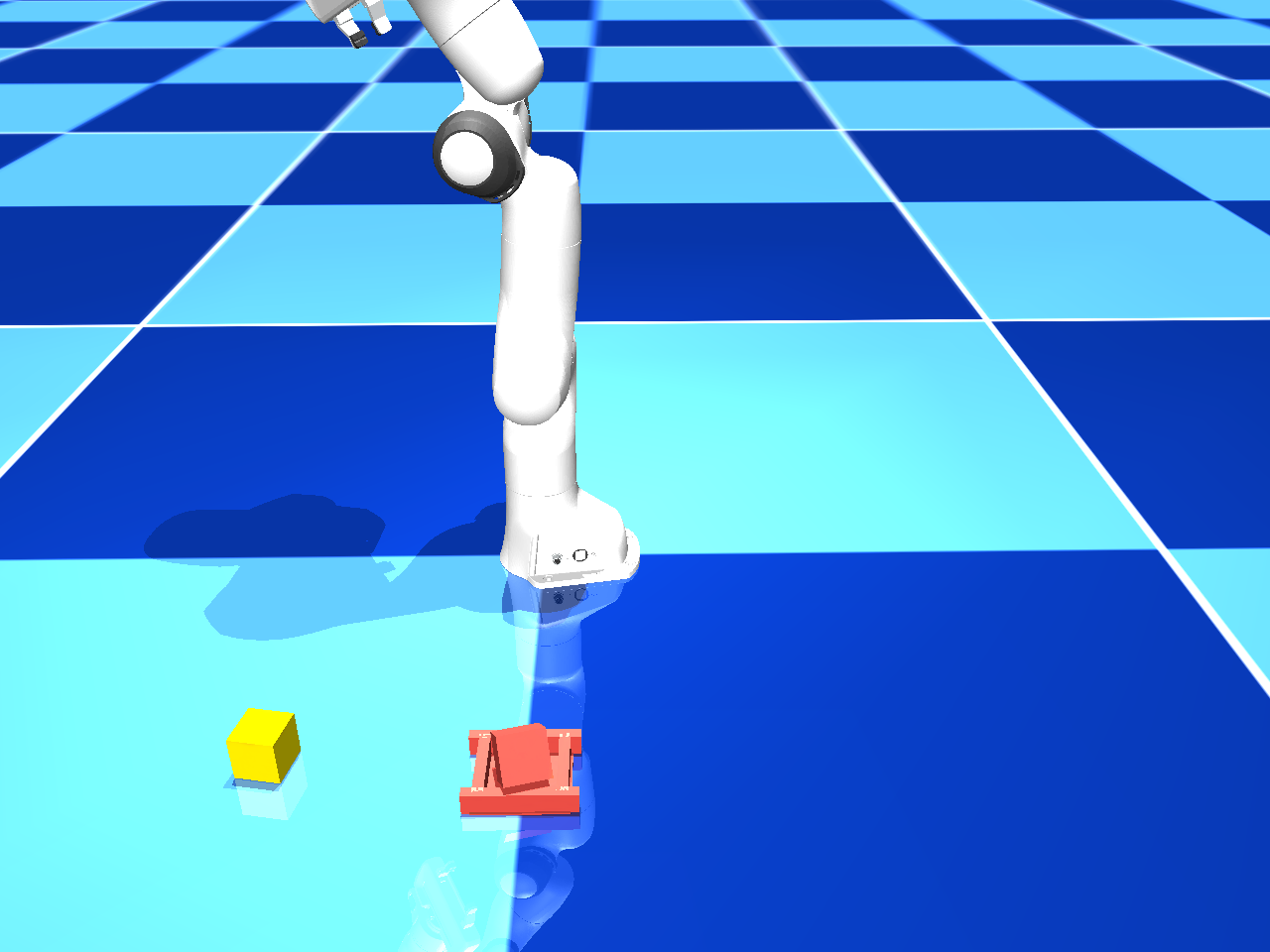}
         \caption{\texttt{TwoCube-Blocked} - complete}
         \label{fig:blockblocks-done}         
    \end{subfigure}
    \\
    \begin{subfigure}[b]{0.3\textwidth}
         \centering
         \includegraphics[width=\textwidth]{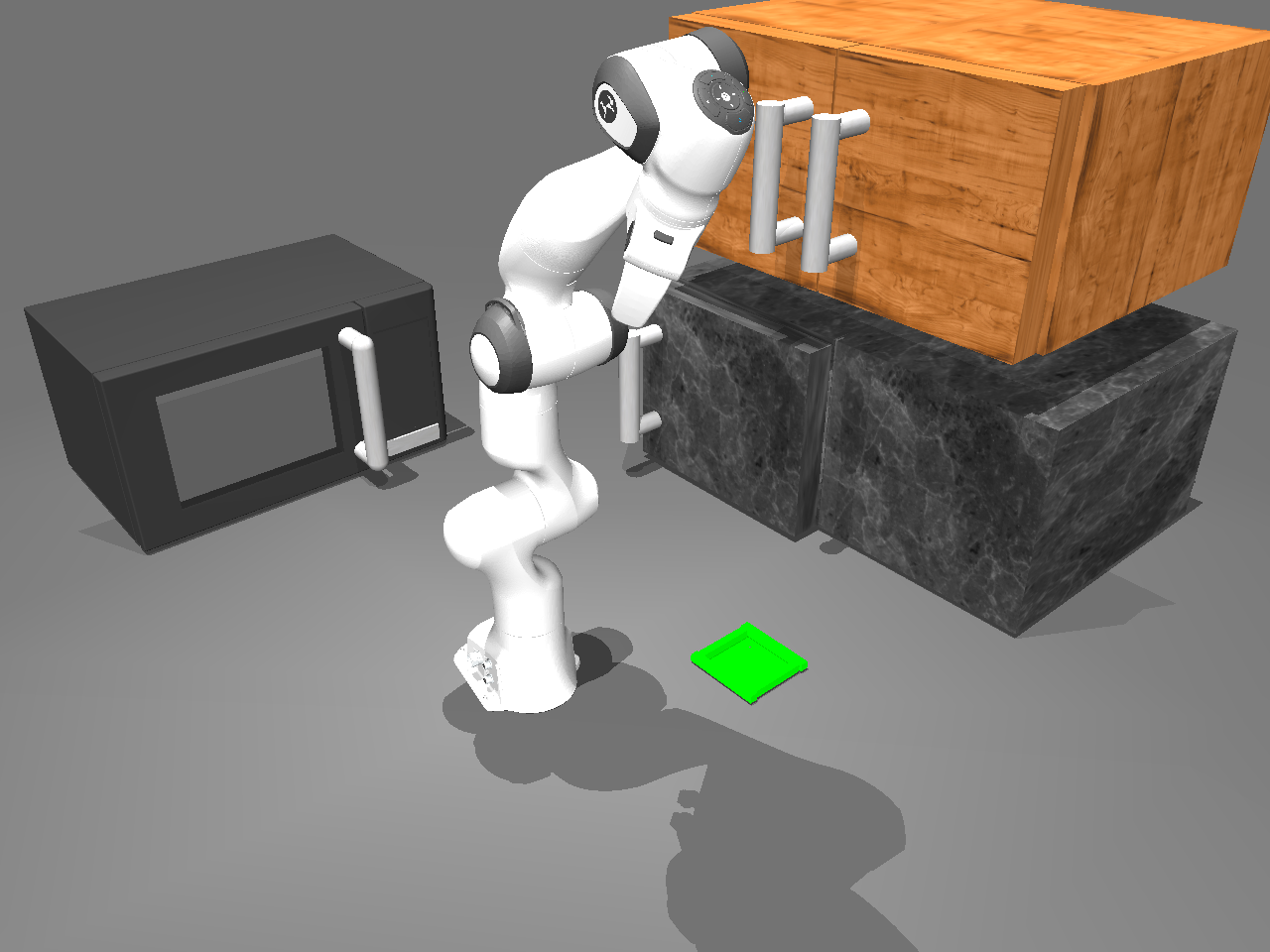}
         \caption{\texttt{Composite-Explore} - initial}
         \label{fig:blockblocks-init}         
    \end{subfigure}
    \begin{subfigure}[b]{0.3\textwidth}
         \centering
         \includegraphics[width=\textwidth]{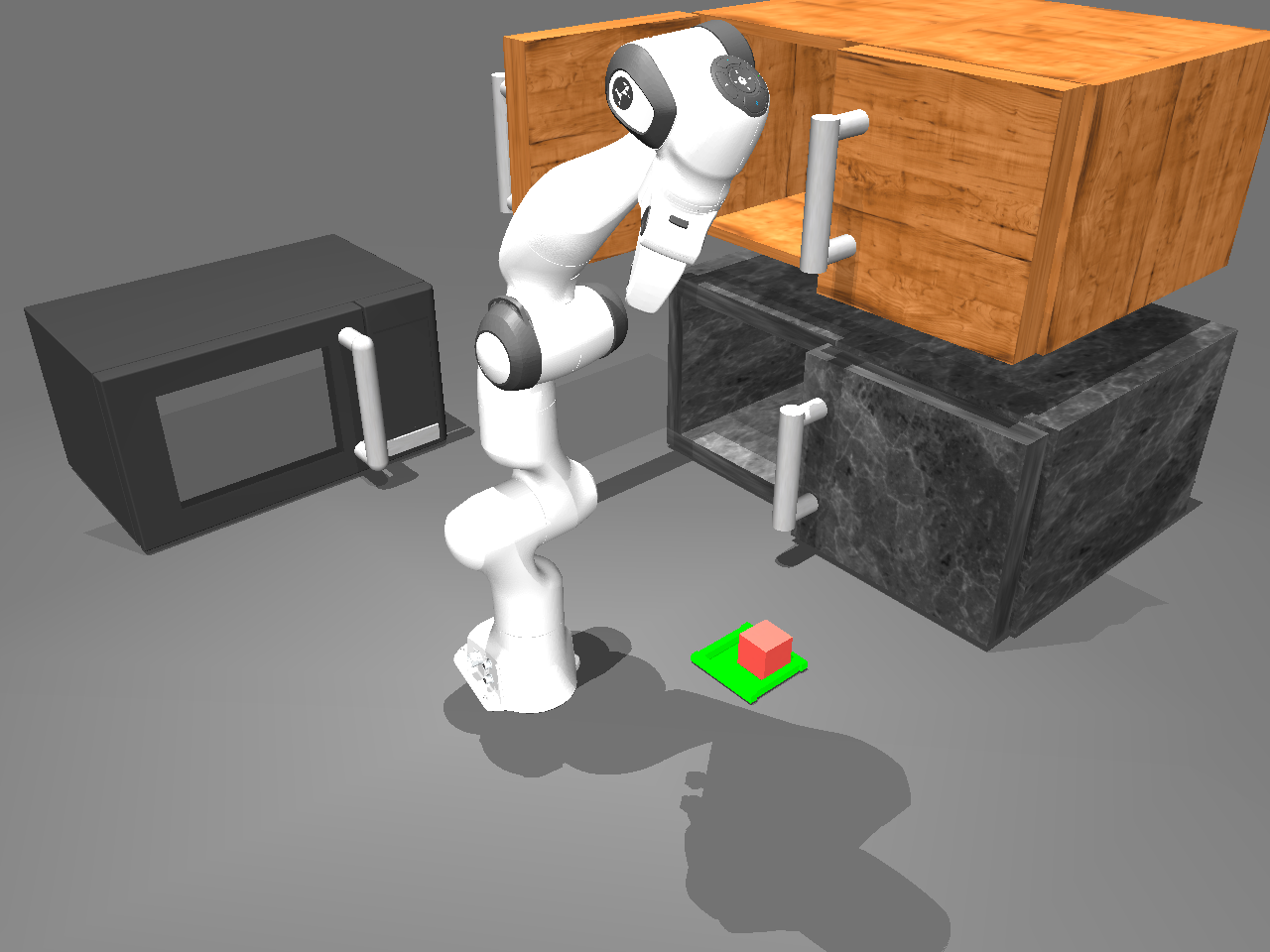}
         \caption{\texttt{Composite-Explore} - complete}
         \label{fig:blockblocks-done}         
    \end{subfigure}
    \caption{Initial and final scenes of the tested environments.}
    \label{fig:env-init}
\end{figure}

\clearpage
\section{Prompts} \label{sec:prompts}
\setcounter{figure}{-1}

\begin{figure}[!htb]
    \centering         
    \includegraphics[width=\textwidth]{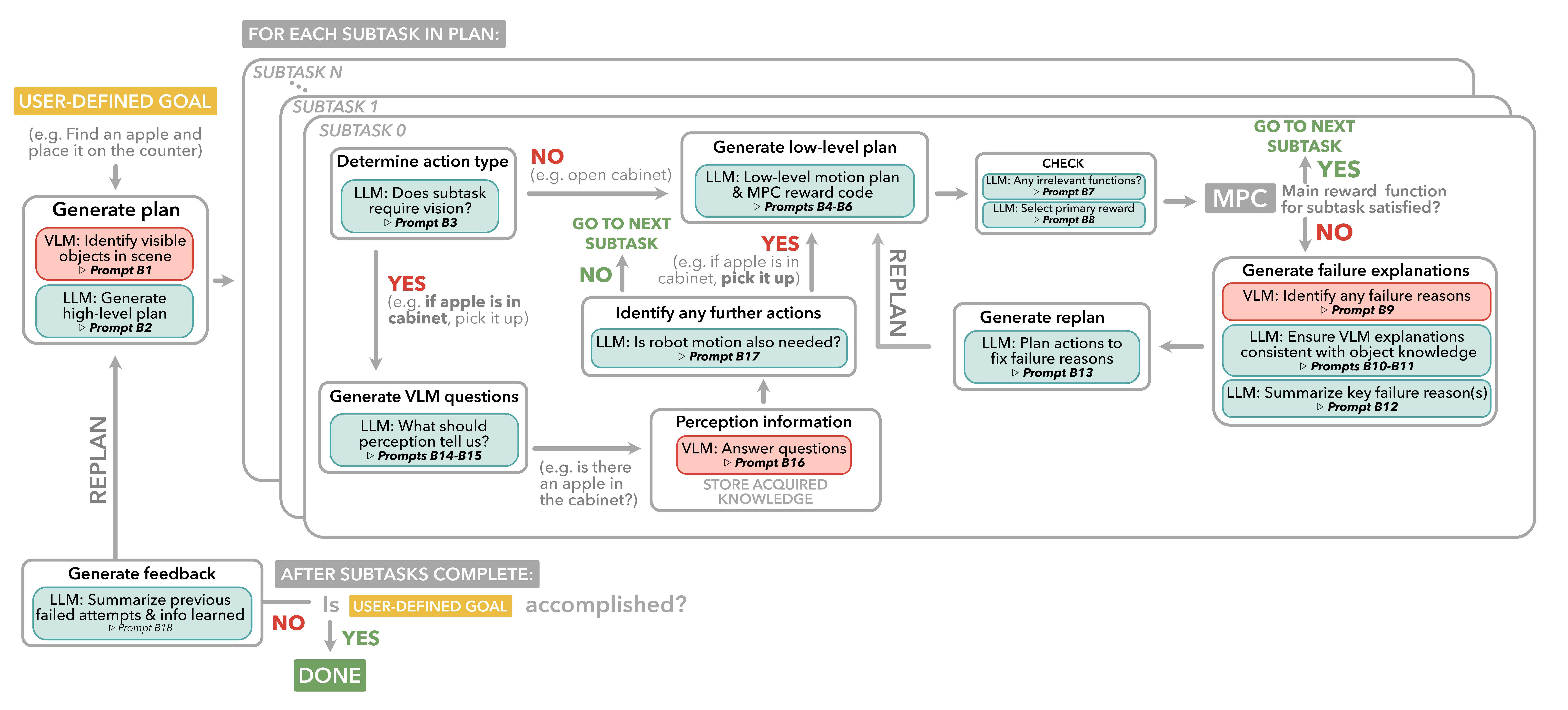}
    \caption{\ourmodel prompt roadmap. We show key modules in the pipeline and indicate under each one what prompts are utilized. This is an expanded version of the Figure~\ref{fig:pipeline}
    }
    \label{fig:roadmap}    
\end{figure}

In Figure~\ref{fig:cabinet_rollout}, we show a example rollout of \ourmodel solving the \texttt{cabinet-closed} task In Figure~\ref{fig:roadmap}, we also show a prompt roadmap.
Furthermore, in Figures~\ref{fig:vlm_perceive_env}-\ref{fig:llm_new_plan} we show step-by-step all the prompts we use for the Planners, Perceiver, and Verifier. We show the prompts as the robot would receive them while excuting a task. The prompts are coloured according to the module it comes from -- \textbf{LLM Planners: blue, VLM Perceiver: pink, Verifier: gray}.

\begin{figure}[!htb]
        \centering  
        \vspace{-3mm}
        \includegraphics[width=0.88\textwidth, trim={0 1cm 0 1cm}, clip]{figures/cabinet_rollout-compressed.pdf}
        \caption{Rollout of robot solving \texttt{cabinet-closed}. The high-level plan is shown in the top row. The second row shows each subtask and the corresponding reward functions generated by the Low-Level Planner, as well as Perceiver feedback. If the subtask fails, its box is colored in red. If the plan is completed and the goal is achieved, its box is green. }
        \label{fig:cabinet_rollout}
\end{figure}
\FloatBarrier

\FloatBarrier

\newpage
\begin{figure}[!htb]
\begin{lstlisting}[backgroundcolor=\color{lightpink}, language=TeX]
Do you see a(n) {0}?
\end{lstlisting}
\caption{VLM prompt for perceiving objects in the environment. \{0\} is an object that the robot knows how to interact with. The VLM is prompted with the list of objects the robot knows how to interact with. If the VLM replies with "yes", that object is added to a list of observed objects.}
\label{fig:vlm_perceive_env}
\end{figure}

\FloatBarrier

\begin{figure}[!htb]
\begin{lstlisting}[backgroundcolor=\color{lightblue}, language=TeX]
A stationary robot arm is in a location where it sees the following list of objects:

{0}

The robot has the following goal: {1}

Propose high-level, abstract subtasks of what the robot needs to do to {1}. The plan can only use one object. 

For example, if the goal is to find a fork, one plan might be:

<thought>To find the fork, I will start by looking inside the drawer.</thought>
[start plan]
    >Open the drawer
    >Look inside the drawer 
    >Grab the fork 
[end plan]

Rules:
1. You have access to the following objects: {0}. Do not create new objects.
2. Generate a plan that interacts with only one object from the list at a time. Keep it as short as possible. Most plans should be under 5 steps.
3. Assume that every action is completed successfully.
4. Assume the first thing you try works.
5. Your plan should only propose one way of accomplishing the task.
6. The robot only has one arm and it cannot hold two things at a time. Remember that when you are deciding on the order of actions.
7. Enclose your thought process with a single pair of tag <thought> and </thought>
8. Enclose your plan with the a single pair of tag [start plan] and [end plan]

{2}
\end{lstlisting}

\caption{LLM prompt for generating high-level task plans. \{0\} is the list of objects the robot can see (for example: [cabinet, blue kettle, microwave], \{1\} is the overall task goal (for example: find the green apple), \{2\} are previous plans that were attempted but failed (see \ref{fig:llm_new_plan}). }
\label{fig:llm_high_level_plan}
\end{figure}

\FloatBarrier

\begin{figure}[!htb]
\begin{lstlisting}[backgroundcolor=\color{lightblue}, language=TeX]
A robot was asked to do this action:
    > {0}
If the central verb is related to vision, answer yes.
\end{lstlisting}
\caption{LLM prompt to determine whether the action that the Planner asked the robot to do involves vision or not. If no, then the Planner is called to generate MPC reward functions (see \ref{fig:llm_motion_plan}-\ref{fig:llm_replan}). If yes, the Perceiver is called (see \ref{fig:llm_vlm_question}-\ref{fig:vlm_completion}). Examples for \{0\}: ``Compare the color of the left cube with the crate", ``Open the microwave".}
\label{fig:llm_reward_type}
\end{figure}

\FloatBarrier

\begin{figure}[!htb]
\begin{lstlisting}[backgroundcolor=\color{lightblue}, language=TeX]

We have a stationary robot arm and we want you to help plan how it should move to perform tasks using the following template:
[start of description]
The manipulator's palm should move close to {{CHOICE: {0}}}.{1}{2}
[end of description] 
Rules:
0. You cannot use one line twice!!!!
1. If you see phrases like [NUM: default_value], replace the entire phrase with a numerical value.
2. If you see phrases like {{CHOICE: choice1, choice2, ...}}, it means you should replace the entire phrase with one of the choices listed.
3. If you see [optional], it means you only add that line if necessary for the task, otherwise remove that line.
4. The environment contains {0}. Do not invent new objects not listed here.
5. I will tell you a behavior/skill/task that I want the manipulator to perform and you will provide the full plan, even if you may only need to change a few lines. Always start the description with [start of description] and end it with [end of description].
6. You can assume that the robot is capable of doing anything, even for the most challenging task.
7. Your plan should be as close to the provided template as possible. Do not include additional details.
8. Your plan should be as concise as possible. Do not include or make up unncessary tasks.
9. Each object can only be close to or far from one thing. 

This is the entire procedure:
{4}

These are the observations we have made so far:
{5}

Create a plan for the following action:
    > {6}
\end{lstlisting}
\caption{LLM prompt to determine the low-level motion plan for the robot. \{0\} is the list of objects the robot can interact with. \{1\} and \{2\} are modifiers, depending on what type of motion is involved. The Planner is asked to determine whether motion is involved (see \ref{fig:llm_motion_plan_mod}. If yes, then \{1\} becomes: \texttt{object1=\{\{CHOICE: \{0\}\}\} should be \{\{CHOICE: close to, far from\}\} object2=\{\{CHOICE: \{0\}\}\}.} If no, then \{1\} becomes \texttt{[optional] object1=\{\{CHOICE: \{0\}\}\} should be close to object2=\{\{CHOICE: \{0\}\}\}.
[optional] object1=\{\{CHOICE: \{0\}\}\} should be far from object2=\{\{CHOICE: \{0\}\}\}.} The modifier \{2\} is added if there are any joints in the scene that are involved with objects the robot can interact with: \texttt{[optional] joint=\{\{CHOICE: \{3\}\}\} needs to be \{\{CHOICE: open, closed\}\}.} (where \{3\} is the list of joints) (adapted from \citep{yu2023language}) \{4\} is the entire plan the robot was given to execute the goal. \{5\} includes any observations made by the Perceiver (see \ref{fig:vlm_object_state}) using the following format: \texttt{Q: <question to Perceiver>, A: <answer from Perceiver>}. \{6\} is the action the motion plan should be made for.
}
\label{fig:llm_motion_plan}
\end{figure}

\FloatBarrier

\begin{figure}[!htb]
\begin{lstlisting}[backgroundcolor=\color{lightblue}, language=TeX]
A robot arm has to do this action:
    > {0}
Does this action necessarily involve relocating an object to a different location that does not involve the robot arm? Answer with yes or no.
\end{lstlisting}
\caption{LLM prompt to determine if relocation is needed in order to determine the motion plan modifier (see \ref{fig:llm_motion_plan}.}
\label{fig:llm_motion_plan_mod}
\end{figure}

\FloatBarrier

\begin{figure}[!htb]
\begin{lstlisting}[backgroundcolor=\color{lightblue}, language=TeX]
We have a plan of a robot arm with palm to manipulate objects and we want you to turn that into the corresponding program with following functions:

    def minimize_l2_distance_reward(name_obj_A, name_obj_B)

where name_obj_A and name_obj_B are selected from {0}. This term sets a reward for minimizing l2 distance between name_obj_A and name_obj_B so they get closer to each other. rest_position is the default position for the palm when it's holding in the air.

    def maximize_l2_distance_reward(name_obj_A, name_obj_B, distance=0.5)

This term encourages the orientation of name_obj to be close to the target (specified by x_axis_rotation_radians).

    def execute_plan(duration=2)

This function sends the parameters to the robot and execute the plan for "duration" seconds, default to be 2.

    def set_joint_fraction_reward(name_joint, fraction)

This function sets the joint to a certain value between 0 and 1. 0 means close and 1 means open. name_joint needs to be select from {1}.

    def reset_reward()

This function resets the reward to default values.
Example plan: To perform this task, the manipulator's palm should move close to object1=faucet_handle. object1 needs to be lifted to a height of 1.0.
This is the first plan for a new task.
Example answer code:

    import numpy as np

    reset_reward()
        # This is a new task so reset reward; otherwise we don't need it
    minimize_l2_distance_reward("palm", "faucet_handle")
    set_joint_fraction_reward("faucet", 1.0)

    execute_plan(4)

Remember:
1. Always format the code in code blocks. In your response execute_plan should be called exactly once at the end.
2. Do not invent new functions or classes. The only allowed functions you can call are the ones listed above. Do not leave unimplemented code blocks in your response.
3. The only allowed library is numpy. Do not import or use any other library.
4. If you are not sure what value to use, just use your best judge. Do not use None for anything.
5. Do not calculate the position or direction of any object (except for the ones provided above). Just use a number directly based on your best guess.
6. You do not need to make the robot do extra things not mentioned in the plan such as stopping the robot.

The action to perform is {2} and the plan is:
{3}
\end{lstlisting}
\caption{LLM prompt to generate MPC reward functions. \{0\} is the list of objects the robot can interact with, to which we also append the word ``palm" to represent the robot hand. \{1\} is the list of object joints. \{2\} is the high-level action  the robot needs to perform and \{3\} is the motion plan  generated from \ref{fig:llm_motion_plan}. Adapted from \citep{yu2023language}.}
\label{fig:llm_reward_func}
\end{figure}

\FloatBarrier

\begin{figure}[!htb]
\begin{lstlisting}[language=TeX]
This is a motion plan generated for a robot:

{0}

This is a reward function generated to complete one step in the motion plan:

{1}

The function minimize_l2_distance_reward() refers to bringing two objects close together.
The function maximize_l2_distance_reward() refers to moving two objects further apart.
The function set_joint_fraction_reward() refers to opening or closing an object (0 for closed, 1 for open)
The function set_obj_z_position_reward() specifies the target height of an object.
The function set_obj_orientation_reward() specifies the target rotation of an object.

Which step in the motion plan is the function referring to? Return the step using <step></step> tags. If it does not refer to any of them, return <step>-1</step>
\end{lstlisting}
\caption{The Verifier checks that every reward function generated corresponds to a step in the motion plan. If it does not, the function is removed. \{0\} is the motion plan generated from Figure~\ref{fig:llm_motion_plan} and \{1\} is one of the generated reward functions (they are looped over individually).}
\label{fig:verifier_reward_func}
\end{figure}

\FloatBarrier

\begin{figure}[!htb]
\begin{lstlisting}[language=TeX]
A stationary robot arm was asked to do the following motion plan to complete the task '{0}':

{1} 

After which step in the motion plan will the task '{0}' be satisfied? First, explain your thought then answer the step number enclosed with the tag <step> and </step>. Opening a joint can also mean activating it depending on the context. You must select one. If you think none of the steps does, select the closest one.
\end{lstlisting}
\caption{The verifier selects which step in the motion plan is considered to be the most important. The reward function generated for that step in the motion plan becomes labelled as the primary reward function. \{0\} is the action the robot is currently doing and \{1\} is the motion plan.}
\label{fig:verifier_reward_func}
\end{figure}

\FloatBarrier

\begin{figure}[!htb]
\begin{lstlisting}[backgroundcolor=\color{pink}, language=TeX]
A robot is in a simulation environment where it can interact with any object like in the real world. The robot would like to {0} but it cannot. Is there something in this scene preventing that, other than the robot? Assume the robot can interact with anything. These are the names of the objects in our scene: {1}
In a simulation, a robot wants to {0} but can't. Is anything else, besides the robot, blocking it? Check the objects in the scene: {1}.
Robot in a simulation wants to {0}, can't. Something else stopping it? Objects in scene: {1}.
A robot can engage with any item. It wants to {0} but can't. Is an object in this scene, apart from the robot, hindering it? Objects present: {1}
I would like to {0} but I cannot. Is there something in this scene preventing that, other than the robot? These are the objects in the scene: {1}
I would like to {0} but I am unable to. Is there something in this scene preventing me from doing that? Ignore the robot. These are the names of the objects: {1}
\end{lstlisting}
\caption{If the robot is unable to satisfy the primary reward function, the Perceiver is queried on whether there are any obstacles in the scene. The Perceiver is called once for each question (total of 6). Questions 2-4 and 6 were variations generated for Q1 and Q5, respectively, using ChatGPT.}
\label{fig:vlm_action_failed_new}
\end{figure}

\FloatBarrier

\begin{figure}[!htb]
\begin{lstlisting}[language=TeX]
We have access to the following objects in our scene: {0}

You are given a sentence describing an image of the scene, but it may have gotten the names of the objects wrong. Does this sentence contain objects that are not in our scene or get the names of the objects wrong? Start your answer with yes or no.

{1}
\end{lstlisting}
\caption{For every explanation from the Perceiver, the Verifier is called to determine whether the explanation lists objects that do not exist in the scene. \{0\} is the list of objects in the scene and \{1\} is the explanation from the Perceiver. If the Verifier answers with 'yes', the explanation is passed to \ref{fig:verifier_remap2} for object remapping.}
\label{fig:verifier_remap1}
\end{figure}

\FloatBarrier

\begin{figure}[!htb]
\begin{lstlisting}[language=TeX]
We have access to the following objects in our scene: {0}

You are given a sentence describing an image of the scene, but got the names of the objects wrong. Rewrite this sentence using the closest object(s) in our environment: 

{1}

Rules:
You can only use objects in the scene. Use your best judgement.
\end{lstlisting}
\caption{If the Verifier identifies that the Perceiver explanation contains objects that are not listed in the scene, the Verifier rewrites the explanation (\{1\}) using the closest objects in our scene (\{0\}).}
\label{fig:verifier_remap2}
\end{figure}

\FloatBarrier

\begin{figure}[!htb]
\begin{lstlisting}[backgroundcolor=\color{lightblue}, language=TeX]
The stationary robot arm would like to {0} but it cannot. Here are possible reasons why based on images of the scene:

    {1}

Based on the above explanations, what are the top reason(s) why the robot cannot {0}? List each reason on a separate line, enclosed with the tag <reason> </reason>. Provide up to two reasons. Be as succinct as possible. You must not include any reasons related to the robot, only reasons related to objects in the scene.
\end{lstlisting}
\caption{The Planner receives all explanations from the Perceiver (\{1\}, see \ref{fig:vlm_action_failed_new} -\ref{fig:verifier_remap2}) and summarizes them into key reasons explaining why the robot could not do the action \{0\}.}
\label{fig:llm_summary}
\end{figure}

\FloatBarrier

\begin{figure}[!htb]
\begin{lstlisting}[backgroundcolor=\color{lightblue}, language=TeX]

<Prompt from *@\ref{fig:llm_high_level_plan}@*>

One or more previous attempts failed. Below are the details.
---------------------------------- attempt #1 ----------------------------------
This attempt failed when executing '\{0\}'.The plan failed because the robot was not able to execute this action: '\{1\}'. This was identified as a possible reason the action failed: '{2}'.
...
---------------------------------- attempt #*@\emph{R}@* ----------------------------------
...
---------------------------- end of failed attempts ----------------------------
Reminder to propose a different plan than the above failed attempts.
\end{lstlisting}
\caption{If the robot does not succeed in performing an action, the Planner is able to replan how the robot does the action by providing the failure reason(s) from the Perceiver from the \emph{R} failures reasons. Example for \{0\}: 'Place the red\_cube on the crate'. Example for \{1\}: 'Place the red\_cube on the crate (incidentally the action is the same as the overall goal, but it doesn't have to be)'. Example for \{2\}: 'The most probable reason why the robot cannot place the red\_cube on top of the crate is that the yellow cube is currently on top of the crate, which would prevent the robot from doing so.'}
\label{fig:llm_replan}
\end{figure}

\FloatBarrier

\begin{figure}[!htb]
\begin{lstlisting}[backgroundcolor=\color{lightblue}, language=TeX]
You are a robot in the process of executing this plan, with the overall goal to '{0}':

{1}

You are currently performing this action: '{2}'. You have access to a perception model that can answer your questions related to vision. 

{3}

What question do you want to ask the perception model in order to get the answer to '{2}'? You can ask up to two questions. You don't have to ask if the information is already sufficient. Avoid asking the vision model to compare things. Enclose each of your questions with the tag <question> </question>.
\end{lstlisting}
\caption{If the action requires calling the Perceiver, this prompt is used to determine what questions the Planner wants to ask the Perceiver. \{0\} is the overall goal, \{1\} is the high-level plan, \{2\} is the action the robot is currecntly performing, \{3\} are the observed objects in the scene.}
\label{fig:llm_vlm_question}
\end{figure}

\FloatBarrier

\begin{figure}[!htb]
\begin{lstlisting}[backgroundcolor=\color{lightblue}, language=TeX]
What type of question is this asking perception model: '{0}'? Choose your answer from [OBJECT_PRESENCE, OBJECT_ATTRIBUTE, NEITHER]
\end{lstlisting}
\caption{Prompt to field what category of question the Planner wants to ask the Perceiver model. \{0\} is the output from \ref{fig:llm_vlm_question}.}
\label{fig:llm_vlm_question_type}
\end{figure}

\FloatBarrier

\begin{figure}[!htb]
\begin{lstlisting}[backgroundcolor=\color{lightpink}, language=TeX]
<Output from *@\ref{fig:llm_vlm_question}@*. The names of the objects in our scene are: {0}. {1} \end{lstlisting}
\caption{Perceiver query on information about the state of objects in the scene from \ref{fig:llm_vlm_question}. States are related to object presence or object attributes. Examples of queries: `Look for the apple in the cabinet', `Check the color of the crate'.}
\label{fig:vlm_object_state}
\end{figure}

\FloatBarrier

\begin{figure}[!htb]
\begin{lstlisting}[backgroundcolor=\color{lightblue}, language=TeX]
A robot was tasked to do this plan: 

{0}

The robot is currently doing this action: '{1}'.

To do the action, the robot asked a perception model the following questions (Q) and got the answers (A):

{2}

After receiving this answer, has the robot completed the action '{1}'?  Begin your answer with yes or no. If your answer begins with no, write the remaining action that needs to be completed using <Action></Action> tags.\end{lstlisting}
\caption{After the Perceiver has provided information, the Planner is asked to determine whether the action is completed. If not, it generates MPC reward functions to finish the action (see \ref{fig:llm_motion_plan}-\ref{fig:llm_new_plan}).}
\label{fig:vlm_completion}
\end{figure}

\FloatBarrier

\begin{figure}[!htb]
\begin{lstlisting}[backgroundcolor=\color{lightblue}, language=TeX]
<Prompt from *@\ref{fig:llm_high_level_plan}@*>

One or more previous attempts failed. Below are the details.
---------------------------------- attempt #1 ----------------------------------
The proposed plan was:
<thought>{0}</thought>
[start plan]
{1}
[end plan]
The plan failed because {2}. 
...
---------------------------------- attempt #*@\emph{P}@* ----------------------------------
...
---------------------------- end of failed attempts ----------------------------
Reminder to propose a different plan than the above failed attempts.
\end{lstlisting}
\caption{If by executing the plan or the replan and the goal is still not accomplished, the Planner is prompted to generate a new plan using the prompt in Figure~\ref{fig:llm_high_level_plan}. The Planner is allowed to generate a new plan \emph{P} times. before the task is considered undoable.}
\label{fig:llm_new_plan}
\end{figure}

\clearpage
\section{Error Evaluation} \label{sec:error_analysis}
Here we provide three error cases of \ourmodel and their analyses.

\paragraph{Case 1 (from \texttt{Cabinet-Blocked}, Figure~\ref{fig:error-cases}(a))} The robot tried to open the cabinet door but failed and the Perceiver gave a correct diagnosis to remove the bar from the handle. However, when generating the reward functions to remove the bar, the LLM selected the wrong primary reward function, as demonstrated below:
 
 \begin{lstlisting}[language=Python]
 reset_reward()
 minimize_l2_distance_reward("palm", "red_block_right_side", primary_reward=True)
 maximize_l2_distance_reward("red_block_right_side", "target_position_in_cabinet")
 execute_plan()
 \end{lstlisting}
 
 The correct primary function should be the second one. As a result, MPC ended prematurely before the robot could remove the bar. The robot was not able to remove the bar in the following steps.
 
 \paragraph{Case 2 (from \texttt{Kitchen-Explore}, Figure~\ref{fig:error-cases}(b)))} The robot tried to open the microwave door but failed due to a kettle obstructing the path. The Perceiver gave five diagnoses, of which three claimed that the kettle was blocking the way, one claimed the cabinet door was blocking the way, and one did not give any conclusive diagnosis. The summary LLM concluded that it was the cabinet door that blocked the action. The robot went on to interact with the cabinet and never removed the kettle.
 
 \paragraph{Case 3 (from \texttt{TwoCube-Color}, Figure~\ref{fig:error-cases}(c)))} The high-level planner proposed a plan where the first step was \emph{``Identify the cube with the same colour as the crate''}. This was a VLM action. However, the LLM proposed to ask the VLM \emph{``Which cube has the same colour?''}, which was a bit vague, resulting in the VLM answering \emph{``The same color cube is the yellow cube and the yellow cube in the middle of the blue cube group.''}. This answer did not provide the necessary information to solve the task. Eventually, the robot put the wrong cube on the crate.
 
 \paragraph{Case 4 (from \texttt{TwoCube-Blocked}, Figure~\ref{fig:error-cases}(d)))} After the robot was not able to execute the task ``Place the red cube on the crate", the Perceiver was called to help identify any issues. The Perceiver's diagnoses all mentioned that the robot was holding the red cube but did not identify the yellow cube as blocking the crate, and so the Planner's summary of the VLMs diagnoses was: ``Based on the given information, the most probable reason why the robot cannot place the red\_cube on the crate is because it is currently holding the red cube." However, it's also important to note that \texttt{TwoCube-Blocked} used GPT-4V which severly limits the number of ouput tokens from the model, and so a lot of explanations were cut off (for example: ``In the image provided, the robot is holding the red cube, which is currently"). 

\subsection{LLM Diagnosis with Ground-truth Data}\label{sec:no_vlm_diagnosis}

An alternative way to diagnose errors is to input the simulator ground-truth state of the objects to an LLM and ask the LLM to infer a possible reason. We show a scenario here where the cabinet door is locked by a red bar (from \texttt{Cabinet-Blocked}) and test the capability of LLM in this regard. The prompt we use is:

\begin{lstlisting}
A robot is in a simulation environment where it can interact with any object like in the real world. The robot would like to open the wooden cabinet but it cannot. Is there something in this scene preventing that, other than the robot? Assume the robot can interact with anything. These are the names of the objects in our scene: wooden cabinet, red block, yellow cube.

The data of the objects are:
{
    "joints": {
        "wooden_cabinet": 0.16235716765387215
    },
    "objects": {
        "wooden_cabinet_handle": [0.08292539, 0.50163332, 0.21],
        "target_position_in_wooden_cabinet": [0.2,    0.85,   0.0775],
        "red_block_right_side": [0.2158331,  0.55110136, 0.15143364],
        "yellow_cube": [-0.24545329,  0.01631876,  0.03599184],
        "palm": [0.2231031,  -0.00875834,  0.76900653]
    }
}
\end{lstlisting}

And the response from GPT-4 is:

\begin{lstlisting}
The data does not indicate any object that would prevent the robot from opening the cabinet. There's no information about any object blocking the cabinet or its handle. The problem could be due to the robot itself as it might not possess the necessary functionality to interact with the objects as intended. The cabinet's joint also has a non-zero value, which might mean it's not locked or stuck. Hence, there is likely a mechanical limitation with the robot or an issue with its programming.
\end{lstlisting}

This example demonstrates that LLM is not capable of directly inferring spatial relations of objects through ground-truth data.

\section{Additional Experiments}

\subsection{VLM Ablation}
\label{sec:vlm_ablation}

We conduct a VLM ablation study for Tasks 3-4 using two open-source models (Qwen-VL-Chat7B~\citep{Qwen-VL} and Llava-1.5-7B~\citep{liu2023improvedllava}) and GPT-4V\footnote{https://openai.com/research/gpt-4}. The results are shown in Table~\ref{tab:VLM_ablation}. We measure (a) their ability to recognize objects, reported as a percentage of the number of object successfully detected in the scene (column 2), (b) their ability to reason about obstacles in the scene, reported as a percentage of times the correct reason was identified out of all the prompts it was given in Figure~\ref{fig:vlm_action_failed_new}, and (c) the ability for the LLM to consilidate the outputs in (b) into a single summary reason of why the robot cannot perform an action given its knowledge about the scene. We found that Qwen somtimes struggled with object detection of smaller objects, and so we coupled it with Segment Anything Model (SAM)~\cite{kirillov2023segany} to first segment the objects in the scene. We found that all models did well with object recognition (except for Qwen when not used with SAM). For object reasoning, Qwen + SAM was able to get the correct scene error in 50-67\% of the prompts it was given, and then the LLM was able to summarize the prompts to generate the correct error reason overall. The reason the LLM was able to do this despite the VLM not giving perfect answers was that the remaining VLM answers pertained where the robot was located or a general comment about the objects in the scene. Llava tended to reply that it was unable to reason because the scene was a simulation and not real life. GPT-4V had the best overall performance in all categories, but API calls to it are still restricted. All ablations were repeated over 5 runs.

\FloatBarrier
\begin{table}[!htb]
    \begin{subtable}[h]{\textwidth}
    \centering
    \begin{tabular}{ccccc}
        \toprule
        & \multicolumn{4}{c}{Models}\\
        \midrule
        Scenarios & Qwen + SAM & Qwen & Llava & GPT-4V \\
        \midrule
        VLM object detection & $100\%$ & $66\%$ & $100\%$ & $100\%$ \\
        VLM Reasoning & $67\%$ & $0\%$ & $23\%$ & $100\%$ \\ 
        LLM summarization and consistency step & $100\%$ & $0\%$ & $100\%$ & $100\%$ \\
        \bottomrule
    \end{tabular}
        \caption{\texttt{Cabinet-Blocked}}
        \vspace{3mm}
    \end{subtable}
    \begin{subtable}[h]{\textwidth}
    \centering
    \begin{tabular}{ccccc}
        \toprule
        & \multicolumn{4}{c}{Models}\\
        \midrule
        Scenarios & Qwen + SAM & Qwen & Llava & GPT-4V \\
        \midrule
        VLM object detection & $100\%$ & $100\%$ & $100\%$ & $100\%$ \\
        VLM Reasoning & $50\%$ & $66\%$ & $40\%$ & $83\%$ \\ 
        LLM summarization and consistency step & $100\%$ & $100\%$ & $20\%$ & $100\%$ \\
        \bottomrule
    \end{tabular}
        \caption{\texttt{Kitchen-Explore}}
    \end{subtable}
    \caption{VLM ablation study.}
    \label{tab:VLM_ablation}
\end{table}

\subsection{GPT-4V Experiments} 
\label{sec:gpt_4v_baseline}

We run the initial High-Level Planner prompt (Prompt \ref{fig:llm_high_level_plan}) using GPT-4V\footnote{https://openai.com/research/gpt-4v-system-card} on initial task scenes to investigate the ability of GPT-4V to find the correct solution in a single step.
\FloatBarrier
\begin{figure}[!htb]
    \centering
     \begin{subfigure}[b]{0.6\textwidth}   \includegraphics[width=\textwidth]{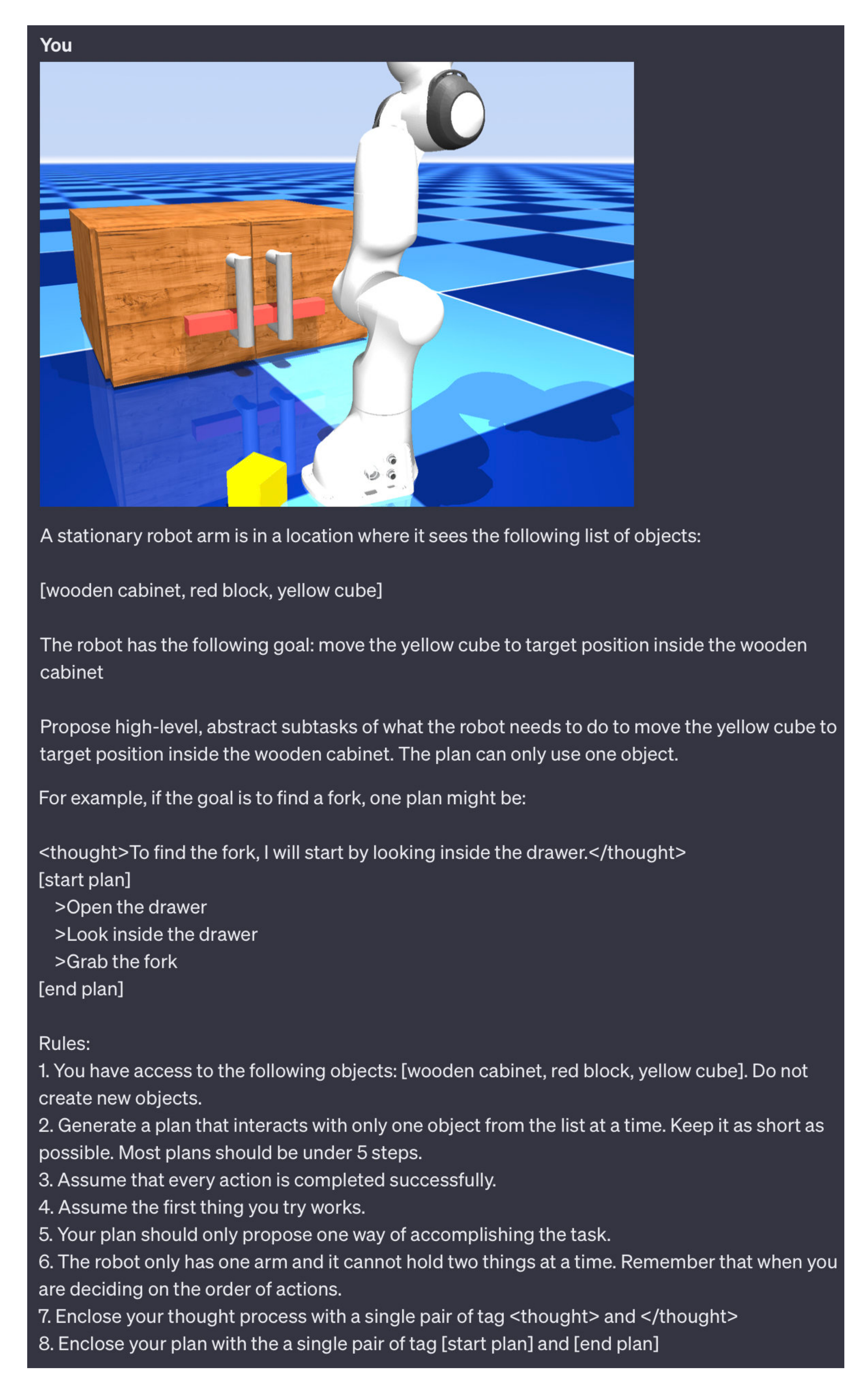}
         \caption{Prompt \ref{fig:llm_high_level_plan} to GPT-4V.}
        \label{fig:gpt4-cabinet-q}
     \end{subfigure}
     
    \vspace{10pt}
      \centering
     \begin{subfigure}[b]{0.55\textwidth}         \includegraphics[width=\textwidth]{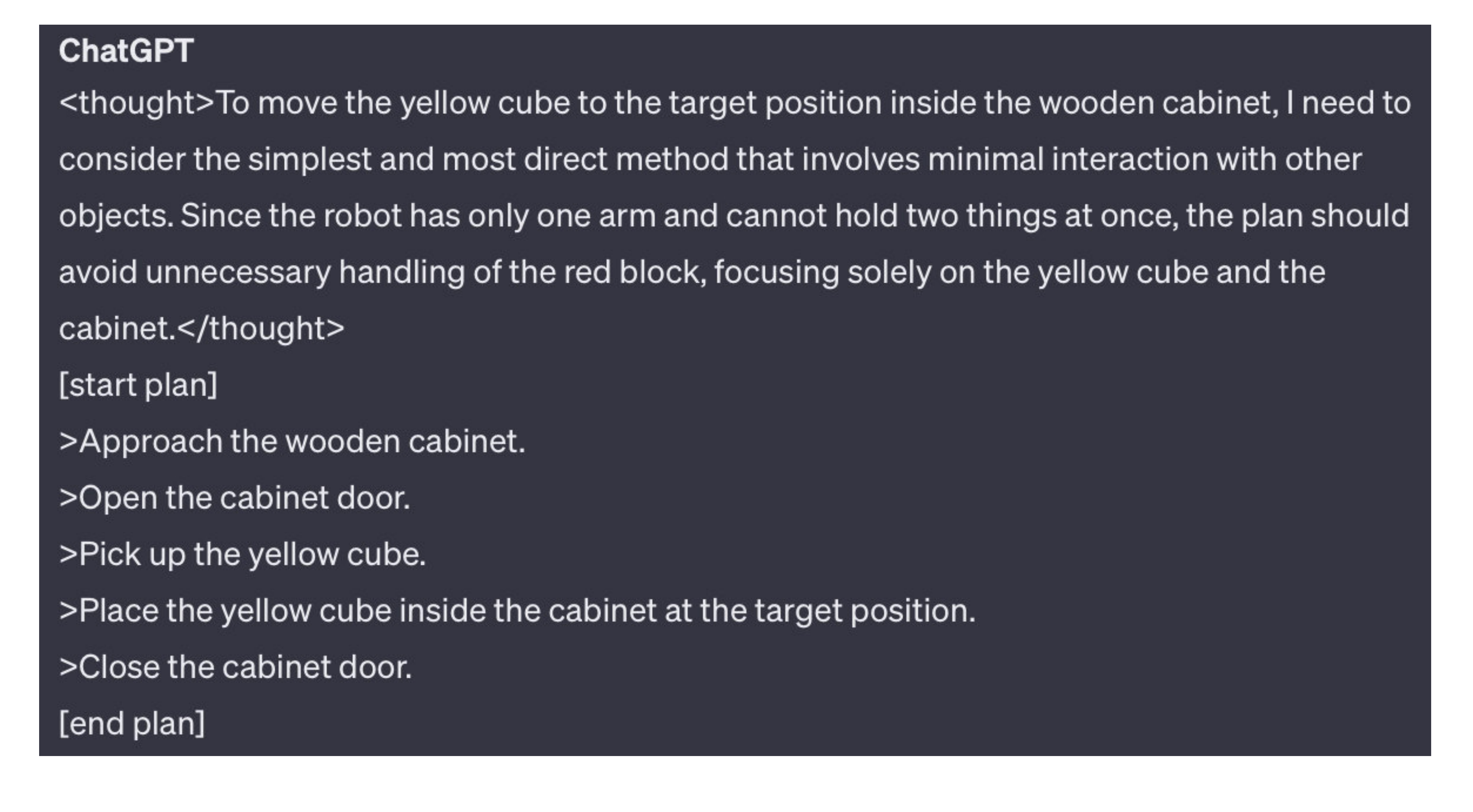}
         \caption{GPT-4V output.}
         \label{fig:gpt4-cabinet-a}
     \end{subfigure}
        \caption{GPT-4V high-level plan for moving the yellow cube inside the wooden cabinet (\texttt{Cabinet-Blocked} in Section \ref{sec:env_1}).}
        \label{fig:gpt4v-cabinet}
\end{figure}

\FloatBarrier

\begin{figure}[!htb]
    \centering
     \begin{subfigure}[b]{0.6\textwidth}   \includegraphics[width=\textwidth]{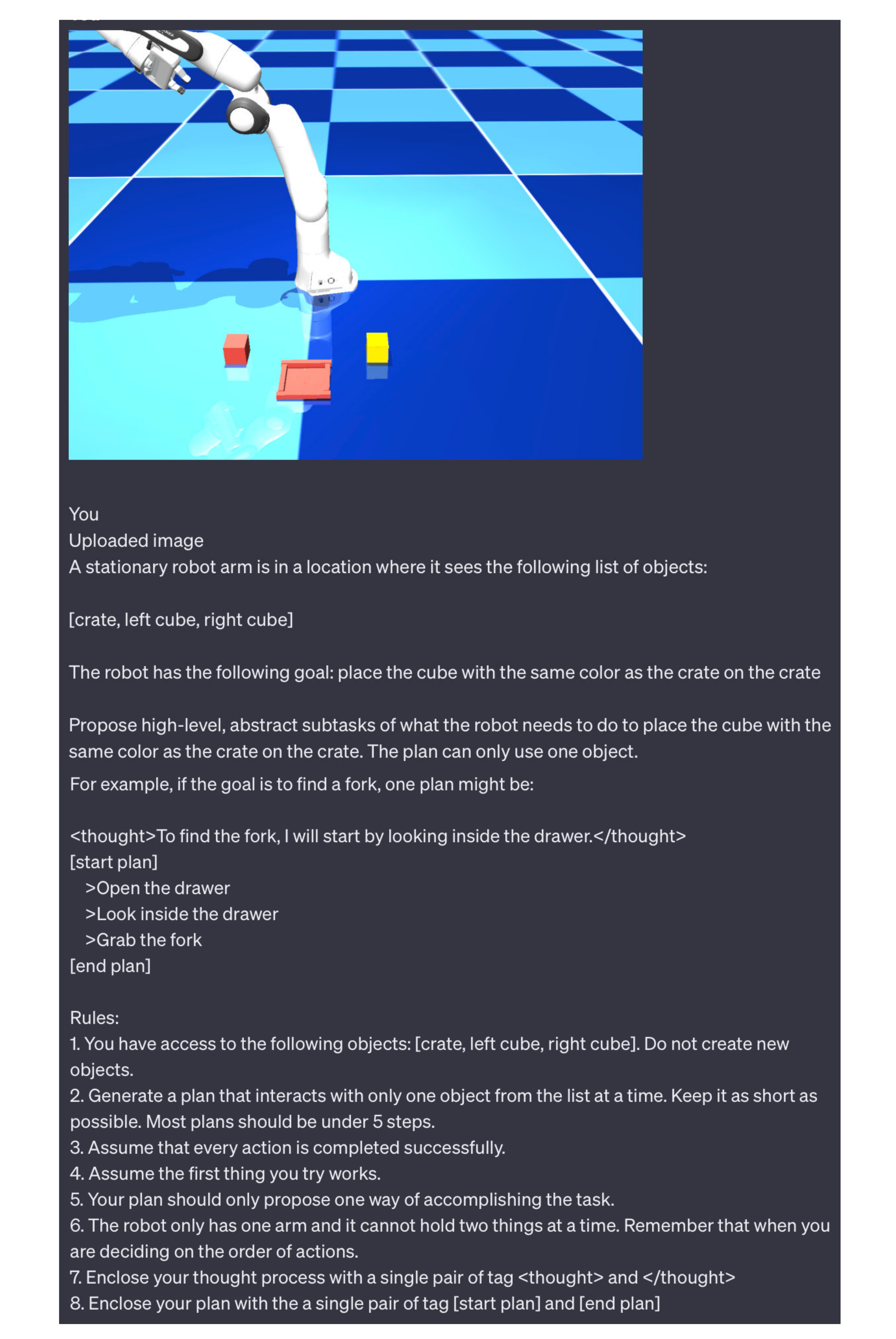}
         \caption{Prompt \ref{fig:llm_high_level_plan} to GPT-4V.}
        \label{fig:gpt4-colour-q}
     \end{subfigure}
     
    \vspace{10pt}
      \centering
     \begin{subfigure}[b]{0.55\textwidth}         \includegraphics[width=\textwidth]{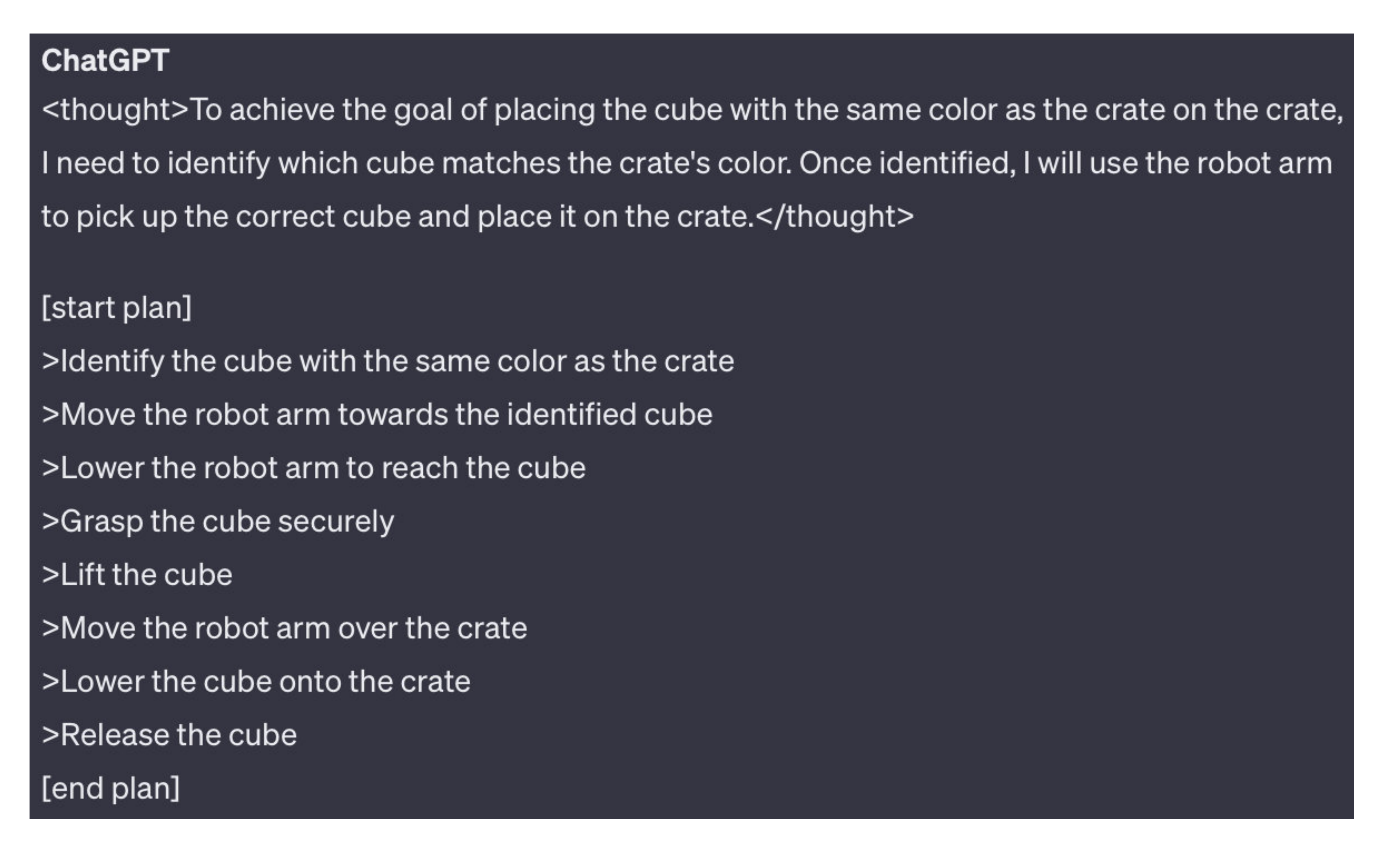}
         \caption{GPT-4V output.}
         \label{fig:gpt4-colour-a}
     \end{subfigure}
        \caption{GPT-4V high-level plan for placing the cube with the same color as the crate on the crate (\texttt{TwoCube-Color} in Section \ref{sec:env_4}).}
        \label{fig:gpt4v-colour}
\end{figure}
\FloatBarrier


\subsection{Comparison with TAMP Experiments} \label{sec:pddl_exps}

To compare the \ourmodel with a TAMP framework, we use Planning Domain Definition Language (PDDL)~\citep{mcdermott1998pddl} to define the domain of Cabinet Tasks 1-3 in Table~\ref{tab:TaskDescription} in Section~\ref{sec:env-details} as follows:

\begin{lstlisting}[language=lisp, breaklines=true, label=lst:pddldomain, numbers=left, caption={PDDL domain definition for Tasks 1-3.}, escapeinside={<*@}{@*>}]
(define (domain pick-place-domain)
  (:requirements :strips :typing :negative-preconditions :conditional-effects)

  ;; Define the object and its possible locations
  (:types
      object
      location
      conf
      robot
      door cabinet cube block - object
      area
      remove_area cabinet_area - area

  )
  ;; define constants
  (:constants
      cube_loc cabinet_loc block_loc remove_loc init_loc door_loc open_door_loc - location
      init_conf robot_conf_cube robot_conf_block robot_conf_cabinet robot_conf_remove robot_conf_door open_door_conf - conf
      robot - robot
      door - door
      cabinet - cabinet
      cube - cube
      block - block
      remove_area - remove_area
      cabinet_area - cabinet_area
  )

  ;; Define predicates
  (:predicates
    (at ?obj - object ?loc - location)  ; the Object at location loc
    (grasped ?obj -object)     ; the object is grasped 
    (at_conf ?conf -conf)               ; the robot is at conf configuration
    (rob_at_loc ?loc -location)    ; the robot is location loc
    (is_free ?rob - robot)      ; the robot hand is free
    (in ?loc -location ?area -area)   ; sampling and certifying the loc location is inside an area type 
    <*@\textbf{\textcolor{blue}{(is\_closed ?door - door)}}@*>        ; the door is closed
    <*@\textbf{\textcolor{blue}{(is\_blocked ?door)}}@*>             ; the door is blocked
    <*@\textbf{\textcolor{blue}{(is\_moveable ?init\_conf -conf ?final\_loc -location ?final\_conf - conf)}}@*>
    ; certifying the robot can move from the initial condition to the goal location/pose with the sampled final configuration 

  )

    ;; Define actions
  (:action remove
      :parameters (?block - block ?door - door ?rob - robot ?init_conf ?final_conf - conf ?init_loc ?final_loc -location)
      :precondition (and 
          (is_blocked ?door)
          (grasped ?block)
          (not (is_free ?rob))
          (at_conf ?init_conf )
          (at ?block ?init_loc)
          (rob_at_loc ?init_loc)
          (in ?final_loc remove_area)
          (is_moveable ?init_conf ?final_loc ?final_conf )
      )
      :effect (and 
          (rob_at_loc ?final_loc )
          (not (rob_at_loc ?init_loc))
          (at ?block ?final_loc)
          (at_conf ?final_conf )
          (is_free ?rob)
          (not (grasped ?block))
          (not (is_blocked ?door))
          (not (at_conf ?init_conf ))
      )
  )

      (:action place
      :parameters (?cube - cube ?door - door ?rob - robot ?init_conf ?final_conf - conf ?init_loc ?final_loc -location)
      :precondition (and 
          (not (is_closed ?door))
          (not (is_blocked ?door))
          (grasped ?cube)
          (not (is_free ?rob))
          (at_conf ?init_conf )
          (at ?cube ?init_loc)
          (rob_at_loc ?init_loc)
          (in ?final_loc cabinet_area) 
          (is_moveable ?init_conf ?final_loc ?final_conf )
      )
      :effect (and
          (rob_at_loc ?final_loc )
          (at ?cube ?final_loc)
          (at_conf ?final_conf )
          (not (at_conf ?init_conf ))
          (is_free ?rob)
          (not (grasped ?cube))
      )
  )

  (:action pick
    :parameters (?init_conf ?final_conf -conf ?obj - object ?loc - location ?rob -robot)
    :precondition (and
      (at_conf ?init_conf )
      (at ?obj ?loc)
      (is_free ?rob )
      (is_moveable ?init_conf ?loc ?final_conf )
      )
    :effect (and
      (at_conf ?final_conf)
      (not (at_conf ?init_conf))
      (rob_at_loc ?loc)
      (not (is_free ?rob))
      (grasped ?obj)
    )
  )

  (:action open
      :parameters (?init_conf -conf ?door -door ?rob -robot )
      :precondition (and 
        (is_closed ?door)
        (not (is_blocked ?door))
        (grasped ?door)
        (not (is_free ?rob))
        (at_conf ?init_conf )
        (is_moveable ?init_conf open_door_loc open_door_conf )
      )
      :effect (and 
        (not (is_closed ?door))
        (not (grasped ?door))
        (is_free ?rob)
        (not (at_conf ?init_conf ))
        (at_conf open_door_conf)
        (rob_at_loc open_door_loc)
        (at door open_door_loc)
        )
      )
  
     <*@\textbf{\textcolor{olive}{(:action is\_moveable\_cube}}@*>
         <*@\textbf{\textcolor{olive}{:parameters ()}}@*>
         <*@\textbf{\textcolor{olive}{:precondition (and }}@*>
           <*@\textbf{\textcolor{olive}{(not (is\_closed door))}}@*>
           <*@\textbf{\textcolor{olive}{(not (is\_blocked door))}}@*>
         <*@\textbf{\textcolor{olive}{)}}@*>
         <*@\textbf{\textcolor{olive}{:effect (is\_moveable robot\_conf\_cube cabinet\_loc robot\_conf\_cabinet) }}@*>
     <*@\textbf{\textcolor{olive}{)}}@*>
  
)
\end{lstlisting}

\newpage
Tasks 1-3 problem definition are written as follows:

\begin{lstlisting}[language=lisp, label=lst:pddlproblem, numbers=left, caption={PDDL problem definition for Cabinet Tasks 1-3.}, escapeinside={<*@}{@*>}]

(define (problem pick-place-problem)
  (:domain pick-place-domain)

  ;; Define objects
  (:objects

  )

  ;; Define initial state
  (:init
    (at cube cube_loc)
    (at door door_loc)
    (at block block_loc)
    (at_conf init_conf)
    (rob_at_loc init_loc)
    (in remove_loc remove_area)
    (in cabinet_loc cabinet_area)
    (is_free robot)
    <*@\textbf{\textcolor{blue}{(is\_moveable init\_conf cube\_loc robot\_conf\_cube)}}@*>
    <*@\textbf{\textcolor{blue}{(is\_moveable open\_door\_conf cube\_loc robot\_conf\_cube)}}@*>
    <*@\textbf{\textcolor{blue}{(is\_moveable robot\_conf\_block remove\_loc robot\_conf\_remove)}}@*>
    <*@\textbf{\textcolor{blue}{(is\_moveable robot\_conf\_remove door\_loc robot\_conf\_door)}}@*>
    <*@\textbf{\textcolor{blue}{(is\_moveable init\_conf block\_loc robot\_conf\_block)}}@*>
    <*@\textbf{\textcolor{blue}{(is\_moveable init\_conf door\_loc robot\_conf\_door)}}@*>
    <*@\textbf{\textcolor{blue}{(is\_moveable robot\_conf\_door open\_door\_loc open\_door\_conf)}}@*>
    
    ;=====================================================================
    ;; commenting the following two initial conditions can change the robot behavior greatly in terms of task plan
    ;; easy mode (Task 1): comment both of the following lines [comment (is_closed door) and (is_blocked door)]
    ;; hard mode (Task 2): comment the second condition [(is_blocked door)]
    ;; expert mode (Task 3): keep both of the following conditions uncommented.
    ;=====================================================================
    
    ; door is closed at the begining
    <*@\textbf{\textcolor{blue}{(is\_closed door)}}@*>
    ; door is blocked at the begining
    <*@\textbf{\textcolor{blue}{(is\_blocked door)}}@*>

    )

  ;; Define goal
  (:goal
    (and
        ;; only picking the cube
        ; (grasped block)
        
        ;; picking and placing the cube inside the cabinet
        (at cube cabinet_loc)
          ; (at block remove_loc)
          ; (grasped door)

    )
  )
)

\end{lstlisting}

The cabinet task problem is solved using the lpg-td~\citep{gerevini2006approach} solver from the planutils library\footnote{\url{https://github.com/AI-Planning/planutils}}.

As seen in Snippets~\ref{lst:pddldomain} and~\ref{lst:pddlproblem}, even for three cabinet tasks 1-3, domain and problem definitions require careful and laborious attention. Task-solving details are outlined in Snippets with blue and olive colours. While our method discovers information through interaction and reasoning over perceiver's feedback, the PDDL solver relies on ground truth (highlighted in blue) and rules (example in olive) provided by the user for problem resolution.

For the same task, PDDLStream~\citep{ICAPS20paper186} offers an alternative using a Task and Motion Planning (TAMP) framework. Rather than human-grounded truth information, a motion planning framework certifies predicates through \textit{streams}. However, this requires user-defined rules for success or failure and a motion planner. Solving long-horizon problems with PDDLStream may become computationally expensive~\citep{khodeir2023learning}.

\ourmodel can robustly solve long-horizon multi-stage problems through interaction with the environment and reasoning based on the perceiver's feedback. This capability enables \ourmodel to uncover underlying rules without the need for an additional domain description and ground truth information.

%% file: rss-main.bbl
\begin{thebibliography}{62}
\providecommand{\natexlab}[1]{#1}
\providecommand{\url}[1]{\texttt{#1}}
\expandafter\ifx\csname urlstyle\endcsname\relax
  \providecommand{\doi}[1]{doi: #1}\else
  \providecommand{\doi}{doi: \begingroup \urlstyle{rm}\Url}\fi

\bibitem[Aeronautiques et~al.(1998)Aeronautiques, Howe, Knoblock, McDermott, Ram, Veloso, Weld, SRI, Barrett, Christianson, et~al.]{aeronautiques1998pddl}
Constructions Aeronautiques, Adele Howe, Craig Knoblock, ISI~Drew McDermott, Ashwin Ram, Manuela Veloso, Daniel Weld, David~Wilkins SRI, Anthony Barrett, Dave Christianson, et~al.
\newblock {PDDL} - the planning domain definition language.
\newblock \emph{Tech. Rep.}, 1998.

\bibitem[Bai et~al.(2023)Bai, Bai, Yang, Wang, Tan, Wang, Lin, Zhou, and Zhou]{Qwen-VL}
Jinze Bai, Shuai Bai, Shusheng Yang, Shijie Wang, Sinan Tan, Peng Wang, Junyang Lin, Chang Zhou, and Jingren Zhou.
\newblock Qwen-vl: A frontier large vision-language model with versatile abilities.
\newblock \emph{arXiv preprint arXiv:2308.12966}, 2023.

\bibitem[Baier et~al.(2009)Baier, Bacchus, and McIlraith]{baier2009heuristic}
Jorge~A Baier, Fahiem Bacchus, and Sheila~A McIlraith.
\newblock A heuristic search approach to planning with temporally extended preferences.
\newblock \emph{Artif. Intell.}, 2009.

\bibitem[Bang et~al.(2023)Bang, Cahyawijaya, Lee, Dai, Su, Wilie, Lovenia, Ji, Yu, Chung, et~al.]{bang2023multitask}
Yejin Bang, Samuel Cahyawijaya, Nayeon Lee, Wenliang Dai, Dan Su, Bryan Wilie, Holy Lovenia, Ziwei Ji, Tiezheng Yu, Willy Chung, et~al.
\newblock A multitask, multilingual, multimodal evaluation of chatgpt on reasoning, hallucination, and interactivity.
\newblock \emph{arXiv preprint arXiv:2302.04023}, 2023.

\bibitem[Baumli et~al.(2023)Baumli, Baveja, Behbahani, Chan, Comanici, Flennerhag, Gazeau, Holsheimer, Horgan, Laskin, et~al.]{baumli2023vision}
Kate Baumli, Satinder Baveja, Feryal Behbahani, Harris Chan, Gheorghe Comanici, Sebastian Flennerhag, Maxime Gazeau, Kristian Holsheimer, Dan Horgan, Michael Laskin, et~al.
\newblock Vision-language models as a source of rewards.
\newblock \emph{arXiv preprint arXiv:2312.09187}, 2023.

\bibitem[Brohan et~al.(2022)Brohan, Brown, Carbajal, Chebotar, Dabis, Finn, Gopalakrishnan, Hausman, Herzog, Hsu, et~al.]{brohan2022rt}
Anthony Brohan, Noah Brown, Justice Carbajal, Yevgen Chebotar, Joseph Dabis, Chelsea Finn, Keerthana Gopalakrishnan, Karol Hausman, Alex Herzog, Jasmine Hsu, et~al.
\newblock Rt-1: Robotics transformer for real-world control at scale.
\newblock \emph{arXiv preprint arXiv:2212.06817}, 2022.

\bibitem[Brohan et~al.(2023{\natexlab{a}})Brohan, Brown, Carbajal, Chebotar, Chen, Choromanski, Ding, Driess, Dubey, Finn, et~al.]{brohan2023rt}
Anthony Brohan, Noah Brown, Justice Carbajal, Yevgen Chebotar, Xi~Chen, Krzysztof Choromanski, Tianli Ding, Danny Driess, Avinava Dubey, Chelsea Finn, et~al.
\newblock Rt-2: Vision-language-action models transfer web knowledge to robotic control.
\newblock \emph{arXiv preprint arXiv:2307.15818}, 2023{\natexlab{a}}.

\bibitem[Brohan et~al.(2023{\natexlab{b}})Brohan, Chebotar, Finn, Hausman, Herzog, Ho, Ibarz, Irpan, Jang, Julian, et~al.]{brohan2023can}
Anthony Brohan, Yevgen Chebotar, Chelsea Finn, Karol Hausman, Alexander Herzog, Daniel Ho, Julian Ibarz, Alex Irpan, Eric Jang, Ryan Julian, et~al.
\newblock Do as i can, not as i say: Grounding language in robotic affordances.
\newblock In \emph{Conference on Robot Learning}, pages 287--318. PMLR, 2023{\natexlab{b}}.

\bibitem[Chen et~al.(2024)Chen, Xu, Kirmani, Ichter, Driess, Florence, Sadigh, Guibas, and Xia]{chen2024spatialvlm}
Boyuan Chen, Zhuo Xu, Sean Kirmani, Brian Ichter, Danny Driess, Pete Florence, Dorsa Sadigh, Leonidas Guibas, and Fei Xia.
\newblock Spatialvlm: Endowing vision-language models with spatial reasoning capabilities.
\newblock \emph{arXiv preprint arXiv:2401.12168}, 2024.

\bibitem[Dai et~al.(2023)Dai, Liu, Ji, Su, and Fung]{dai-etal-2023-plausible}
Wenliang Dai, Zihan Liu, Ziwei Ji, Dan Su, and Pascale Fung.
\newblock Plausible may not be faithful: Probing object hallucination in vision-language pre-training.
\newblock In \emph{Proceedings of the 17th Conference of the European Chapter of the Association for Computational Linguistics}, pages 2136--2148, Dubrovnik, Croatia, May 2023. Association for Computational Linguistics.
\newblock \doi{10.18653/v1/2023.eacl-main.156}.
\newblock URL \url{https://aclanthology.org/2023.eacl-main.156}.

\bibitem[Darvish et~al.(2024)Darvish, Skreta, Zhao, Yoshikawa, Som, Bogdanovic, Cao, Hao, Xu, Aspuru-Guzik, Garg, and Shkurti]{darvish2024organa}
Kourosh Darvish, Marta Skreta, Yuchi Zhao, Naruki Yoshikawa, Sagnik Som, Miroslav Bogdanovic, Yang Cao, Han Hao, Haoping Xu, Al{\'a}n Aspuru-Guzik, Animesh Garg, and Florian Shkurti.
\newblock Organa: A robotic assistant for automated chemistry experimentation and characterization.
\newblock \emph{arXiv preprint arXiv:2401.06949}, 2024.

\bibitem[Driess et~al.(2023)Driess, Xia, Sajjadi, Lynch, Chowdhery, Ichter, Wahid, Tompson, Vuong, Yu, et~al.]{driess2023palm}
Danny Driess, Fei Xia, Mehdi~SM Sajjadi, Corey Lynch, Aakanksha Chowdhery, Brian Ichter, Ayzaan Wahid, Jonathan Tompson, Quan Vuong, Tianhe Yu, et~al.
\newblock Palm-e: An embodied multimodal language model.
\newblock \emph{arXiv preprint arXiv:2303.03378}, 2023.

\bibitem[Garcia et~al.(1989)Garcia, Prett, and Morari]{garcia1989model}
Carlos~E Garcia, David~M Prett, and Manfred Morari.
\newblock Model predictive control: Theory and practice—a survey.
\newblock \emph{Automatica}, 25\penalty0 (3):\penalty0 335--348, 1989.

\bibitem[Garrett et~al.(2020)Garrett, Lozano-P{\'{e}}rez, and Kaelbling]{ICAPS20paper186}
Caelan~Reed Garrett, Tom{\'{a}}s Lozano-P{\'{e}}rez, and Leslie~Pack Kaelbling.
\newblock {PDDLStream}: Integrating symbolic planners and blackbox samplers via optimistic adaptive planning.
\newblock In \emph{Proceedings of the 30th Int. Conf. on Automated Planning and Scheduling ({ICAPS})}, pages 440--448. {AAAI} Press, 2020.

\bibitem[Garrett et~al.(2021)Garrett, Chitnis, Holladay, Kim, Silver, Kaelbling, and Lozano-P{\'e}rez]{garrett2021integrated}
Caelan~Reed Garrett, Rohan Chitnis, Rachel Holladay, Beomjoon Kim, Tom Silver, Leslie~Pack Kaelbling, and Tom{\'a}s Lozano-P{\'e}rez.
\newblock Integrated task and motion planning.
\newblock \emph{Annual review of control, robotics, and autonomous systems}, 4:\penalty0 265--293, 2021.

\bibitem[Gerevini et~al.(2006)Gerevini, Saetti, and Serina]{gerevini2006approach}
Alfonso Gerevini, Alessandro Saetti, and Ivan Serina.
\newblock An approach to temporal planning and scheduling in domains with predictable exogenous events.
\newblock \emph{Journal of Artificial Intelligence Research}, 25:\penalty0 187--231, 2006.

\bibitem[Goyal et~al.(2019)Goyal, Niekum, and Mooney]{goyal2019using}
Prasoon Goyal, Scott Niekum, and Raymond~J Mooney.
\newblock Using natural language for reward shaping in reinforcement learning.
\newblock \emph{arXiv preprint arXiv:1903.02020}, 2019.

\bibitem[Guo et~al.(2023)Guo, Wang, Zha, Jiang, and Chen]{guo2023doremi}
Yanjiang Guo, Yen-Jen Wang, Lihan Zha, Zheyuan Jiang, and Jianyu Chen.
\newblock Doremi: Grounding language model by detecting and recovering from plan-execution misalignment.
\newblock \emph{arXiv preprint arXiv:2307.00329}, 2023.

\bibitem[Ha et~al.(2023)Ha, Florence, and Song]{ha2023scaling}
Huy Ha, Pete Florence, and Shuran Song.
\newblock Scaling up and distilling down: Language-guided robot skill acquisition.
\newblock \emph{arXiv preprint arXiv:2307.14535}, 2023.

\bibitem[Heo et~al.(2023)Heo, Lee, Lee, and Lim]{heo2023furniturebench}
Minho Heo, Youngwoon Lee, Doohyun Lee, and Joseph~J Lim.
\newblock Furniturebench: Reproducible real-world benchmark for long-horizon complex manipulation.
\newblock \emph{arXiv preprint arXiv:2305.12821}, 2023.

\bibitem[Howell et~al.(2022)Howell, Gileadi, Tunyasuvunakool, Zakka, Erez, and Tassa]{howell2022}
Taylor Howell, Nimrod Gileadi, Saran Tunyasuvunakool, Kevin Zakka, Tom Erez, and Yuval Tassa.
\newblock {Predictive Sampling: Real-time Behaviour Synthesis with MuJoCo}.
\newblock dec 2022.
\newblock \doi{10.48550/arXiv.2212.00541}.
\newblock URL \url{https://arxiv.org/abs/2212.00541}.

\bibitem[Huang et~al.(2022{\natexlab{a}})Huang, Abbeel, Pathak, and Mordatch]{huang2022language}
Wenlong Huang, Pieter Abbeel, Deepak Pathak, and Igor Mordatch.
\newblock Language models as zero-shot planners: Extracting actionable knowledge for embodied agents.
\newblock \emph{arXiv preprint arXiv:2201.07207}, 2022{\natexlab{a}}.

\bibitem[Huang et~al.(2022{\natexlab{b}})Huang, Xia, Xiao, Chan, Liang, Florence, Zeng, Tompson, Mordatch, Chebotar, et~al.]{huang2022inner}
Wenlong Huang, Fei Xia, Ted Xiao, Harris Chan, Jacky Liang, Pete Florence, Andy Zeng, Jonathan Tompson, Igor Mordatch, Yevgen Chebotar, et~al.
\newblock Inner monologue: Embodied reasoning through planning with language models.
\newblock \emph{arXiv preprint arXiv:2207.05608}, 2022{\natexlab{b}}.

\bibitem[Hussein et~al.(2017)Hussein, Gaber, Elyan, and Jayne]{hussein2017imitation}
Ahmed Hussein, Mohamed~Medhat Gaber, Eyad Elyan, and Chrisina Jayne.
\newblock Imitation learning: A survey of learning methods.
\newblock \emph{ACM Computing Surveys (CSUR)}, 50\penalty0 (2):\penalty0 1--35, 2017.

\bibitem[Khodeir et~al.(2023)Khodeir, Agro, and Shkurti]{khodeir2023learning}
Mohamed Khodeir, Ben Agro, and Florian Shkurti.
\newblock Learning to search in task and motion planning with streams.
\newblock \emph{IEEE Robotics and Automation Letters}, 8\penalty0 (4):\penalty0 1983--1990, 2023.

\bibitem[Kirillov et~al.(2023{\natexlab{a}})Kirillov, Mintun, Ravi, Mao, Rolland, Gustafson, Xiao, Whitehead, Berg, Lo, Doll{\'a}r, and Girshick]{kirillov2023segany}
Alexander Kirillov, Eric Mintun, Nikhila Ravi, Hanzi Mao, Chloe Rolland, Laura Gustafson, Tete Xiao, Spencer Whitehead, Alexander~C. Berg, Wan-Yen Lo, Piotr Doll{\'a}r, and Ross Girshick.
\newblock Segment anything.
\newblock \emph{arXiv:2304.02643}, 2023{\natexlab{a}}.

\bibitem[Kirillov et~al.(2023{\natexlab{b}})Kirillov, Mintun, Ravi, Mao, Rolland, Gustafson, Xiao, Whitehead, Berg, Lo, et~al.]{kirillov2023segment}
Alexander Kirillov, Eric Mintun, Nikhila Ravi, Hanzi Mao, Chloe Rolland, Laura Gustafson, Tete Xiao, Spencer Whitehead, Alexander~C Berg, Wan-Yen Lo, et~al.
\newblock Segment anything.
\newblock \emph{arXiv preprint arXiv:2304.02643}, 2023{\natexlab{b}}.

\bibitem[Kwon et~al.(2023)Kwon, Xie, Bullard, and Sadigh]{kwon2023reward}
Minae Kwon, Sang~Michael Xie, Kalesha Bullard, and Dorsa Sadigh.
\newblock Reward design with language models.
\newblock \emph{arXiv preprint arXiv:2303.00001}, 2023.

\bibitem[Li et~al.(2023)Li, Zhang, Wong, Gokmen, Srivastava, Mart{\'\i}n-Mart{\'\i}n, Wang, Levine, Lingelbach, Sun, et~al.]{li2023behavior}
Chengshu Li, Ruohan Zhang, Josiah Wong, Cem Gokmen, Sanjana Srivastava, Roberto Mart{\'\i}n-Mart{\'\i}n, Chen Wang, Gabrael Levine, Michael Lingelbach, Jiankai Sun, et~al.
\newblock Behavior-1k: A benchmark for embodied ai with 1,000 everyday activities and realistic simulation.
\newblock In \emph{Conference on Robot Learning}, pages 80--93. PMLR, 2023.

\bibitem[Liang et~al.(2022)Liang, Huang, Xia, Xu, Hausman, Ichter, Florence, and Zeng]{liang2022code}
Jacky Liang, Wenlong Huang, Fei Xia, Peng Xu, Karol Hausman, Brian Ichter, Pete Florence, and Andy Zeng.
\newblock Code as policies: Language model programs for embodied control.
\newblock \emph{arXiv preprint}, 2022.
\newblock \doi{10.48550/arXiv.2209.07753}.

\bibitem[Lin et~al.(2022)Lin, Fried, Klein, and Dragan]{lin2022inferring}
Jessy Lin, Daniel Fried, Dan Klein, and Anca Dragan.
\newblock Inferring rewards from language in context.
\newblock \emph{arXiv preprint arXiv:2204.02515}, 2022.

\bibitem[Lin et~al.(2023)Lin, Agia, Migimatsu, Pavone, and Bohg]{lin2023text2motion}
Kevin Lin, Christopher Agia, Toki Migimatsu, Marco Pavone, and Jeannette Bohg.
\newblock Text2motion: From natural language instructions to feasible plans.
\newblock \emph{arXiv preprint arXiv:2303.12153}, 2023.

\bibitem[Liu et~al.(2023{\natexlab{a}})Liu, Li, Li, and Lee]{liu2023improvedllava}
Haotian Liu, Chunyuan Li, Yuheng Li, and Yong~Jae Lee.
\newblock Improved baselines with visual instruction tuning, 2023{\natexlab{a}}.

\bibitem[Liu et~al.(2022)Liu, Wei, Gu, Wu, Vosoughi, Cui, Zhou, and Dai]{liu2022mind}
Ruibo Liu, Jason Wei, Shixiang~Shane Gu, Te-Yen Wu, Soroush Vosoughi, Claire Cui, Denny Zhou, and Andrew~M Dai.
\newblock Mind's eye: Grounded language model reasoning through simulation.
\newblock \emph{arXiv preprint arXiv:2210.05359}, 2022.

\bibitem[Liu et~al.(2023{\natexlab{b}})Liu, Zeng, Ren, Li, Zhang, Yang, Li, Yang, Su, Zhu, et~al.]{liu2023grounding}
Shilong Liu, Zhaoyang Zeng, Tianhe Ren, Feng Li, Hao Zhang, Jie Yang, Chunyuan Li, Jianwei Yang, Hang Su, Jun Zhu, et~al.
\newblock Grounding dino: Marrying dino with grounded pre-training for open-set object detection.
\newblock \emph{arXiv preprint arXiv:2303.05499}, 2023{\natexlab{b}}.

\bibitem[Liu et~al.(2023{\natexlab{c}})Liu, Bahety, and Song]{liu2023reflect}
Zeyi Liu, Arpit Bahety, and Shuran Song.
\newblock Reflect: Summarizing robot experiences for failure explanation and correction.
\newblock \emph{arXiv preprint arXiv:2306.15724}, 2023{\natexlab{c}}.

\bibitem[Mahmoudieh et~al.(2022)Mahmoudieh, Pathak, and Darrell]{mahmoudieh2022zero}
Parsa Mahmoudieh, Deepak Pathak, and Trevor Darrell.
\newblock Zero-shot reward specification via grounded natural language.
\newblock In \emph{International Conference on Machine Learning}, pages 14743--14752. PMLR, 2022.

\bibitem[McDermott et~al.(1998)McDermott, Ghallab, Howe, Knoblock, Ram, Veloso, Weld, and Wilkins]{mcdermott1998pddl}
Drew McDermott, Malik Ghallab, Adele Howe, Craig Knoblock, Ashwin Ram, Manuela Veloso, Daniel Weld, and David Wilkins.
\newblock Pddl-the planning domain definition language.
\newblock 1998.

\bibitem[Mehr et~al.(2020)Mehr, Craven, Leonov, Keenan, and Cronin]{10.1126/science.abc2986}
S~Hessam~M Mehr, Matthew Craven, Artem~I Leonov, Graham Keenan, and Leroy Cronin.
\newblock A universal system for digitization and automatic execution of the chemical synthesis literature.
\newblock \emph{Science}, 370\penalty0 (6512):\penalty0 101--108, 2020.

\bibitem[Miyaoka et~al.(2023)Miyaoka, Inoue, and Nii]{miyaoka2023chatmpc}
Yuya Miyaoka, Masaki Inoue, and Tomotaka Nii.
\newblock Chatmpc: Natural language based mpc personalization.
\newblock \emph{arXiv preprint arXiv:2309.05952}, 2023.

\bibitem[{OpenAI}(2023)]{gpt4v-2023}
{OpenAI}.
\newblock {GPT-4V(ision) system card}, 2023.
\newblock URL \url{https://openai.com/research/gpt-4v-system-card}.

\bibitem[OpenAI et~al.(2023)]{openai2023gpt4}
OpenAI et~al.
\newblock Gpt-4 technical report, 2023.

\bibitem[Radford et~al.(2021)Radford, Kim, Hallacy, Ramesh, Goh, Agarwal, Sastry, Askell, Mishkin, Clark, et~al.]{radford2021learning}
Alec Radford, Jong~Wook Kim, Chris Hallacy, Aditya Ramesh, Gabriel Goh, Sandhini Agarwal, Girish Sastry, Amanda Askell, Pamela Mishkin, Jack Clark, et~al.
\newblock Learning transferable visual models from natural language supervision.
\newblock In \emph{International conference on machine learning}, pages 8748--8763. PMLR, 2021.

\bibitem[Raman et~al.(2023)Raman, Cohen, Paulius, Idrees, Rosen, Mooney, and Tellex]{raman2023cape}
Shreyas~Sundara Raman, Vanya Cohen, David Paulius, Ifrah Idrees, Eric Rosen, Ray Mooney, and Stefanie Tellex.
\newblock Cape: Corrective actions from precondition errors using large language models.
\newblock In \emph{2nd Workshop on Language and Robot Learning: Language as Grounding}, 2023.

\bibitem[Rana et~al.(2023)Rana, Haviland, Garg, Abou-Chakra, Reid, and Suenderhauf]{rana2023sayplan}
Krishan Rana, Jesse Haviland, Sourav Garg, Jad Abou-Chakra, Ian Reid, and Niko Suenderhauf.
\newblock Sayplan: Grounding large language models using 3d scene graphs for scalable task planning.
\newblock In \emph{7th Annual Conference on Robot Learning}, 2023.
\newblock URL \url{https://openreview.net/forum?id=wMpOMO0Ss7a}.

\bibitem[Rawlings(2000)]{rawlings2000tutorial}
James~B Rawlings.
\newblock Tutorial overview of model predictive control.
\newblock \emph{IEEE control systems magazine}, 20\penalty0 (3):\penalty0 38--52, 2000.

\bibitem[Singh et~al.(2022)Singh, Blukis, Mousavian, Goyal, Xu, Tremblay, Fox, Thomason, and Garg]{singh2022progprompt}
Ishika Singh, Valts Blukis, Arsalan Mousavian, Ankit Goyal, Danfei Xu, Jonathan Tremblay, Dieter Fox, Jesse Thomason, and Animesh Garg.
\newblock Progprompt: Generating situated robot task plans using large language models.
\newblock \emph{arXiv preprint arXiv:2209.11302}, 2022.

\bibitem[Skreta et~al.(2023)Skreta, Yoshikawa, Arellano-Rubach, Ji, Kristensen, Darvish, Aspuru-Guzik, Shkurti, and Garg]{skreta2023errors}
Marta Skreta, Naruki Yoshikawa, Sebastian Arellano-Rubach, Zhi Ji, Lasse~Bj{\o}rn Kristensen, Kourosh Darvish, Al{\'a}n Aspuru-Guzik, Florian Shkurti, and Animesh Garg.
\newblock Errors are useful prompts: Instruction guided task programming with verifier-assisted iterative prompting.
\newblock \emph{arXiv preprint arXiv:2303.14100}, 2023.

\bibitem[Todorov et~al.(2012)Todorov, Erez, and Tassa]{todorov2012mujoco}
Emanuel Todorov, Tom Erez, and Yuval Tassa.
\newblock Mujoco: A physics engine for model-based control.
\newblock In \emph{2012 IEEE/RSJ International Conference on Intelligent Robots and Systems}, pages 5026--5033. IEEE, 2012.
\newblock \doi{10.1109/IROS.2012.6386109}.

\bibitem[Wang et~al.(2023{\natexlab{a}})Wang, Xie, Jiang, Mandlekar, Xiao, Zhu, Fan, and Anandkumar]{wang2023voyager}
Guanzhi Wang, Yuqi Xie, Yunfan Jiang, Ajay Mandlekar, Chaowei Xiao, Yuke Zhu, Linxi Fan, and Anima Anandkumar.
\newblock Voyager: An open-ended embodied agent with large language models.
\newblock \emph{arXiv preprint arXiv: Arxiv-2305.16291}, 2023{\natexlab{a}}.

\bibitem[Wang et~al.(2023{\natexlab{b}})Wang, Ma, Feng, Zhang, Yang, Zhang, Chen, Tang, Chen, Lin, et~al.]{wang2023survey}
Lei Wang, Chen Ma, Xueyang Feng, Zeyu Zhang, Hao Yang, Jingsen Zhang, Zhiyuan Chen, Jiakai Tang, Xu~Chen, Yankai Lin, et~al.
\newblock A survey on large language model based autonomous agents.
\newblock \emph{arXiv preprint arXiv:2308.11432}, 2023{\natexlab{b}}.

\bibitem[Wang et~al.(2023{\natexlab{c}})Wang, Cai, Liu, Ma, and Liang]{wang2023describe}
Zihao Wang, Shaofei Cai, Anji Liu, Xiaojian Ma, and Yitao Liang.
\newblock Describe, explain, plan and select: Interactive planning with large language models enables open-world multi-task agents.
\newblock \emph{arXiv preprint arXiv:2302.01560}, 2023{\natexlab{c}}.

\bibitem[Wei et~al.(2022)Wei, Wang, Schuurmans, Bosma, Xia, Chi, Le, Zhou, et~al.]{wei2022chain}
Jason Wei, Xuezhi Wang, Dale Schuurmans, Maarten Bosma, Fei Xia, Ed~Chi, Quoc~V Le, Denny Zhou, et~al.
\newblock Chain-of-thought prompting elicits reasoning in large language models.
\newblock \emph{Advances in Neural Information Processing Systems}, 35:\penalty0 24824--24837, 2022.

\bibitem[Wu et~al.(2023)Wu, Martin-Martin, and Fei-Fei]{wu2023m}
Bohan Wu, Roberto Martin-Martin, and Li~Fei-Fei.
\newblock M-ember: Tackling long-horizon mobile manipulation via factorized domain transfer.
\newblock \emph{arXiv preprint arXiv:2305.13567}, 2023.

\bibitem[Xi et~al.(2023)Xi, Chen, Guo, He, Ding, Hong, Zhang, Wang, Jin, Zhou, et~al.]{xi2023rise}
Zhiheng Xi, Wenxiang Chen, Xin Guo, Wei He, Yiwen Ding, Boyang Hong, Ming Zhang, Junzhe Wang, Senjie Jin, Enyu Zhou, et~al.
\newblock The rise and potential of large language model based agents: A survey.
\newblock \emph{arXiv preprint arXiv:2309.07864}, 2023.

\bibitem[Xie et~al.(2023)Xie, Zhao, Wu, Liu, Luo, Zhong, Yang, and Yu]{xie2023text2reward}
Tianbao Xie, Siheng Zhao, Chen~Henry Wu, Yitao Liu, Qian Luo, Victor Zhong, Yanchao Yang, and Tao Yu.
\newblock Text2reward: Automated dense reward function generation for reinforcement learning.
\newblock \emph{arXiv preprint arXiv:2309.11489}, 2023.

\bibitem[Yao et~al.(2023)Yao, Zhao, Yu, Du, Shafran, Narasimhan, and Cao]{yao2023react}
Shunyu Yao, Jeffrey Zhao, Dian Yu, Nan Du, Izhak Shafran, Karthik~R Narasimhan, and Yuan Cao.
\newblock React: Synergizing reasoning and acting in language models.
\newblock In \emph{The Eleventh International Conference on Learning Representations}, 2023.
\newblock URL \url{https://openreview.net/forum?id=WE_vluYUL-X}.

\bibitem[Yoshikawa et~al.(2023)Yoshikawa, Skreta, Darvish, Arellano-Rubach, Ji, Bj{\o}rn~Kristensen, Li, Zhao, Xu, Kuramshin, Aspuru-Guzik, Shkurti, and Garg]{yoshikawa2023large}
Naruki Yoshikawa, Marta Skreta, Kourosh Darvish, Sebastian Arellano-Rubach, Zhi Ji, Lasse Bj{\o}rn~Kristensen, Andrew~Zou Li, Yuchi Zhao, Haoping Xu, Artur Kuramshin, Al{\'a}n Aspuru-Guzik, Florian Shkurti, and Animesh Garg.
\newblock Large language models for chemistry robotics.
\newblock \emph{Autonomous Robots}, 47\penalty0 (8):\penalty0 1057--1086, 2023.

\bibitem[Yu et~al.(2023)Yu, Gileadi, Fu, Kirmani, Lee, Arenas, Chiang, Erez, Hasenclever, Humplik, et~al.]{yu2023language}
Wenhao Yu, Nimrod Gileadi, Chuyuan Fu, Sean Kirmani, Kuang-Huei Lee, Montse~Gonzalez Arenas, Hao-Tien~Lewis Chiang, Tom Erez, Leonard Hasenclever, Jan Humplik, et~al.
\newblock Language to rewards for robotic skill synthesis.
\newblock \emph{arXiv preprint arXiv:2306.08647}, 2023.

\bibitem[Zha et~al.(2023)Zha, Cui, Lin, Kwon, Arenas, Zeng, Xia, and Sadigh]{zha2023distilling}
Lihan Zha, Yuchen Cui, Li-Heng Lin, Minae Kwon, Montserrat~Gonzalez Arenas, Andy Zeng, Fei Xia, and Dorsa Sadigh.
\newblock Distilling and retrieving generalizable knowledge for robot manipulation via language corrections.
\newblock \emph{arXiv preprint arXiv:2311.10678}, 2023.

\bibitem[Zhou and Garg(2023)]{zhou2023learning}
Zihan Zhou and Animesh Garg.
\newblock Learning achievement structure for structured exploration in domains with sparse reward, 2023.

\bibitem[Zou et~al.(2023)Zou, Chen, Shi, Guo, and Ye]{zou2023object}
Zhengxia Zou, Keyan Chen, Zhenwei Shi, Yuhong Guo, and Jieping Ye.
\newblock Object detection in 20 years: A survey.
\newblock \emph{Proceedings of the IEEE}, 2023.

\end{thebibliography}
